\documentclass[journal]{IEEEtran} % Comment this line out if you need a4paper

\IEEEoverridecommandlockouts                              % This command is only needed if 
\pdfoutput=1

\usepackage{graphicx}
\usepackage{epsfig} % for postscript graphics files
\usepackage{mathptmx}
\usepackage{amsmath} % assumes amsmath package installed

\usepackage{amssymb}  % assumes amsmath package installed
\usepackage{amsthm}
\usepackage{wasysym}
\usepackage[mathcal]{euscript}
\usepackage{mathrsfs}
\usepackage{bm}
\usepackage{bbm}
\usepackage{color}
\usepackage{lipsum}
\usepackage[ruled,vlined,linesnumbered]{algorithm2e}
\usepackage[colorlinks=true,linkcolor=black,anchorcolor=black,citecolor=black,filecolor=black,menucolor=black,runcolor=black,urlcolor=black]{hyperref}
\usepackage{etoolbox}
\usepackage[table,xcdraw,dvipsnames]{xcolor}
\usepackage{multirow}
\usepackage{mathtools}
\usepackage{algpseudocode}
\usepackage{xspace}

\hypersetup{
    colorlinks=true,
    linkcolor=blue,
    filecolor=magenta,      
    urlcolor=cyan,
    pdftitle={Overleaf Example},
    pdfpagemode=FullScreen,
    }

\usepackage{outlines}
\let\labelindent\relax
\usepackage{enumitem}
\setenumerate[1]{label=\Roman*.}
\setenumerate[2]{label=\Alph*.}
\setenumerate[3]{label=\arabic*.}
\setenumerate[4]{label=\roman*.}

\usepackage{pifont}% http://ctan.org/pkg/pifont

\usepackage{booktabs}

\makeatletter
\patchcmd{\@makecaption}
  {\scshape}
  {}
  {}
  {}
\makeatother
\newif\ifcommentson
\commentsontrue
\newcounter{PatrickCount}
\addtocounter{PatrickCount}{1}

\newcounter{JonCount}
\addtocounter{JonCount}{1}

\newcounter{BohaoCount}
\addtocounter{BohaoCount}{1}

\newcounter{ThoughtCount}
\addtocounter{ThoughtCount}{1}

\newcounter{ShreyCount}
\addtocounter{ShreyCount}{1}

\newcounter{RamCount}
\addtocounter{RamCount}{1}

\newcounter{FixCount}
\addtocounter{FixCount}{1}

\newtheorem{defn}{Definition}
\newtheorem{rem}[defn]{Remark}
\newtheorem{lem}[defn]{Lemma}

\newtheorem{assum}[defn]{Assumption}
\newtheorem{ex}[defn]{Example}
\newtheorem{thm}[defn]{Theorem}
\newtheorem{cor}[defn]{Corollary}

\providecommand{\R}{\ensuremath \mathbb{R}}

\providecommand{\N}{\ensuremath \mathbb{N}}

\newcommand{\regtext}[1]{\mathrm{\textnormal{#1}}}

\newcommand{\defemph}[1]{\emph{#1}}
\newcommand{\ts}[1]{\textsuperscript{#1}}

\newcommand{\norm}[1]{\left\Vert#1\right\Vert}
\newcommand{\abs}[1]{\left\vert#1\right\vert}
\newcommand{\pow}[1]{\regtext{pow}\!\left(#1\right)} % shreyas uncommented this 20 july 2022

\newcommand{\trans}{^\top}

\newcommand{\lbl}[1]{_{\regtext{#1}}}

\newcommand{\zeros}{\textit{0}}
\newcommand{\ones}{\textit{1}}
\newcommand{\eye}{\regtext{\textit{I}}}

\newcommand{\interval}[1]{[ #1 ]}
\newcommand{\iv}[1]{[ #1 ]}
\newcommand{\nom}[1]{#1}
\newcommand{\pz}[1]{\mathbf{#1}}
\newcommand{\pzgreek}[1]{\bm{#1}}
\newcommand{\PZ}[1]{\mathcal{PZ}\left(#1\right)}
\newcommand{\pzi}[1]{\pz{ #1 }(\pz{T_i};\pz{K})}
\newcommand{\pzki}[1]{\pz{ #1 }(\pz{T_i};k)}

\newcommand{\setop}[1]{{\mathrm{\textnormal{\texttt{#1}}}}}
\newcommand{\numop}[1]{{\mathrm{\textnormal{\texttt{#1}}}}}
\newcommand{\lb}[1]{\underline{#1}}
\newcommand{\ub}[1]{\overline{#1}}

\newcommand{\pzqi}{\pzi{q}}
\newcommand{\pzqAi}{\pzi{\qA}}
\newcommand{\pzqdi}{\pzi{\dot{q}}}
\newcommand{\pzqdai}{\pzi{\dot{q}_{a}}}
\newcommand{\pzqddi}{\pzi{\ddot{q}}}
\newcommand{\pzqddai}{\pzi{\ddot{q}_{a}}}
\newcommand{\pzqdesi}{\pzi{q_{d}}}
\newcommand{\pzqddesi}{\pzi{\dot{q}_{d}}}
\newcommand{\pzqdddesi}{\pzi{\ddot{q}_{d}}}

\newcommand{\pzqki}{\pzki{q}}
\newcommand{\pzqAki}{\pzki{\qA}}

\newcommand{\pzqji}{\pzi{q_j}}

\newcommand{\pzqli}{\pzi{q_l}}
\newcommand{\pzqdji}{\pzi{\dot{q}_j}}

\newcommand{\pzqdesji}{\pzi{q_{d,j}}}
\newcommand{\pzqddesji}{\pzi{\dot{q}_{d,j}}}
\newcommand{\pzqdddesji}{\pzi{\ddot{q}_{d,j}}}
\newcommand{\pzqdesjki}{\pzki{q_{d,j}}}
\newcommand{\pzqddesjki}{\pzki{\dot{q}_{d,j}}}
\newcommand{\pzqdddesjki}{\pzki{\ddot{q}_{d,j}}}

\newcommand{\pzqjki}{\pzki{q_j}}
\newcommand{\pzqdjki}{\pzki{\dot{q}_j}}
\newcommand{\pzujKi}{\pz{u}(\pzqAi, \nomparams, \intparams)}
\newcommand{\pzujki}{\pz{u}(\pzqAki, \nomparams, \intparams)}
\newcommand{\pzFKjki}{\pz{FK_j}(\pzqki)}
\newcommand{\pzFOjki}{\pz{FO_j}(\pzqki)}
\newcommand{\pzFKjKi}{\pz{FK_j}(\pzqi)}
\newcommand{\pzFOjKi}{\pz{FO_j}(\pzqi)}

\newcommand{\pzpboundj}{\bm{\epsilon}_{\mathbf{p, j}}}
\newcommand{\pzvboundj}{\bm{\epsilon}_{\mathbf{v, j}}}

\newcommand{\pzg}{g}
\newcommand{\pzv}{x}

\newcommand{\pzn}{{n_g}}
\newcommand{\pzgi}{g_i}
\newcommand{\pzei}{\alpha_i}

\newcommand{\q}{q(t)}
\newcommand{\qd}{\dot{q}(t)}
\newcommand{\qdd}{\ddot{q}(t)}

\newcommand{\qj}{q_j(t)}
\newcommand{\ql}{q_l(t)}
\newcommand{\qdj}{\dot{q}_{j}(t)}

\newcommand{\qajdot}{\dot{q}_{a, j}(t)}

\newcommand{\qdesj}{q_{d, j}}

\newcommand{\qdeskj}{q_{d, j}(t; k)}

\newcommand{\nq}{n_q}
\newcommand{\nt}{n_t}
\newcommand{\nf}{n_f}
\newcommand{\Nq}{ N_q }
\newcommand{\Nt}{ N_t }

\newcommand{\err}{e}
\newcommand{\errdot}{\dot{e}}

\newcommand{\errj}{e_j}
\newcommand{\errjdot}{\dot{e}_j}
\newcommand{\errddot}{\ddot{e}}
\newcommand{\roblyap}{V(\qA(t),\Delta)}
\newcommand{\robh}{h(\qA(t),\Delta)}
\newcommand{\robv}{v}
\newcommand{\robr}{r}
\newcommand{\robrj}{r_j}
\newcommand{\robw}{w}
\newcommand{\robrdot}{\dot{r}}
\newcommand{\robH}{H}
\newcommand{\robhmin}{\underline{h}(\qA(t),[\Delta])}
\newcommand{\robhdot}{\dot{h}(\qA(t))}
\newcommand{\roblevel}{V_M}
\newcommand{\robcoeff}{\gamma}
\newcommand{\robKinf}{\alpha}

\newcommand{\ultbound}{\sqrt{\frac{2 \roblevel}{\sigma_m}}}

\newcommand{\vbound}{\epsilon_v}
\newcommand{\pboundj}{\epsilon_{p, j}}

\newcommand{\pboundvec}{E_p}
\newcommand{\vboundvec}{E_v}

\newcommand{\epvarj}{x_{e_{p, j}}}
\newcommand{\evvarj}{x_{e_{v, j}}}
\newcommand{\qA}{q_A}

\providecommand{\R}{\ensuremath \mathbb{R}}
\newcommand{\plan}{\lbl{p}}
\newcommand{\elapsed}{\lbl{e}}

\providecommand{\tfin}{t\lbl{f}}

\newcommand{\bM}{M}
\newcommand{\Mq}{M(\q, \Delta)}
\newcommand{\Mqdot}{\dot{M}(q(t), \Delta)}

\newcommand{\bC}{C}

\newcommand{\Cqd}{C(\q, \qd, \Delta)}

\newcommand{\bG}{G}

\newcommand{\Gqd}{G(\q, \Delta)}

\newcommand{\intparams}{[\Delta]}
\newcommand{\intparamsdb}{[\Delta]\lbl{db}}
\newcommand{\nomparams}{\Delta_0}
\newcommand{\trueparams}{\Delta}

\makeatletter
\newcommand{\smalloplus}{\mathbin{\mathpalette\make@small\oplus}}
\newcommand{\smallotimes}{\mathbin{\mathpalette\make@small\otimes}}

\newcommand{\homtrans}{H}

\newcommand{\wdist}{w}

\newcommand{\wdistinterval}{w}
\newcommand{\lambdamin}{\lambda_m}
\newcommand{\lambdamax}{\lambda_M}
\newcommand{\sigm}{\sigma_{m}}
\newcommand{\sigM}{\sigma_{M}}

\newcommand{\Hquad}{\hspace{0.5em}}

\newcommand{\FK}{\regtext{\small{FK}}}

\newcommand{\FO}{\regtext{\small{FO}}}

\newcommand{\Aobs}{A_O}
\newcommand{\bobs}{b_O}

\newcommand{\nObs}{n_\mathscr{O}}
\newcommand{\obsset}{\mathscr{O}}

\newcommand{\SO}{\regtext{\small{SO}}}

\newcommand{\kj}{k_j}
\newcommand{\Kj}{K_j}

\newcommand{\kjscale}{\eta_{j, 1}}

\newcommand{\kjoffset}{\eta_{j, 2}}
\newcommand{\kvar}{x_k}
\newcommand{\kjvar}{x_{k_j}}

\newcommand{\tvari}{x_{t_{i}}}

\newcommand{\qlim}{q_{j,\regtext{lim}}}
\newcommand{\dqlim}{\dot{q}_{j,\regtext{lim}}}

\newcommand{\ulim}{u_{j,\regtext{lim}}}

\newcommand{\initq}{q_{d_0}}
\newcommand{\initqj}{q_{d, j_{0}}}
\newcommand{\initdq}{\dot{q}_{d_0}}
\newcommand{\initdqj}{\dot{q}_{d, j_{0}}}
\newcommand{\initddq}{\ddot{q}_{d_0}}
\newcommand{\initddqj}{\ddot{q}_{d, j_{0}}}

\newcommand{\jss}{_{j}^{j}}
\newcommand{\jssm}{_{j-1}^{j}}
\newcommand{\jssmu}{_{j}^{j-1}}
\newcommand{\jssmj}{_{j-1, j}^{j}}
\newcommand{\jssp}{_{j+1}^{j}}
\newcommand{\jssmm}{_{j-1}^{j-1}}

\newcommand{\jssa}{_{a, j}^{j}}
\newcommand{\jssc}{_{c, j}^{j}}

\newcommand{\jssmma}{_{a, j-1}^{j-1}}

\newcommand{\qgoal}{q\lbl{goal}}
\newcommand{\qstart}{q\lbl{start}}
\newcommand{\methodname}{{ARMOUR}\xspace}

\newcommand{\wmax}{w_M}

\newcommand{\timestep}{\Delta t}

\usepackage[
    style=ieee,
    doi=false,
    isbn=false,
    url=false,
    eprint=false,
    backend=biber,
    natbib=true
    ]{biblatex}
    
\bibliography{references}

\title{\LARGE \bf
Can't Touch This: Real-Time, Safe Motion Planning and Control for Manipulators Under Uncertainty
}
\author{Jonathan Michaux$^{1}$, Patrick Holmes$^{1}$, Bohao Zhang$^{1}$, Che Chen$^{1}$, Baiyue Wang$^{1}$, Shrey Sahgal$^{1}$, Tiancheng Zhang$^{1}$, Sidhartha Dey$^{4}$, Shreyas Kousik$^{2}$, and Ram Vasudevan$^{1,3}$
\thanks{This work is supported by the Ford Motor Company via the Ford-UM
Alliance under award N022977, National Science Foundation Career Award \#1751093, and by the Office of Naval Research under Award Number N00014-18-1-2575 }
\thanks{$^{1}$Robotics Institute, University of Michigan, Ann Arbor, MI $\langle$\texttt{jmichaux,pdholmes,jimzhang,ramv, cctom, baiyuew, shreyps, zhangtc}$\rangle$\texttt{@umich.edu}.}
\thanks{$^{2}$Mechanical Engineering, Georgia Institute of Technology, Atlanta, GA \texttt{shreyas.kousik@me.gatech.edu}.}
\thanks{$^{3}$Mechanical Engineering, University of Michigan, Ann Arbor, MI \texttt{ramv}\texttt{@umich.edu}.}
\thanks{$^{4}$Agility Robotics, Albany, OR \texttt{sid.dey}\texttt{@agilityrobotics.com}.}
}

\begin{document}

\maketitle
\thispagestyle{plain}
\pagestyle{plain} %% SHREYAS ADDED THIS SO WE HAVE PAGE NUMBERS

\begin{abstract}
Ensuring safe, real-time motion planning in arbitrary environments requires a robotic manipulator to avoid collisions, obey joint limits, and account for uncertainties in the mass and inertia of objects and the robot itself. 
This paper proposes Autonomous Robust Manipulation via Optimization with Uncertainty-aware Reachability (ARMOUR), a provably-safe, receding-horizon trajectory planner and tracking controller framework for robotic manipulators to address these challenges. 
ARMOUR first constructs a robust controller that tracks desired trajectories with bounded error despite uncertain dynamics. 
ARMOUR then uses a novel recursive Newton-Euler method to compute all inputs required to track any trajectory within a continuum of desired trajectories. 
Finally, ARMOUR over-approximates the swept volume of the manipulator; this enables one to formulate an optimization problem that can be solved in real-time to synthesize provably-safe motions. 
This paper compares ARMOUR to state of the art methods on a set of challenging manipulation examples in simulation and demonstrates its ability to ensure safety on real hardware in the presence of model uncertainty without sacrificing performance.
Project page: \url{https://roahmlab.github.io/armour/}.
\end{abstract}
\section{Introduction}

\begin{figure}[t]
    \centering
    \includegraphics[width=0.98\columnwidth]{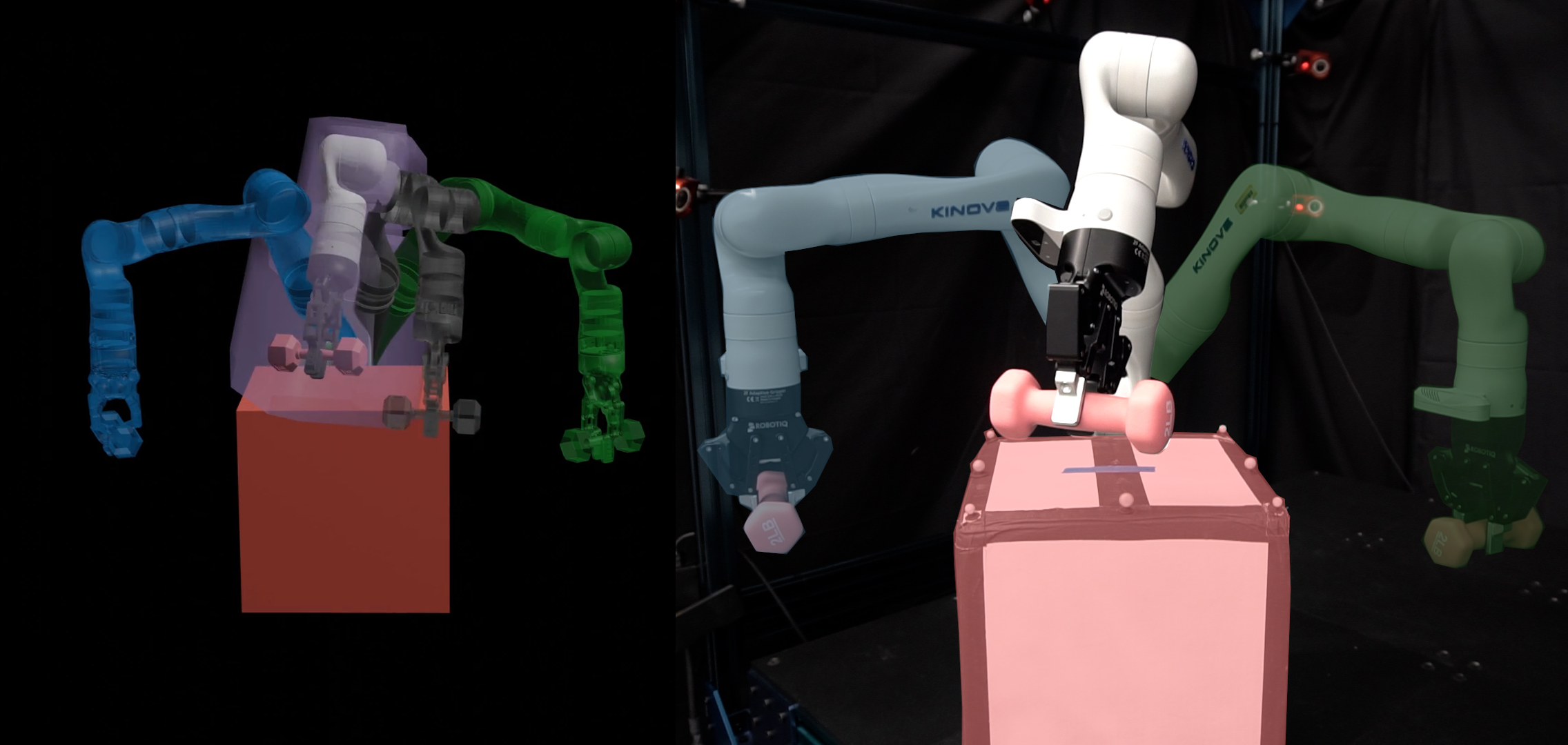}
    \caption{
    % An overview of the proposed method entitled \methodname.
    This paper considers the problem of safe motion planning with uncertain dynamics; for example, when manipulating a heavy dumbbell (pink) with uncertain mass in clutter (red obstacles).
    \methodname operates in a receding horizon way, moving from the start (right panel, blue) arm to the goal (right panel, green arm) by repeatedly generating new trajectory plans in real time.
    In each planning iteration, \methodname first computes a Forward Reachable Set (FRS) for a continuum of possible motion plans (left panel,purple volume), representing the swept volume of the arm under uncertain dynamics.
    Many of these motion plans may be in collision, so \methodname solves a constrained trajectory optimization problem to find a collision-free plan that makes progress towards an intermediate waypoint (left panel, black arm) and the global goal (right panel, green arm).
    }
    \label{fig:front_figure_method_overview}
    \vspace*{-0.2cm}
\end{figure}

Robotic manipulators have the potential to assist humans in a wide variety of collaborative settings, such as manufacturing, package delivery, and in-home care.
However, such settings are typically constrained and uncertain; nevertheless, the robot must operate in a safety-critical fashion. 
This makes it challenging to directly apply high-torque manipulators that ignore model uncertainty.
Instead, it is necessary to develop motion planning and control strategies that can operate safely by accounting for these types of uncertainty in real-time.
In this context, safety means avoiding collisions while obeying joint position, velocity, and torque limits.
To address the safety challenge, this paper proposes \textbf{Autonomous Robust Manipulation via Optimization with Uncertainty-aware Reachability (\methodname)}, a method for guaranteed-safe, real-time manipulator motion planning and control.
An overview of this method is given in Fig. \ref{fig:front_figure_method_overview}.
% This work significantly extends our previous work \cite{holmes2020armtd}, which ensured safety for a kinematic model of a manipulator given a fixed feedback controller.
The work enables safety for uncertain manipulator dynamics, which includes uncertain payloads, by proposing a combined planning and control framework.
% To proceed, we outline our research scope, state the contributions of this paper, and present a brief overview of the proposed method and the remainder of the paper.

% \shrey{figure suggestion:
% similar to ARMTD paper, show a time lapse of the arm moving a heavy dumbbell from one spot to another, and plot the FRS for one time instance somewhere in the middle.
% maybe show the FRS with and without dynamics uncertainty?}

\begin{figure*}[t]
    \centering
    \includegraphics[width=1.0\textwidth]{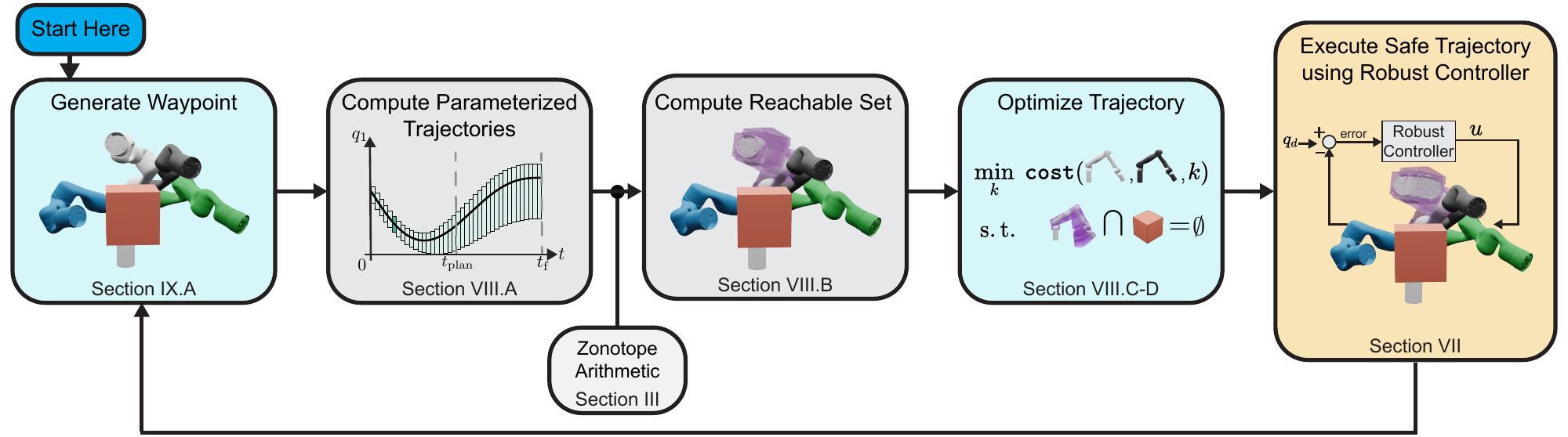}
    \caption{
    Overview of \methodname{}, a receding-horizon planning and control framework for robot manipulators with uncertain dynamics.
    First, a high-level planner generates waypoints between the robot's initial and goal states.
    Second, \methodname{} represents the time horizon as a finite set of polynomial zonotopes (Sec. \ref{subsubsec:time_horizon_PZs}) and computes a family of desired trajectories that are overapproximated by a finite set of polynomial zonotopes (Sec. \ref{subsubsec:trajectory_PZs}).
    Third, \methodname{} computes the forward occupancy of the robot over this entire set  (Sec. \ref{sec:reachset_construction}).
    Fourth, \methodname{} generates safety constraints (Sec. \ref{subsec:implementation_input_constraints}) and performs optimization over the family of parameterized desired trajectories of each joint (Sec.  \ref{sec:algorithm_theory:optimization}).
    Note that torque constraints and obstacle collision-avoidance constraints guarantee the safety of our method.
    Finally, to ensure that these trajectories can be safely followed, \methodname implements a robust passivity-based controller that uniformly bounds the tracking performance despite uncertainty in the robot's dynamics (Sec. \ref{sec:controller_design}).
    }
    \label{fig:paper_overview}
\end{figure*}

The focus of this paper is manipulator planning and control under uncertainty in a robot's dynamic model.
Typically, one relies on an accurate model of the robot that is assumed to be true.
However, it is challenging to construct a perfect model.
For example, the dynamics of a robotic arm depend on inertial properties that may not be known perfectly well (e.g., link masses, center of mass locations, and inertia matrices).
Similarly, a manipulator may not have access to an object's true inertial properties when grasping and transporting it.
This work seeks to robustly account for set-based uncertainty in manipulator and object inertial properties.

Safety is a critical property for deploying manipulators in real-world scenarios, as collisions with people or objects (i.e., obstacles) could cause grave harm.
Planning and control for collision avoidance is challenging because most obstacle-avoidance constraints are non-convex, and manipulators typically have nonlinear dynamics.
This challenge is compounded by the need to obey position, velocity, and torque constraints at each joint to avoid damaging the robot.
Furthermore, it is necessary to guarantee safety in continuous time, to ensure no safety violations between discrete time steps.
This work ensures safety by making continuous-time guarantees on collision avoidance, actuator limits, joint position limits, and joint velocity limits.
% Robots are equipped with sensors that have finite sensing horizons, and occlusions may block obstacles from view.
% Thus, a robot must update its map of the world as it moves and collects new information, and must react to these updates online.
% As a result, 
Because mobile robots must update their maps of the world as they move and collect new information, this work focuses on the development of a real-time, receding horizon planning and control method.
\textbf{Contribution:}
To the best of our knowledge, no motion planning and control framework exists that guarantees the continuous time safety of a manipulator robot with set-based uncertainty and operates in real-time.
Our framework guarantees continuous time collision avoidance and satisfaction of a robot's torque bounds and joint limits while explicitly accounting for tracking error caused by uncertainty in the robot's dynamics.
We make the following contributions:
First, we propose a novel robust passivity-based controller to safely account for uncertain robot dynamics by providing explicit tracking error guarantees.
Second, we construct a novel variation of the modified Recursive Newton-Euler Algorithm (RNEA) \cite{deluca2009rnea}, which we call polynomial zonotope RNEA (PZRNEA).
% This approach significantly reduces the conservativeness of ARMTD's numerical reachable set representation by using polynomial zonotopes \cite{kochdumper2020sparse}, and allows us to satisfy torque limit constraints during motion planning.
This approach uses polynomial zonotopes \cite{kochdumper2020sparse} to compute sets of possible inputs required to track trajectories, and allows us to satisfy torque limit constraints during motion planning.
Third, we use polynomial zonotopes to represent the forward occupancy of the robot, which is used to guarantee obstacle avoidance.
Each of these contributions is used to formulate the motion planning and control problem as an optimization problem that can be tractably solved in real-time with formal safety guarantees in a receding horizon fashion 
In particular, we demonstrate how to differentiate these polynomial zonotope-based representations to perform real-time optimization while providing safety guarantees. 
We demonstrate our method in simulation and hardware and compare it to state of the art alternatives.
These experiments confirm that our method enables safe manipulation of grasped objects with uncertain inertial parameters, while other methods may cause collisions or require motions that are not realizable.

\textbf{Relationship to Prior Work:}
\methodname builds upon prior work entitled Autonomous, Reachability-based Manipulator Trajectory Design (ARMTD) \cite{holmes2020armtd}. 
The prior work develops a planning algorithm for deterministic kinematic manipulator models. 
In contrast, this work is able to account for model uncertainty while using a dynamic model of the manipulator. 
As we show in this paper, \methodname significantly reduces the conservativeness of ARMTD's numerical reachable set.
Lastly, \methodname describes a strategy for control design and is able to account for input constraints. 
% Our code is open-source and available at \pat{add repo}.

% \subsection{Method Overview and Paper Organization}\label{subsec:intro_method_overview_and_paper_org}
\textbf{Method and Paper Overview:}
% We give a brief overview of our method and the structure of this document.
An overview of the different components of \methodname is illustrated in Fig. \ref{fig:paper_overview}.
The remainder of this paper is organized as follows:
First, we review related work in Sec. \ref{sec:related_work}.
Next, relevant notation and mathematical objects are introduced in Sec. \ref{sec:preliminaries}, and the robot's kinematics, dynamics and environment are discussed in Sec. \ref{sec:arm}.
Our motion planning and control hierarchy operates in a receding-horizon fashion, and is divided into a trajectory planner and a low-level robust controller.
The trajectory planner performs optimization over a family of parameterized desired trajectories of each joint (Sec. \ref{sec:trajectory}).
To ensure that these trajectories can be safely followed we implement a robust passivity-based controller that uniformly bounds the tracking performance despite uncertainty in the robot's dynamics (Sec. \ref{sec:controller_design}).
% Our framework plans in terms of parameterized desired trajectories of each joint (Sec. \ref{sec:trajectory}).
Online, the desired trajectories buffered by this uniform bound are assembled into reachable sets of the full arm to bound the robot's motion through the workspace and utilized within PZRNEA to bound the inputs required to track any desired trajectory (Sec. \ref{sec:algorithm_theory}).
Torque constraints and obstacle collision-avoidance constraints guarantee the safety of our method within an online trajectory optimization program, which generates a desired trajectory to be tracked by the robust controller.
% The cost function of the optimization program may be designed based on output from a high-level planner, such as reaching a desired waypoint.
We then demonstrate the efficacy of our method for safely manipulating uncertain grasped objects in simulation and in the real-world, and we compare it to other state of the art methods (Sec. \ref{sec:experiments}).

% We conclude and describe future work in Sec. \ref{sec:conclusion}.
\section{Background and Related Work}\label{sec:related_work}

This paper addresses the problem of real-time, safe manipulation planning and control that accounts for uncertain dynamics.
A common approach to solving similar problems is to use a global, sampling-based path planner to generate coarse, high-level motions, then a local planner to generate trajectories, and finally a controller to track the trajectories (e.g., \cite{yoshida2005humanoid,dai2014whole}, and our own prior work \cite{kousik2020bridging,holmes2020armtd}).
% We now review methods for both planning and control.

% \subsection{Planning}

It is challenging to generate motion plans to complete tasks while obeying constraints in real time due to the high-dimensional models used to describe manipulators.
Planning methods can be categorized depending upon how physics are represented: one can ignore physics (i.e., path or kinematic planning), or represent dynamics.
Methods can also be categorized by how plans are synthesized, using either sampling \cite{karaman2011sampling,kavraki1996probabilistic} or optimization \cite{zucker2013chomp,schulman2014motion}.
Note, further categorization can be applied depending upon how the robot and its environment are represented \cite{elbanhawi2014sampling,kingston2018sampling}.
Recall that the present work considers real-time, optimization-based motion planning for manipulators with uncertain dynamics.
To proceed, we review how kinematics, dynamics, and uncertainty have been considered in sampling- and optimization-based approaches; note that a thorough survey is available \cite{kingston2018sampling}.

A common approach to enable fast performance is to plan a path or a kinematic trajectory, then rely on an underlying feedback controller to compensate for the robot's true dynamics.
Our previous work \cite{holmes2020armtd} employed this strategy, as do a variety of optimization-based planners \cite{zucker2013chomp,schulman2014motion,kalakrishnan2011stomp}.
This approach is also used in sampling-based motion planning for complex robots such as manipulators or humanoids, often by directly modeling the configuration manifold \cite{jaillet2012_atlasrrt_kinematic_constraints, porta2012randomized_plan_on_manifolds}, though these approaches are not typically fast enough for real time replanning.
To increase computational efficiency, one can instead precompute some constraint manifolds \cite{csucan2012plan_with_constraints}.
With these methods, computational speed comes at the expense of not modeling the dynamics of the robot.

% On the other hand, one can directly model a robot's dynamics in planning.
An alternate approach is to incorporate a model of the robot's dynamics during planning.
Notably, researchers have extended the well-known Rapidly-exploring Random Trees (RRT) algorithm \cite{lavalle1997motion} to incorporate torque and friction constraints in discrete time by projection to constraint manifolds \cite{berenson2009manipulation_dumbbell}, though this approach cannot plan in real time.
More recently, it has been shown how to verify robot motion near humans by rapidly computing capsule-shaped safety zones for manipulator dynamics \cite{beckert2017online,scalera2021optimal,scalera2022online}; however, these methods do not account for uncertainty in the dynamics.
These previous methods focus on collision avoidance for manipulators; however, we note that there is extensive work on humanoid motion planning with dynamics, typically by planning a sequence of static, dynamically-feasible poses \cite{kuffner2002dynamically_stable_humanoid,kanehiro2008integrating_dyn_plan_humanoid,han2020can}.
These latter methods do not operate in real time due to the high-dimensional nature of humanoid robots.
Altogether, it remains a challenge to rapidly plan collision-free manipulator motion while incorporating uncertain dynamics.
% \shrey{want to say:
%     We consider two axes: sampling/optimization methods, and path/kinematic/dynamics.
%     A common approach to get fast performance is to consider path planning or kinematic planning, and lean on feedback controller to account for uncertainty/dynamics.
%     Or, one can directly attempt to plan for the dynamics of the robot.
%     One can also attempt to account for uncertainty, but this again adds more computational expense and requires approximations.}
% \subsection{Control}

As mentioned above, to achieve fast performance, most approaches plan kinematic paths or trajectories, then rely on a lower-level controller to obey dynamic constraints.
We briefly note that some approaches instead merge kinematic planning and control \cite{ratliff2018riemannian,bylard2021_pbds} by generalizing the notion of potential fields.
% This concept can also extend to multi-manipulator systems \cite{sina2016coordinated_multiarm_uncertainty} and humanoids \cite{koptev2021real}.
While these methods can guarantee kinematic collision avoidance in discrete time \cite{bylard2021_pbds}, they do not incorporate dynamics.
% To proceed, we review classical adaptive methods, and more recent interval arithmetic methods, for robust manipulator control.
There is a long history of constructing control strategies to account for manipulator dynamics \cite{abdallah1991survey,sage1999robust}.
The core challenge is to develop a controller that guarantees convergence to a given desired trajectory despite disturbance and uncertain, nonlinear dynamics.
A further challenge is to ensure these controllers are continuous, to avoid chattering or exciting high-frequency dynamics that are not typically modeled \cite{slotine1983tracking,slotine1985robust}.
To handle parametric uncertainty, a variety of adaptive and sliding-mode control approaches have been proposed that prove both convergence and bounds on trajectory tracking error by simultaneously controlling a robot and estimating its uncertain parameters \cite{islam2010robust,neila2011adaptive,baek2016new,zhu2020estimation}.

One of the key challenges with such techniques is to estimate bounds on the nonlinear perturbations to the controlled system \cite{neila2011adaptive}.
A recent alternative has been to directly compute the reachable set of states for a manipulator with parameter uncertainty via interval arithmetic for verification  \cite{giusti2017efficient,wagner2018interval,giusti2021interval} while building off of existing techniques for linear systems \cite{malan1997robust,smagina2002using} and nonlinear systems \cite{jaulin2002nonlinear,bravo2005computation}.
% The present work develops a novel controller with tighter bounds based on these previous interval arithmetic methods (as is discussed in App. \ref{app:input_bound_comparison}). 
The present work develops a novel controller that improves on previous interval arithmetic-based methods in three ways.
First, our controller generates smaller inputs that achieve the same tracking error performance.
Second, this allows us to build tighter tracking error bounds in practice (see \href{https://github.com/roahmlab/armour/blob/main/assets/TRO_Armour_Appendix_F.pdf}{online appendix}).
Third, we demonstrate that our controller is not only useful for verification, but also planning.
\section{Preliminaries}
\label{sec:preliminaries}

To describe the swept volume of a manipulator subject to uncertain dynamics, we leverage two numerical set representations: intervals and polynomial zonotopes.
This section describes our notation conventions and each set representation.
% and its significance in our overall framework.

\subsection{Notation}
The $n$-dimensional real numbers are $\R^n$, natural numbers $\N$, the unit circle is $\mathbb{S}^1$, and the set of $3 \times 3$ orthonormal matrices is $\SO(3)$.
% Upper case letters represent sets (or matrices), while lowercase letters represent vectors or scalars; we indicate minor exceptions when necessary.
Subscripts may index elements of a vector or a set, or provide additional context.
For a matrix $A$, we use $A\trans$ to denote its transpose.
We use the product notation $\prod_{i=1}^{m} A_i = A_1 A_2 \cdots A_m$ to denote successive matrix multiplications, where $A_i \in \R^{n \times n}$. 
Vector concatenation is denoted $(x_1, \ldots, x_n)$ unless the shape is critical to understanding, in which case we write $[x_1\trans, \ldots, x_n\trans]\trans$.

Let $U$, $V \subset \R^n$.
For a point $u \in U$, $\{ u \} \subset U$ is the set containing only $u$.
The \defemph{power set} of $U$ is $\pow{U}$.
The \defemph{Minkowski Sum} is $U \oplus V  = \{ u + v \, \mid \, u \in U, v \in V \}$; 
the \defemph{Minkowski Difference} is $U \ominus V = U \oplus (-V)$.
For vectors $a, b \in \R^3$, we write the cross product as $a^\times b$, where
\begin{equation}
    \label{eq:cross_product_matrix}
    a^\times= \begin{bmatrix}
    0 & -a_3 & a_2 \\
    a_3 & 0 & -a_1 \\
    -a_2 & a_1 & 0
    \end{bmatrix}.
\end{equation}
If $n = 3$, the \defemph{set-based cross product} is defined as $U \otimes V = \{ u \times v \, \mid \, u \in U, v \in V \}$.
If $\{U_i \subset \R^n\}_{i=1}^{m}$ then let $\bigtimes_{i=1}^n U_i$ denote the Cartesian product of the $U_i$'s.
Let $W \subset \R^{n \times n}$ be a set of square matrices.
Then, \defemph{set-based multiplication} is defined as $WV = \{ Av, \, \mid \, A \in W, v \in V \}$.
% Minksowski sum: $A \oplus B = \{ x = a + b \, \mid \, a \in A, b \in B \}$.
% Set multiplication: $A \odot B = \{ x = ab, \, \mid \, a \in A, b \in B \}$ (well-defined if $a$ and $b$ are appropriately sized).
% Set-based cross product: $A \subset \R^{3}, B \subset \R^{3}$, $A \otimes B = \{ x = a \times b \, \mid \, a \in A, b \in B \}$.
% \shrey{Need to add notation for identity matrix and matrices of zeros or ones?}
% \fix{added:}
Let $\zeros$ (resp. $\ones$) denote a matrix of zeros (resp. ones) of appropriate size, and let $\eye_n$ be the $n \times n$ identity matrix.

\vspace*{-0.22cm}
\subsection{Intervals}

% \shrey{Need to add a paragraph before the intervals/zonotopes/PZ subsections explaining all the different set representations that will be discussed, and why we discuss/use each of them.
% Maybe each set representation should be its own subsubsection, or else (if you guys want to keep this format), the set representations should just get their own entire section with an intro paragraph.
% The goal of the intro paragraph about set representations is that a reader can just read it and then skip all the mathematical details and still have an idea of what is going on.
% For example, ``In this work, we seek to conservatively represent the swept volume of a manipulator subject to uncertain dynamics.
% To achieve this, we leverage several numerical set representations of increasing complexity: intervals, zonotopes, and polynomial zonotopes.
% Next we describe each representation and its significance in our overall framework.''}
% \shrey{It is important in the beginning of each subsection to very briefly explain why intervals, zonotopes, operations like Minkowski sums, matrix/zonotope products, etc. are useful before diving into definitions.
% That is, we need to preempt the question, ``why should the reader care to memorize all this stuff right now?''}
% % \fix{Tried to add a little more exposition}
% Various set representations are used extensively throughout this paper.
We describe uncertain manipulator parameters as intervals (see \cite{hickey2001interval,althoff2010reachability} for an overview of interval arithmetic).
An \defemph{n-dimensional interval} is a set
\begin{align}
    \interval{x} &= \{ y \in \R^n \, \mid \, \lb{x}_i \leq y_i \leq \ub{x}_i, \Hquad \forall i = 1, \dots, n\}.
\end{align}
% Note we have abused notation and used a lowercase letter to denote a set here; however, the $\interval{\cdot}$ notation hopefully makes clear that this indeed means a set. 
When the bounds are important, we denote an interval $\interval{x}$ by $\interval{\lb{x}, \ub{x}}$, where $\lb{x}$ and $\ub{x}$ are the infimum and supremum, respectively, of $\interval{x}$.
Let $\setop{inf}(\iv{x}) := \lb{x}$ and $\setop{sup} \setop{sup}(\iv{x}) := \ub{x}$.
% return the infimum and supremum of an interval:
% \begin{align}
%     \setop{inf}(\iv{x}) = \lb{x} \quad \text{and} \quad \setop{sup}(\iv{x}) = \ub{x}.
% \end{align}
Let $\mathbb{I}\mathbb{R}^{n}$ be the set of all real-valued $n$-dimensional interval vectors. 
The Minkowski sum and difference of $\interval{x}$ and $\interval{y}$ are
\begin{gather}
    \interval{x} \oplus \interval{y} = \interval{\lb{x} + \lb{y}, \ub{x} + \ub{y}}, \\
    \interval{x} \ominus \interval{y} = \interval{\lb{x} - \ub{y}, \ub{x} - \lb{y}}.
\end{gather}
The product of $\interval{x}$ and $\interval{y}$ is
\begin{equation}
    \interval{x} \interval{y} =
        \big[\min\big(\lb{x} \lb{y}, \lb{x} \ub{y},\ub{x}\lb{y}, \ub{x} \ub{y}\big),
        \max\big(\lb{x} \lb{y}, \lb{x} \ub{y},\ub{x}\lb{y}, \ub{x} \ub{y}\big) \big].
\end{equation}
The $i$\ts{th} and $j$\ts{th} entry of the product of an 
% Given an 
% scalar interval $\interval{a}$ or 
interval matrix $\interval{Y}$ multplied by an interval matrix $\interval{X}$ is
\begin{gather}
    % (\interval{a} \interval{X})_{ij} = \interval{a} \interval{X}_{ij}, \\
    (\interval{X} \interval{Y})_{ij} = \bigoplus_{k=1}^n (\interval{X}_{ik} \interval{Y}_{kj}),
\end{gather}
where $n$ is the number of columns of $\interval{X}$ and number of rows of $\interval{Y}$.
Given interval vectors $\interval{x},\interval{y} \subset \R^3$, their cross product is
\begin{align}
    \interval{x} \otimes \interval{y} = \interval{x}^{\times} \interval{y},
\end{align}
where $\interval{x}^{\times}$ is the skew-symmetric matrix representation of $\interval{x}$ as in \eqref{eq:cross_product_matrix} (i.e., a matrix with interval entries).

\begin{figure}[t]
    \centering
    \includegraphics[width=0.98\columnwidth]{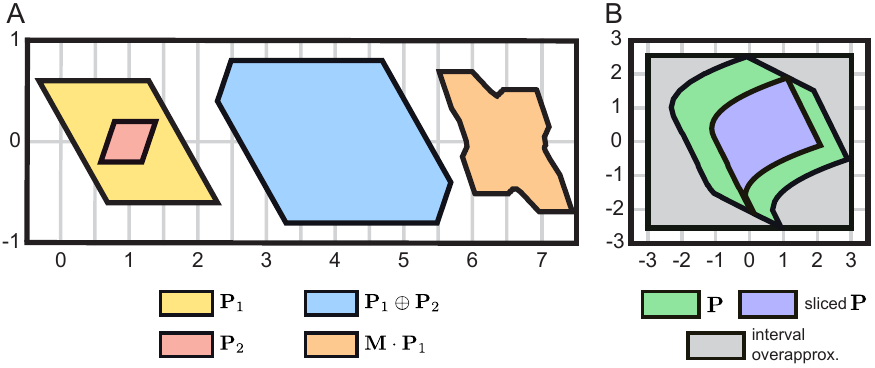}
    \caption{Illustration of polynomial zonotope arithmetic. (A) The middle panel shows the minkowski sum (blue) of polynomial zonotopes $\pz{P}_1$ (gold) and $\pz{P}_2$ (red). The right panel shows the multiplication (orange) of $\pz{P}_1$ by a matrix $M$. (B) A polynomial zonotope $\pz{P}$, shown in green, is sliced to produce the polynomial zonotope shown in purple. The grey rectangle represents an interval overapproximation of $\pz{P}$. }
    \label{fig:pz_figure}
    \vspace*{-0.2cm}
\end{figure}

\subsection{Polynomial Zonotopes}\label{subsec:polyzono}
% \pat{not sure i'm loving how close $a_i$ and $\pzei$ look... maybe i'll change $a$'s to be $c$'s}
We present an overview of the necessary definitions and operations of polynomial zonotopes (PZs) in this subsection.
A thorough introduction is available \cite{kochdumper2020sparse} and illustration of the operations we perform on PZs can be found in Fig. \ref{fig:pz_figure}.
Given \defemph{generators} $\pzgi \in \R^n$ and \defemph{exponents} $\pzei \in \N\pzn$ for $i \in \{0 ,\ldots, \pzn\}$, a \defemph{polynomial zonotope} is a set:
\begin{align}\label{eq:pz_definition}
    \hspace*{-0.2cm} \pz{P} = \PZ{\pzgi, \pzei, \pzv} = \big\{
                z \in \R^n \, \mid \,
                z = \sum_{i=0}^{\pzn} \pzgi \pzv ^{\pzei}, \, \pzv \in [-1, 1]^\pzn
            \big\}.
    \end{align}
In this paper, we set the exponent associated with the zeroth generator to be zero, i.e., $\alpha_0 = [0, 0, \ldots, 0]$.
We refer to $\pzv ^{\pzei}$ as a \defemph{monomial}, $\pzv \in [-1,1]^\pzn$ as \defemph{indeterminates}, and $\pzg_0$ as the \defemph{center}.
    % A polynomial zonotope $\pz{P} \subset \R^n$ is given by its generators $\pzgi \in \R^n$ (of which there are $\pzn$), exponents $\pzei \in \N^\pzn$, and indeterminates $\pzv \in [-1,1]^\pzn$ as
    % \begin{align}\label{eq:pz_definition}
    %     \pz{P} = \PZ{\pzgi, \pzei, \pzv} = \left\{
    %             z \in \R^n \, \mid \,
    %             z = \sum_{i=0}^{\pzn} \pzgi \pzv ^{\pzei}, \, \pzv \in [-1, 1]^\pzn
    %         \right\}.
    % \end{align}
    % We refer to $\pzv ^{\pzei}$ as a \defemph{monomial}.
    A \defemph{term} $\pzgi\pzv ^{\pzei}$ is produced by multiplying a monomial by the associated generator $\pzgi$.
    % When we need to emphasize the generators and exponents, we write $\pz{P} = \PZ{\pzgi, \pzei, \pzv}$.
% \end{defn}
% A polynomial zonotope $\pz{P} \subset \R^n$ is given by its generators, exponents, and indeterminates.
% \shrey{I think this bracket notation is a bit more confusing than some sort of functional notation, such as $P = \PZ{\pzgi,\pzei,\pzv}$, which would avoid abusing notation and separate the object $P$ from its parameters (the arguments to $\PZ{$).}
% When we need to emphasize the generators and exponents, we write $\pz{P} = \PZ{\pzgi, \pzei, \pzv}$.
Throughout this document, we exclusively use bold symbols to denote polynomial zonotopes.

Next, we introduce several useful operations on polynomial zonotopes: interval conversions, set addition and multiplication, slicing, computing bounds, and set-valued function evaluation.
% The usual order of operations for addition, multiplication and exponentiation apply for polynomial zonotope operations as well.
Table \ref{tab:poly_zono_operations} summarizes these operations.
Importantly, these operations can either be computed exactly or in an overapproximative fashion using polynomial zonotopes.

\begin{table}[b]
    \caption{Summary of polynomial zonotope operations.}
    \centering
    \begin{tabular}{l|c}
        \multicolumn{1}{c|}{Operation} & Computation \\
        \hline
        $\iv{z} \to \pz{z}$ (Interval Conversion) \eqref{eq:int_to_pz} & Exact \\
        % $Z \to \pz{Z}$ (Zonotope Conversion) \eqref{eq:first_pz_definition} & Exact \\
        $\pz{P}_1 \oplus \pz{P}_2$ (Minkowski Sum) \eqref{eq:pz_minkowski_sum} & Exact \\
        $\pz{P}_1\pz{P}_2$ (Set Multiplication) \eqref{eq:pz_multiplication} & Exact \\
        $\pz{P}_1^m$ (Exponentiation) & Exact \\
        $\pz{P}_1 \otimes \pz{P}_2$ (Set Cross Product) \eqref{eq:pz_cross_product_matrix} & Exact \\
        % Cartesian Product & x \\
        $\setop{slice}(\pz{P}, \pzv_j, \sigma)$ \eqref{eq:pz_slice} & Exact \\
        $\setop{sup}(\pz{P})$ \eqref{eq:pz_sup} and $\setop{inf}(\pz{P})$ \eqref{eq:pz_inf} & Overapproximative \\
        % $\setop{reduce}(\pz{P}, n_h)$ \eqref{eq:pz_reduce} & Overapproximative \\
        % $\setop{free}$ & o \\
        $f(\pz{P}_1) \subseteq \pz{P}_2$ (Taylor expansion) \eqref{eq:pz_taylor_fin} & Overapproximative
    \end{tabular}
    \label{tab:poly_zono_operations}
    \vspace*{-1cm}
\end{table}

\subsubsection{Intervals Conversion}
Intervals can also be written as polynomial zonotopes.
For example, let $[z] = [\underline{z}, \overline{z}] \subset \mathbb{R}^n$, then one can convert $[z]$ to a polynomial zonotope $\pz{z}$ using
\begin{equation}
    \label{eq:int_to_pz}
    \pz{z} = \frac{\overline{z} + \underline{z}}{2} + \sum_{i = 1}^{n}\frac{\overline{z}_i - \underline{z}_i}{2}x_i,
\end{equation}
where $x \in [-1, 1]^n$ is the indeterminate vector.

\subsubsection{Set Addition and Multiplication}
The Minkowski Sum of two polynomial zonotopes $\pz{P}_1 \subset \R^n = \PZ{ \pzgi, \pzei, \pzv } $ and $\pz{P}_2 \subset \R^n = \PZ{ h_j, \beta_j, y }$ follows from polynomial addition:
\begin{align}
    \pz{P}_1 \oplus \pz{P}_2
    % &=  \left\{     z \in \R^n \, \mid \, z = p_1 + p_2, p_1 \in \pz{P}_1, p_2 \in \pz{P}_2 \right\} \\
   &=
        \left\{
            z \in \R^n \mid z = \sum_{i=0}^{\pzn} \pzgi \pzv ^{\pzei} + \sum_{j=0}^{n_h} h_j y^{\beta_j}
        \right\}. \label{eq:pz_minkowski_sum}
\end{align}
Similarly, we may write the matrix product of two polynomial zonotopes $\pz{P}_1$ and $\pz{P}_2$ when the sizes are compatible.
% (i.e., elements in $\pz{P}_1$ have the same number of columns as elements of $\pz{P}_2$ have rows).
% \shrey{What does ``compatible'' mean?
% I mean, it's clear from reading the equations below that it has to do with objects being the right size for matrix products, but it's only clear to me because I already know what is going on.}
% \fix{\checkmark}
Letting $\pz{P}_1 \subset \R^{n \times m}$ and $\pz{P}_2 \subset \R^{m \times k}$, we obtain $\pz{P}_1 \pz{P}_2 \subset \R^{n \times k}$:
\begin{align}
    \pz{P}_1 \pz{P}_2 
    % &=     \left\{      z \in \R^{n \times k} \, \mid \, z = p_1p_2, p_1 \in \pz{P}_1, p_2 \in \pz{P}_2 \right\} \\
    &= \left\{
            z \in \R^{n \times k} \, \mid \, z = \sum_{i=0}^{\pzn} \pzgi(\sum_{j=0}^{q} h_j y^{\beta_j}) \pzv ^{\pzei}
        \right\}. \label{eq:pz_multiplication}
\end{align}
When $\pz{P}_1 \subset \R^{n \times n}$ is square, exponentiation $\pz{P}_1^m$ may be performed by multiplying $\pz{P}_1$ by itself $m$ times.
Furthermore, if $\pz{P}_1 \subset \R^{3}$ and $\pz{P}_2 \subset \R^{3}$, we implement a set-based cross product as matrix multiplication.
% \shrey{Avoid ``may'' and ``will,'' which just take up space and make things sound wishy-washy.}
% \fix{\checkmark}
We create $\pz{P}_1^\times \subset \R^{3 \times 3}$ as
\begin{align}
    \pz{P}_1^\times = \left\{ A \in \R^{3 \times 3} \, \mid \, A = \sum_{i = 0}^{\pzn}
        \left[\begin{smallmatrix}
            0 & -\pzg_{i,3} & \pzg_{i,2} \\ \pzg_{i,3} & 0 & -\pzg_{i,1} \\ -\pzg_{i,2} & \pzg_{i,1} & 0
        \end{smallmatrix}\right] \pzv ^{\pzei}
    \right\} \label{eq:pz_cross_product_matrix}
\end{align}
where $g_{i, j}$ refers to the $j$\ts{th} element of $g_i$.
Then, the set-based cross product $\pz{P}_1 \otimes \pz{P}_2 = \pz{P}_1^{\times} \pz{P}_2$ is well-defined. 

\subsubsection{Slicing}
One can obtain subsets of polynomial zonotopes by plugging in values of known indeterminates.
For instance, if a polynomial zonotope $\pz{P}$ represented a set of possible positions of a robot arm operating near an obstacle.
It may be beneficial to know whether a particular choice of $\pz{P}$'s indeterminates yields a subset of positions that could collide with the obstacle.
To this end, ``slicing'' a polynomial zonotope $\pz{P} = \PZ{ \pzgi, \pzei, \pzv }$  corresponds to evaluating an element of the indeterminate $\pzv$.
Given the $j$\ts{th} indeterminate $\pzv_j$ and a value $\sigma \in [-1, 1]$, slicing yields a subset of $\pz{P}$ by plugging $\sigma$ into the specified element $\pzv_j$:
\begin{equation}
    \label{eq:pz_slice}
    \hspace*{-0.25cm} \setop{slice}(\pz{P}, \pzv_j, \sigma) \subset \pz{P} =
        \big\{
            z \in \pz{P} \, \mid \, z = \sum_{i=0}^{\pzn} \pzgi \pzv ^{\pzei}, \, \pzv_j = \sigma
        \big\}.
\end{equation}
% Given values $\sigma \in [-1, 1]^s$ and a corresponding set of indices $\mathcal{S}$ (where $|\mathcal{S}| = s$), slicing yields a subset of a polynomial zonotope by plugging in the values $\sigma$ for corresponding elements of $x$:

\subsubsection{Bounding a PZ}
Our algorithm requires a means to bound the elements of a polynomial zonotope.
% It is possible to efficiently generate these upper and lower bounds on the values of a polynomial zonotope through overapproximation.
We define the $\setop{sup}$ and $\setop{inf}$ operations which return these upper and lower bounds, respectively, by taking the absolute values of generators.
For $\pz{P} \subseteq \R^n$ with center $g_0$ and generators $g_i$,
\begin{align}
    \setop{sup}(\pz{P}) = g_0 + \sum_{i=1}^{\pzn} \abs{\pzgi}, \label{eq:pz_sup}\\
    \setop{inf}(\pz{P}) = g_0 - \sum_{i=1}^{\pzn} \abs{\pzgi}. \label{eq:pz_inf}
\end{align}
Note that for any $z \in \pz{P}$,  $\setop{sup}(\pz{P}) \geq z$ and $\setop{inf}(\pz{P}) \leq z$, where the inequalities are taken element-wise.
These bounds may not be tight, 
% because possible dependencies between indeterminates are not accounted for,
but they are quick to compute.

\subsubsection{Set-Valued Function Evaluation}
% Though we have defined several basic operations like addition and multiplication above, it may be desirable to use polynomial zonotopes as inputs to more complicated functions.
One can overapproximate any analytic function evaluated on a polynomial zonotope using a Taylor expansion, which itself can be represented as a polynomial zonotope \cite[Sec 4.1]{althoff2013reachability},\cite[Prop. 13]{kochdumper2020sparse}.
Consider an analytic function $f: \R \to \R$ and $\pz{P}_1 = \PZ{ \pzgi, \pzei, \pzv }$, with each $\pzgi \in \R$, then $f(\pz{P}_1) = \{ y \in \R \; | \; y = f(z), z \in \pz{P}_1 \}$.
We generate $\pz{P}_2$ such that $f(\pz{P}_1) \subseteq \pz{P}_2$ using a Taylor expansion of degree $d \in \N$, where the error incurred from the finite approximation is overapproximated using a Lagrange remainder.
The method follows the Taylor expansion found in the reachability algorithm in \cite{kochdumper2020sparse}, which builds on previous work on conservative polynomialization found in \cite{althoff2013reachability}.
Recall that the Taylor expansion about a point $c \in \R$ is
\begin{equation}
    f(z) = \sum_{n = 0}^{\infty} \frac{f^{(n)}(c)}{n!} (z - c)^n,
\end{equation}
where $f^{(n)}$ is the $n$\ts{th} derivative of $f$.
The error incurred by a finite Taylor expansion can be bounded using the Lagrange remainder $r$ \cite[7.7]{apostol1991calculus}:
\begin{equation}
    |f(z) - \sum_{n=0}^{d} \frac{f^{(n)}(c)}{n!} (z - c)^n | \leq r,
\end{equation}
where
\begin{align}
    r =  \underset{\delta \in [c,z]} {\max} \frac{( | f^{d+1}(\delta) | ) |z - c |^{d+1}}{(d+1)!}.
\end{align}

For a polynomial zonotope, the infinite dimensional Taylor expansion is given by
\begin{equation}
\label{eq:pz_taylor_inf}
    f(\pz{P}_1) =
    \sum_{n=0}^{\infty}
    \frac{f^{(n)}(c)}{n!} 
    (\pz{P}_1 - c)^n.
\end{equation}
In practice, only a finite Taylor expansion of degree $d \in \N$ can be computed.
Letting $c = \pzg_0$ (i.e., the center of $\pz{P}_1$), and noting that $(z - c) = \sum_{i=1}^{\pzn}\pzgi \pzv ^{\pzei}$ for $z \in \pz{P}_1$, we write
\begin{equation}
\label{eq:pz_taylor_fin}
    \pz{P}_2 \coloneqq
        \left\{ z \in \R \, | \, z \in
            \sum_{n=0}^{d} \left (
            \frac{f^{(n)}(\pzg_0)}{n!} 
            (\sum_{i=1}^{\pzn}\pzgi \pzv ^{\pzei})^n 
            \right )
            \oplus [r]
        \right\},
\end{equation}
% \pat{bleh doesn't quite make sense}
and the Lagrange remainder $[r]$ is computed using interval arithmetic as
\begin{align}
    [r] &= \frac{[f^{(d+1)}([\pz{P}_1])] [(\pz{P}_1 - c)^{d+1}]}{(d+1)!}
\end{align}
where $[(\pz{P}_1 - c)^{d+1}] = [\setop{inf}((\pz{P}_1 - c)^{d+1}), \setop{sup}((\pz{P}_1 - c)^{d+1})]$ is an overapproximation of $(\pz{P}_1 - c)^{d+1}$.
% that can be computed using the $\setop{sup}$ \eqref{eq:pz_sup} and $\setop{inf}$ \eqref{eq:pz_inf} operations 
$\pz{P}_2$ can be expressed as a polynomial zonotope because all terms in the summation are polynomials of $\pzv$, and the interval $[r]$ can be expressed as a polynomial zonotope as in \eqref{eq:int_to_pz}.
We denote the polynomial zonotope overapproximation of a function evaluated on a zonotope using bold symbols (i.e., $\pz{f}(\pz{P}_1)$ is the polynomial zonotope overapproximation of $f$ applied to $\pz{P}$).

\section{Arm and Environment}
\label{sec:arm}
% The goal of this work is to plan guaranteed-safe trajectories for serial manipulators with uncertain inertial properties in real time.
This section introduces the robot arm's kinematics and dynamics, and defines the environment and obstacles.

\subsection{Robotic Manipulator Model}
Given an $\nq$-dimensional serial robotic manipulator with configuration space $Q$ and a compact time interval $T \subset \R$ we define a trajectory for the configuration as $q: T \to Q \subset \R^{\nq}$. 
The trajectory's velocity is $\dot{q}: T \to \R^{\nq}$.
Let $\Nq = \{1,\ldots,\nq\}$.
We assume the following about the robot model:
\begin{assum}[Workspace and Configuration Space]
The robot operates in a three dimensional workspace, $W \subset \R^3$.
The robot is composed of only revolute joints, where the $j$\ts{th} joint actuates the robot's $j$\ts{th} link.
The robot's $j$\ts{th} joint has position and velocity limits given by $\qj \in [\qlim^-, \qlim^+]$ and $\qdj \in [\dqlim^-, \dqlim^+]$ for all $t \in T$, respectively.
During online operation, the robot has encoders that allow it to measure its joint positions and velocities.
The robot is fully actuated, where the robot's input $u: T \to \R^{\nq}$ has limits,  i.e., $u_j(t) \in [\ulim^-, \ulim^+]$ for all $t \in T$ and $j \in \Nq$. 
\end{assum}
\noindent We make the one-joint-per-link assumption with no loss of generality because joints with multiple degrees of freedom (e.g., spherical joints) may be represented in this framework using links with zero length.
In addition, this work can be extended to robots with prismatic joints by extending the RNEA algorithms presented in Sec. \ref{sec:algorithm_theory}.

\subsubsection{Arm Kinematics}
% Next, we introduce the robot's kinematics.
Suppose there exists a fixed inertial reference frame, which we call the \defemph{world} frame.
In addition suppose there exists a \defemph{base} frame, which we denote the $0$\ts{th} frame, that indicates the origin of the robot's kinematic chain.
We assume that the $j$\ts{th} reference frame $\{\hat{x}_j, \hat{y}_j, \hat{z}_j\}$ is attached to the robot's $j$\ts{th} revolute joint, and that $\hat{z}_j = [0, 0, 1]\trans$ corresponds to the $j$\ts{th} joint's axis of rotation.
For a configuration at a particular time, $\q$, the position and orientation of frame $j$ with respect to frame $j-1$ can be expressed using the homogeneous transformation matrix $\homtrans_{j}^{j-1}(\qj)$ \cite[Ch. 2]{spong2005textbook}:
\begin{equation}
\label{eq:homogeneous_transform}
    \homtrans_{j}^{j-1}(\qj) = 
    \begin{bmatrix} R\jssmu (\qj) & p\jssmu \\
    \zeros  & 1 \\
    \end{bmatrix},
\end{equation}
where $R\jssmu (\qj)$ is a configuration-dependent rotation matrix and $p\jssmu$ is the fixed translation vector from frame $j-1$ to frame $j$.
% Note that we use the non-standard notation $\homtrans_{j}^{j-1}$ to avoid confusion with other symbols denoting time.
% With these definitions, we can express the forward kinematics of the robot.
Let $\FK_j: Q \to \R^{4 \times 4}$ map the robot's configuration to the position and orientation of the $j$\ts{th} joint in the world frame:
\begin{equation}\label{eq:fk_j}
    \FK_j(\q) = 
    \prod_{l=1}^{j} \homtrans_{l}^{l-1}(q_l(t)) = 
    \begin{bmatrix} R_j(\q) & p_j(\q) \\
    \zeros  & 1 \\
    \end{bmatrix},
\end{equation}
% \pat{per Zac: should the $\qj$ in the product be $q_l$ instead? also, should the matrix have a $R_{j}$ and $p_{j}$} \jon{fixed}
where 
\begin{align}
    R_j(\q) &\coloneqq R_j^{0}(\q) = \prod_{l=1}^{j}R_{l}^{l-1}(\ql), \\
    p_j(\q) &\coloneqq \sum_{l=1}^{j} R_{l}(\q) p_{l}^{l-1}.
\end{align}

\subsubsection{Arm Occupancy}
% An arm's kinematics can be used to describe the volume occupied by the arm in the workspace. 
Let $L_j \subset \R^3$ denote the volume occupied by the $j$\ts{th} link with respect to the $j$\ts{th} reference frame. 
The forward occupancy of link $j$ is given by $\FO_j: Q \to \pow{W}$ defined as:
\begin{align}\label{eq:forward_occupancy_j}
     \FO_j(\q) &= p_j(\q) \oplus R_j(\q) L_j,
\end{align}
where the first term gives the position of joint $j$ and the second gives the rotated volume of link $j$.
The volume occupied by the arm in the workspace is given by:
\begin{align}\label{eq:forward_occupancy}
    \FO(\q) = \bigcup_{j = 1}^{\nq} \FO_j(\q). 
\end{align}

\subsubsection{Arm Dynamics}
The robot is composed of $n_q$ rigid links with inertial parameters:
\begin{equation}\label{eq:true_intertial_parameters}
    \Delta = (m_1, \ldots, m_{\nq}, c_{x,1}, \ldots, c_{z,{\nq}}, I_{xx,1}, \ldots, I_{zz,{\nq}})
\end{equation}
where $m_j$, $c_j = (c_{x,j}, c_{y,j}, c_{z,j})$, and $(I_{xx,j}, \ldots, I_{zz,j})$  represent the mass, center of mass, and inertia tensor of the $j$\ts{th} link, respectively. 
The dynamics are represented by the standard manipulator equations \cite{spong2005textbook}:
\begin{equation}\label{eq:manipulator_equation}
    % \roboteqn    
    \bM(\q, \Delta) \qdd +
    \bC(\q, \qd, \Delta) \qd +
    \bG(\q, \Delta) = u(t)
\end{equation}
where $\Mq \in \R^{\nq \times \nq}$ is the positive definite inertia matrix, $\Cqd$ is the Coriolis matrix, $\Gqd$ is the gravity vector, and $u(t)$ is the input torque all at time $t$.
For simplicity, we do not explicitly model friction, but it can be incorporated in $\bG(\q, \Delta)$ \cite{giusti2017efficient,giusti2021interval}.
% We make the following assumptions regarding the robot's dynamics:
We assume the following about the inertial parameters:
\begin{assum}[Inertial Parameter Bounds]
\label{assum:arm_model}
The model structure (i.e., number of joints, sizes of links, etc.) of the robot is known, but its true inertial parameters $\trueparams$ are unknown. 
The uncertainty in each inertial parameter is given by
\begin{equation}
   \hspace*{-0.25cm} \intparams = \left(\iv{m_1}, \ldots, \iv{m_{\nq}}, \iv{c_{x,1}}, \ldots, \iv{c_{z,{\nq}}}, \iv{I_{xx,1}}, \ldots \iv{I_{zz,\nq}}\right)\trans
\end{equation}
% where polynomial zonotopes $\pz{m_j} \subset \R$, $\pz{c_j} \subset \R^3$, and $\pz{I_j}\subset \R^{3 \times 3}$ represent the uncertainties in the mass, center of mass, and inertia tensor of the $j$\ts{th} link, respectively.
where $[m_j] \subset \mathbb{IR}$, $[c_j] \subset \mathbb{IR}^3$, and $[I_j]\subset \mathbb{IR}^{3 \times 3}$ represent the uncertainties in the mass, center of mass, and inertia tensor of the $j$\ts{th} link, respectively.
% An interval overapproximation $\pzparams \subseteq \intparams$ is given by
% \begin{equation}
    % [\Delta] = \left([m_1], \cdots, [m_{\nq}], [c_{x,1}], \cdots, [c_{z,{\nq}}], [I_{xx,1}], \cdots [I_{zz,\nq}]\right)\trans
% \end{equation}
% where $[m_j] = [\setop{inf}(\pz{m_j}), \setop{sup}(\pz{m_j})]$, $[c_j] = [\setop{inf}(\pz{c_j}), \setop{sup}(\pz{c_j})]$, and $[I_j] = [\setop{inf}(\pz{I_j}), \setop{sup}(\pz{I_j})]$.
% where $[m_j] = [\setop{inf}(\pz{m_j}), \setop{sup}(\pz{m_j})] \subset \mathbb{IR}$, $[c_j] = [\setop{inf}(\pz{c_j}), \setop{sup}(\pz{c_j})] \subset \mathbb{IR}^3$, and $[I_j] = [\setop{inf}(\pz{I_j}), \setop{sup}(\pz{I_j})] \subset \mathbb{IR}^{3 \times 3}$.
% where $[m_j] \subset \mathbb{IR}$, $[c_j] \subset \mathbb{IR}^3$, and $[I_j]\subset \mathbb{IR}^{3 \times 3}$ represent the uncertainties in the mass, center of mass, and inertia tensor of the $j$\ts{th} link, respectively.
The true parameters lie in this interval.
% , i.e., $\trueparams \in \intparams$.
Any inertia tensor drawn from $[I_j]$ must be positive semidefinite.
If we let $[\Delta]_j$ denote the $j$\ts{th} component of $[\Delta]$, then $\setop{inf}([\Delta]_j ) > -\infty$  and $\setop{sup}([\Delta]_j) < \infty$ for all $j$.
% In addition, if we let $\pzgreek{\Delta}_\pz{j}$ denote the $j$\ts{th} component of $\pzparams$, then $\setop{inf}(\pzgreek{\Delta}_\pz{j}) > -\infty$  and $\setop{sup}(\pzgreek{\Delta}_\pz{j}) < \infty$ for all $j$.
There exists a nominal vector of inertial parameters $\nomparams \in \intparams$ that serves as an estimate of the true parameters of the system.
\end{assum}
\noindent We do not assume that $\Delta_0$ is equal to the true parameters.

\subsection{Environment}

% \shrey{figure suggestion: there should be an environment with obstacles figure in the numerical examples section, so it would be helpful to point to that here and be like ``for an example, look later in the text at Fig. whatever''}

Next, we describe obstacles and the start and goal configurations of the robot.
% The robot's origin is assumed to coincide with the environment's origin without loss of generality.

\subsubsection{Obstacles}
We begin by defining obstacles:
% We denote the set of obstacles as a set $O \subset W$.
% We make the following assumptions about obstacles:
\begin{assum}[Obstacles]
\label{assum:obstacles}
The transformation between the world frame of the workspace and the base frame of the robot is known, and obstacles are expressed in the base frame of the robot.
At any time, the number of obstacles $\nObs \in \N$ in the scene is finite ($\nObs < \infty$).
Let $\obsset$ be the set of all obstacles $\{ O_1, O_2, \ldots, O_{\nObs} \}$.
% \pat{$\obsset$ is a set here, but we use it like it's the union of obstacles when writing the trajopt}
Each obstacle is convex, bounded, and static with respect to time.
% The robot has access to a conservative estimate of the size and location of each obstacle (this work is concerned with planning and control, not perception).
The arm has access to a zonotope overapproximation of each obstacle's volume in workspace.
\end{assum}
\noindent The arm is \defemph{in collision} with an obstacle if $\FO_j(\q) \cap O_i \neq \emptyset$ for any $j \in \Nq$ or $i \in \{1,\ldots,\nObs\}$. 
This work is concerned with planning and control around obstacles, not perception of obstacles.
A convex, bounded object can always be overapproximated as a zonotope \cite{guibas2003zonotopes}.
If one is given a non-convex bounded obstacle, then one can outerapproximate that obstacle by computing its convex hull. 
Dynamic obstacles may also be considered within the \methodname framework by introducing a more general notion of safety \cite[Def. 11]{vaskov2019towards}, but we omit this case in this paper to ease exposition.
We have assumed for convenience that the robot is able to sense all obstacles workspace. 
One could instead assume that the robot has a finite sensing horizon in which it is able to sense obstacles. 
In this instance, one could extend the results regarding the collision free behavior of \methodname{} by applying Thm. 39 in \cite{kousik2020bridging}.
If a portion of the scene is occluded, then one can treat that portion of the scene as an obstacle.

\subsubsection{Task Specification}
We define the arm's task in terms of its start and goal configurations:
\begin{defn}[Start and Goal Configurations]
\label{def:start_and_goal_config}
The robot begins from rest from the known starting configuration $\qstart$.
It has zero initial velocity and zero initial acceleration.
\methodname attempts to find and execute a sequence of safe trajectories from $\qstart$ to a user-specified goal configuration, $\qgoal$.
\end{defn}
\noindent If desired, an end-effector goal position in workspace may be specified instead of the configuration $\qgoal$.
We omit this case as it only affects \methodname{}'s user-specified cost function.

\section{Proposed Method Overview}
\label{sec:planning_formulation}

% \Ram{this section feels like it could just be the introduction to the next section...}
% \shrey{re: Ram, I think this section should come after the current Sec III, and instead be called a ``proposed method overview.''
% Then each following section describes how we enable solving a particular part of the trajopt problem.
% This would make it easier to justify why we use a trajectory parameterization (e.g., instead of optimizing over an infinite-dimensional control signal space), which then justifies the hierarchical planner/controller design, and leads nicely into the theoretical formulation and numerical implementation of reachable sets.}

This section provides a high-level overview of \methodname.
Our objective is to plan safe trajectories in a receding horizon fashion that reach a goal configuration $\qgoal \in Q$.
% , which is specified at each planning iteration by a higher-level planner.
To accomplish this goal, \methodname optimizes over a space of possible desired trajectories.
As described in Sec. \ref{sec:trajectory}, these desired trajectories are chosen from a prespecified continuum of trajectories, with each determined by a \emph{trajectory parameter}, $k \in K$.
During each planning iteration, \methodname selects a  trajectory parameter that can be followed without collisions despite tracking error and model uncertainty while satisfying joint and input limits.
To ensure that \methodname is able to plan in real-time, each desired trajectory is followed while the robot constructs the next desired trajectory for the subsequent step.
Next, we give overviews of the feedback controller, planning time, and trajectory optimization problem.

\subsection{Feedback Controller}
As described in Sec. \ref{sec:controller_design}, we associate a feedback control input over a compact time interval $T = [0, \tfin] \subset \mathbb{R}$ with each trajectory parameter $k \in K$.
This feedback control input is a function of the nominal inertial parameters, $\nomparams$, the interval inertial parameters $\intparams$, and the state of the robot; however, it cannot be a function of the true inertial parameters of the robot because these are not known.
Applying this control input to the arm generates an associated trajectory of the arm. 
This position and velocity trajectory is a function of the true inertial parameters. 
We denote the position and velocity trajectories at time $t \in T$ under trajectory $k \in K$ under the true inertia model parameters $\Delta \in \intparams$ by $q(t;k,\Delta)$ and $\dot{q}(t;k,\Delta)$, respectively.
To simplify notation, \emph{just} in this section, we denote the feedback control input at time $t \in T$ under trajectory $k \in K$ by $u(t;k)$. 
% {\it Note in the remainder of the paper, the control input is allowed to be a function of the nominal inertial parameters, $\nomparams$, the interval inertial parameters, $\intparams$, the state of the system, and the trajectory parameter.}
\begin{rem}[Controller Dependency Notation]
In the remainder of the paper, the control input is a function of the nominal inertial parameters, $\nomparams$, the interval inertial parameters, $\intparams$, the state of the system, and the trajectory parameter.
% Although we have suppressed the dependence for ease of reading in this section, we stress that the control input is a function of the state of the robot arm, which is a function of the (unknown) true inertial parameters. 
\end{rem}
% Note that though we do not make it explicit, the control input can be a function of the nominal inertial parameters, $\Delta_0$, and the state of the robot; however, it cannot be a function of the true inertial parameters of the robot because these are not assumed to be known.
% We note that for control design, we use the interval parameters $\intparams$ which overapproximate the polynomial zonotope parameters $\pzparams$ because this simpler set representation allows the controller to operate online at a higher frequency.

\subsection{Timing}
Because \methodname performs receding horizon planning, we assume without loss of generality that the control input and trajectory begin at time $t=0$ and end at a fixed time $\tfin$. 
To ensure real-time operation, \methodname identifies a new trajectory parameter within a fixed planning time of $t\plan$ seconds, where $t\plan < \tfin$.
\methodname must select a new trajectory parameter before completing its tracking of the previously identified desired trajectory.
If no new trajectory parameter is found in time, \methodname defaults to a braking maneuver that brings the robot to a stop at time $t = \tfin$.
As is described in Sec. \ref{sec:trajectory}, each desired trajectory contains a braking maneuver.

\subsection{Online Trajectory Optimization}
During each receding-horizon planning iteration, \methodname generates a trajectory by solving a tractable representation of the following nonlinear optimization:
\begin{align}
    \label{eq:optcost}
    &\underset{k\in K}{\min} &&\texttt{cost}(k) \\
    % \label{eq:optdyn}
    % & s.t.  && u(t) = \bM(\q, \Delta) \qdd + \bC(\q, \qd, \Delta) \qd \qquad \forall t \in T \nonumber  \\ 
    % &&&  \qquad + \bG(\q, \Delta) \qquad \forall t \in T \\
    \label{eq:optpos}
    &&& q_j(t; k, \Delta) \in [\qlim^-, \qlim^+]  &\forall t \in T, \Delta \in [\Delta], j \in \Nq \\
    \label{eq:optvel}
    &&& \dot{q}_j(t; k, \Delta) \in [\dqlim^-, \dqlim^+]  &\forall t \in T,  \Delta \in [\Delta], j \in \Nq \\
    \label{eq:opttorque}
    &&& u_j(t;k) \in [\ulim^-, \ulim^+]  &\forall t \in T,  \Delta \in [\Delta], j \in \Nq \\
    % \label{eq:optposcon}
    % &&& p_j(t; k) \in [p_j^-, p_j^+]  &\forall t \in T, j \in \Nq \\
    % \label{eq:optorncon}
    % &&& R_i(t) \in [R_i^-, R_i^+]  \forall t \in T, i = 1,\ldots, \Nq \\
    % \label{eq:optorncon1}
    % &&& (R_i(t) z_{i})^{T} z_{i,des} \in [b_i^-, b_i^+]  \forall t \in T, i = 1,\ldots, \Nq \\
    % \label{eq:optorncon2}
    % &&& (R_i(t) x_{i})^{T} x_{i, des} \in [b_i^-, b_i^+]  \forall t \in T, i = 1,\ldots, \Nq \\
    % \label{eq:optorncon1}
    % &&& \hat{z}_{d,j}^{T} \hat{z}_{j}(t; k) \geq z_{d,j}^-  &\forall t \in T, j \in \Nq \\
    % \label{eq:optorncon2}
    % &&& \hat{x}_{d,j}^{T} \hat{x}_{j}(t; k) \geq x_{d, j}^-  &\forall t \in T, j \in \Nq \\
    \label{eq:optcolcon}
    &&& \FO_j(q(t; k,\Delta)) \bigcap \obsset = \emptyset  &\forall t \in T,  \Delta \in [\Delta], j \in \Nq
\end{align}
% \shrey{reminder for later that theoretical/mathemagical objects are typeset whichever way (e.g., $x$) whereas their numerical, polyzono implementations are langle-rangle'd (e.g., $\pz{x}$)}
The cost function \eqref{eq:optcost} specifies a user-defined objective, such as bringing the robot close to some desired goal.
Each of the constraints guarantee the safety of any feasible trajectory parameter as we describe next.
% We define an optimal trajectory in the following way.
The trajectory must be executable by the robot, which means the trajectory must not violate the robot's joint position \eqref{eq:optpos}, velocity \eqref{eq:optvel}, or input \eqref{eq:opttorque} limits,
 These constraints must be satisfied for each joint over the entire planning horizon despite model uncertainty; 
% Although we have suppressed the dependence, we stress that {\it the control input in the remainder of the paper is also function of the state of the robot arm, which is a function of the (unknown) true inertial parameters}. 
The robot must not collide with any obstacles in the environment \eqref{eq:optcolcon}.

Implementing a real-time algorithm to solve this problem is challenging for several reasons.
First, the dynamics of the robot are nonlinear and constructing an explicit solution to them is intractable. 
Second, the constraints associated with obstacle avoidance are non-convex. 
Finally, the constraints must be satisfied for all time $t \in T$.
To address these three challenges, Sec. \ref{sec:algorithm_theory} proposes a method to generate tight overapproximations to the trajectories and constraints that are differentiable and enable the application of fast optimization techniques.

\section{Trajectory Design}
\label{sec:trajectory}

At runtime, \methodname  computes safe trajectories in a receding-horizon manner by solving an optimization program over parameterized trajectories at each planning iteration.
This section describes the parameterized trajectories.

In each planning iteration, \methodname chooses a desired trajectory to be followed by the robot.
These trajectories are chosen from a pre-specified continuum of trajectories, with each uniquely determined by a \textit{trajectory parameter} $k \in K$.
% \shrey{Maybe hint that this prespecified continuum is a design choice, and even if it is pre-specified, it can be made rich enough for handling a quite large array of possible tasks?
% Then cite the ARMTD and other RTD papers.}
% \fix{added some... feel free to edit}
% Let $K \subset \R^{n_k}$, $n_k \in \N$ be a compact set.
The set $K \subset \R^{n_k}$, $n_k \in \N$, is compact and represents a user-designed continuum of trajectories. 
% For example, in Sec. \ref{sec:implementation_trajectories}, we let $K$ represent a set of Bernstein polynomials which stop at a desired configuration.
$K$ can be designed to include trajectories designed for a wide variety of tasks and robot morphologies \cite{holmes2020armtd, kousik2020bridging, kousik2019_quad_RTD, liu2022refine}, as long as each trajectory satisfies the following definition.
% We say that $k \in K$ maps to a desired trajectory $q_d(\cdot; k) : T \to \Q$, and use $\qdesk$ to denote the configuration along this trajectory parameterized by $k$ at time $t$.
\begin{defn}[Trajectory Parameters]
\label{def:traj_param}
For each $k \in K$, a \emph{desired trajectory} is an analytic function $q_d(\,\cdot\,; k) : T \to Q$ that satisfies the following properties:
\begin{outline}[enumerate]
% We require that $q_d(\cdot; k): T \to Q$ satisfies three properties for all $k \in K$
\1 The trajectory starts at a specified initial condition $(\initq, \initdq, \initddq)$, so that $q_d(0; k) = \initq$, $\dot{q}_d(0; k) = \initdq$, and $\ddot{q}_d(0; k) = \initddq$.
\1 $\dot{q}_d(\tfin; k) = 0$ and $\ddot{q}_d(\tfin; k) = 0$ (i.e., each trajectory brakes to a stop, and at the final time has zero velocity and acceleration).
\1 $\ddot{q}_d(\,\cdot\,; k)$ is Lipschitz continuous and bounded.
% \item $q_d(\cdot; k) $ is at least once-differential with respect to time (i.e., no discontinuities in joint position or velocity).
% \pat{Third, $q(0; k) = 0$ (each joint starts from zero initial angle).}
\end{outline}
\end{defn}
\noindent The first property allows for desired trajectories to be generated online.
In particular, recall that \methodname performs real-time receding horizon planning by executing a desired trajectory computed at a previous planning iteration while constructing a desired trajectory for the subsequent time interval.
% In particular, recall that \methodname performs real-time receding horizon planning. 
% It does this by constructing desired trajectories to follow for predicted future positions while following a desired trajectory computed at a previous time step.
The first property allows desired trajectories that are generated by \methodname to begin from the appropriate future initial condition of the robot.
The second property ensures that a fail safe braking maneuver is always available.
The third property ensures that desired position, velocity, and acceleration trajectories for the tracking controller are sufficiently continuous to prove a tracking error bound as we describe next.

\section{Controller Design} 
\label{sec:controller_design}

% \jon{motivate this section a little better. talk about developing a robust passivity-based control with ultimate tracking performance. Finish by saying something like "we will show in the next section how the ultimate bound helps to ensure safe planning}
% \fix{Added a little bit}
This section describes a robust passivity-based controller that is used to follow the desired trajectories described in the previous section.
The controller conservatively handles uncertainty in the manipulator's dynamics stemming from uncertain inertial parameters, and provides a bound on the worst-case tracking error.
% This tracking error bound is critical to our planning algorithm because it allows us to account for the worst-case tracking performance when following any desired trajectory.
We use this bound to develop obstacle avoidance and torque limit constraints in Sec. \ref{sec:algorithm_theory} that are guaranteed to be satisfied when tracking a feasible desired trajectory.
% Importantly, we show via experiments in Sec. \ref{sec:experiments} that this robust approach is not overly conservative.
% \pat{I didn't know where to put the k-dependence $( \cdot ; k)$ in this section... it seems like it should go everywhere? But maybe that's too heavy?}
% \shrey{Probably introduce the $k$-dependence up front with $(\cdot ; k)$ then say we drop it to ease notation going forward}
% In this section we follow the method developed by \cite{giusti2016bound, giusti2017efficient, giusti2021interval} for designing a passivity-based tracking controller \cite{theory_of_robot_control} that is robust to disturbances caused by uncertainty in the robot's inertial parameters. 
% \Ram{do you really need to show the dependence on time or k here? Why not just write $\dot{q}_a$ for instance?}
% \fix{Removed all dependence on $t$ and $k$, and added text about dropping:}
% \shrey{add a sentence at a high level explaining all the following subsections and why they matter}
% \fix{reorganized and added}
% We note that for control design, we utilize the interval parameters $\intparams$ which overapproximate the polynomial zonotope parameters $\pzparams$ because this simpler set representation allows the controller to operate online at a higher frequency.

The robust passivity-based controller is a feedback control input that is a function of a desired trajectory parameter. 
Recall that in Sec. \ref{sec:planning_formulation}, we denote the control input at time $t \in T$ under trajectory $k \in K$ by $u(t;k)$ even though it is a function of the nominal inertial parameters, $\nomparams$, the interval inertial parameters, $\intparams$, and the state of the system.
Because this section focuses on proving several important properties regarding the continuity of the controller, we are deliberate in describing the controller's dependencies.
Note that all desired and actual trajectories of the robot depend on the trajectory parameter $k \in K$; however, to avoid overburdening the reader, and because it is clear in context, we drop the dependence of the trajectory on $k$.

To formulate the controller, we define a \defemph{total feedback trajectory}, $\qA$, that is made up of the robot trajectory and the desired trajectory and is defined as:
\begin{equation} \label{eq:total_feedback_traj}
\qA(t) =
    \begin{bmatrix}
        q(t)\trans & \dot{q}(t)\trans  & q_d(t)\trans & \dot{q}_d(t)\trans & \ddot{q}_d(t)\trans 
    \end{bmatrix}\trans.
\end{equation}
% at time $t$.
Using $\qA$, we can define a \defemph{modified desired velocity trajectory}, $\dot{q}_a$, and \defemph{modified desired acceleration trajectory}, $\ddot{q}_a$ as:
\begin{equation}\label{eq:modified_ref} % modified error
\begin{split}
    \dot{q}_a(t) &= \dot{q}_d(t) + K_r \err(t), \Hquad \err(t) = q_d(t) - q(t) \\
    \ddot{q}_a(t) &= \ddot{q}_d(t) + K_r \errdot(t), \Hquad \errdot(t) = \dot{q}_d(t) - \dot{q}(t)
    \end{split}
\end{equation}
where the gain matrix $K_r$ is a diagonal positive definite matrix.
% We concatenate these trajectories together and define a total feedback trajectory, $\qA$ at time $t$ as:
% \begin{equation}
% \qA(t) = \begin{bmatrix} q(t) & \dot{q}(t) & \ddot{q}(t) & q_d(t) & \dot{q}_d(t) & \ddot{q}_d(t) \end{bmatrix}.
% \end{equation}
These trajectories do depend on the trajectory parameter $k \in K$; however, we have dropped this dependence for ease of reading in the remainder of the discussion.

The robust passivity-based control input at time $t \in T$ is $u(\qA(t), \nomparams, \intparams) \in \R^{\nq}$, which is composed of a \defemph{passivity-based nominal input} $\tau(\qA(t), \nomparams) \in \R^{\nq}$, as well as a \defemph{robust input} $\robv(\qA(t),\nomparams, \intparams) \in \R^{\nq}$:
\begin{equation}\label{eq:controller}
u(\qA(t), \nomparams, \intparams) = \tau(\qA(t), \nomparams) - \robv(\qA(t), \nomparams, \intparams).
\end{equation}
This decomposition is similar to the one proposed in \cite{giusti2016bound, giusti2017efficient}.
The justification for this decomposition is that the nominal control input is unable to perfectly execute the desired trajectory because of a possible mismatch between the nominal inertial parameters $\nomparams$ and the true parameters $\trueparams$.
Thus, $\robv$ is introduced as an additional term to guarantee robustness by compensating for the worst possible disturbances stemming from this mismatch in inertial parameters.
% \shrey{Replace ``enhance'' with ``guarantee'' and maybe replace ``derive'' with ``construct'' to sound less wishy-washy and more impressive.}
% \fix{done}
Despite the similarities to the decomposition proposed in \cite{giusti2016bound, giusti2017efficient}, our formulation of $\robv$ is distinct and significantly improves upon prior approaches by providing the same tracking error bound while requiring a smaller bound on the robust input $\robv$, as is illustrated in \href{https://github.com/roahmlab/armour/blob/main/assets/TRO_Armour_Appendix_F.pdf}{online App. F}
% \pat{we probably want to back this claim up somewhere...}
Next, we define $\tau$ and $\robv$, and conclude this section by describing how to compute $\tau$ and $\robv$.

\subsection{Nominal and Robust Controllers}
Per Ass. \ref{assum:arm_model}, there exists a nominal model of the robot's dynamics given by inertial parameters $\nomparams$.
Similar to \cite[(4)]{giusti2016bound}, our nominal input stems from passivity-based control:
\begin{equation}\label{eq:nominal_controller}
\begin{split}
\tau(\qA(t),\nomparams) =~&\bM(q(t), \nomparams) \ddot{q}_a(t) + \\  &+ \bC(q(t), \dot{q}(t), \nomparams) \dot{q}_a(t) + \bG(q(t), \nomparams).
\end{split}
\end{equation}
% where $\dot{q}_a$ and $\ddot{q}_a$ are the modified reference velocity and acceleration defined by
% \begin{align}\label{eq:modified_ref} % modified error
%     \dot{q}_a(t) &= \dot{q}_d(t) + K_r \err(t), \Hquad \err(t) = q_d(t) - q(t) \\
%     \ddot{q}_a(t) &= \ddot{q}_d(t) + K_r \errdot(t), \Hquad \errdot(t) = \dot{q}_d(t) - \dot{q}(t)
% \end{align}
% where the gain matrix $K_r$ is a diagonal positive definite matrix.
% The nominal controller is unable to perfectly control the manipulator because of a possible mismatch between the nominal parameters $\nomparams$ and the true parameters $\trueparams$.
We treat the torques that arise from the possible mismatch between the nominal parameters $\nomparams$ and the true parameters $\trueparams$ as a disturbance $\wdist$, where 
\begin{equation}\label{eq:disturbance_equation}
\begin{split}
    \wdist(\qA(t), &\nomparams, \Delta) =  (\bM(q(t),\trueparams) -\bM(q(t),\nomparams)) \ddot{q}_a(t) + \\
 &+ (\bC(q(t), \dot{q}(t), \trueparams) - \bC(\q(t), \dot{q}(t), \nomparams))\dot{q}_a + \\ &+ \bG(q(t),\trueparams) - \bG(q(t),\nomparams).
 \end{split}
    \end{equation}
\noindent In practice, $\wdist$ is unknown, but can be bounded by
\begin{equation}\label{eq:disturbance_vector}
\wdist(\qA(t), \nomparams, \Delta) \in \iv{\wdistinterval(\qA(t), \nomparams, \intparams)},
\end{equation}
where the right hand side of the previous equation is computed by plugging $\intparams$ into \eqref{eq:disturbance_equation} in place of $\trueparams$ and using interval arithmetic.
We describe how to compute $\iv{\wdistinterval(\qA(t), \nomparams, \intparams)}$ using Alg. \ref{alg:IRNEA} in the next subsection.

To create our controller, we next define a function that measures the worst case disturbance:
\begin{defn}[Worst Case Disturbance]
\label{def:worst_case_disturbance}
Suppose that we have $\setop{inf}([\wdistinterval(\qA(t),\nomparams, \intparams)]) =\underline{\wdistinterval}$ and $\setop{sup}( [\wdistinterval(\qA(t),\nomparams,\intparams)]) = \overline{\wdistinterval}$.
Then the \defemph{measure of worst case disturbance} is the function $\rho: \mathbb{I}\mathbb{R}^{\nq} \xrightarrow{} \mathbb{R}^{\nq}$ defined as
% \begin{equation}
%     \rho([\bPhi]) = \text{max }(|\underline{\bPhi}|, |\overline{\bPhi}|)\label{eq:worst_case_disturbance},
% \end{equation}
\begin{equation}
    \rho([\wdistinterval(\qA(t),\nomparams,\intparams)]) = \max(|\underline{\wdistinterval}|, |\overline{\wdistinterval}|)\label{eq:worst_case_disturbance},
\end{equation}
where the $\max$ operator is applied element-wise.
% over the interval vector argument.
\end{defn}
% The norm of the disturbance $\norm{\wdist(\qA(t),\nomparams,\Delta)}$ is always bounded by the norm of the worst-case disturbance \eqref{eq:worst_case_disturbance}:
% \begin{lem}\label{lem:disturbance_bound}
% For all $\qA(t) \in \mathbb{R}^{6\nq}$ and $\trueparams \in \intparams$
% % \begin{equation*}
% % \rho_i([\Phi]) \geq w_i(\q, \qd, \qd_a, \qdd_a, \Delta).    
% % \end{equation*}
% % \begin{equation}
% % \normwmax \geq \norm{\wdist}.    
% % \end{equation}
% % where $\normwmax = \norm{{\rho([\wdistinterval])}}$.
% \begin{equation}
% \norm{{\rho([\wdistinterval(\qA(t), \nomparams,\intparams)]}} \geq \norm{\wdist(\qA(t), \nomparams, \Delta)}.
% \end{equation}
% \end{lem}
% \begin{proof}
% Throughout the proof, we suppress the dependence on $\qA(t)$, $\Delta$, and $\nomparams$ for convenience.  
% Notice that the true disturbance $w$ is contained in the interval $\iv{w}$ as in \eqref{eq:disturbance_vector}.
% Next, from Def. \ref{def:worst_case_disturbance}, $\rho([\wdistinterval]) \geq |w| \; \forall w \in \iv{w}$.
% Because $\norm{w} = \norm{(\abs{w})}$, we have $\norm{{\rho([\wdistinterval])}} \geq \norm{w}$.
% \end{proof}
\noindent The absolute value of the disturbance $\abs{\wdist(\qA(t),\nomparams,\Delta)}$ is always bounded by the worst-case disturbance \eqref{eq:worst_case_disturbance}:
\begin{lem}[Disturbance Bound]
\label{lem:disturbance_bound}
For all $\qA(t) \in \mathbb{R}^{6\nq}$ and $\trueparams \in \intparams$
% \begin{equation*}
% \rho_i([\Phi]) \geq w_i(\q, \qd, \qd_a, \qdd_a, \Delta).    
% \end{equation*}
% \begin{equation}
% \normwmax \geq \norm{\wdist}.    
% \end{equation}
% where $\normwmax = \norm{{\rho([\wdistinterval])}}$.
\begin{equation}
\rho([\wdistinterval(\qA(t), \nomparams,\intparams)] \geq \abs{\wdist(\qA(t), \nomparams, \Delta)},
\end{equation}
where the inequality is applied element-wise.
\end{lem}
\begin{proof}
Throughout the proof, we suppress the dependence on $\qA(t)$, $\Delta$, and $\nomparams$ for convenience.  
The true disturbance $w$ is contained in the interval $\iv{w}$ as in \eqref{eq:disturbance_vector}.
Therefore, $\abs{w} \in [0, \max(|\underline{\wdistinterval}|, |\overline{\wdistinterval}|)]$.
From Def. \ref{def:worst_case_disturbance}, $\rho([\wdistinterval]) = \max(|\underline{\wdistinterval}|, |\overline{\wdistinterval}|)]$, so $\rho([\wdistinterval]) \geq |w|, \; \forall w \in \iv{w}$.
% Because $\norm{w} = \norm{(\abs{w})}$, we have $\norm{{\rho([\wdistinterval])}} \geq \norm{w}$.
\end{proof}
% \shrey{Proof or citation?}
% \fix{added quick proof}
% The disturbance $\wdist$ is always bounded by the worst-case disturbance \eqref{eq:worst_case_disturbance}:
% \begin{lem}\label{lem:disturbance_bound}
% For all $\q, \qd, \qadot, \qaddot \in \mathbb{R}^{\nq}$ and $\trueparams \in \intparams$
% % \begin{equation*}
% % \rho_i([\Phi]) \geq w_i(\q, \qd, \qd_a, \qdd_a, \Delta).    
% % \end{equation*}
% \begin{equation*}
% \rho_i([\wdistinterval]) \geq \wdistlongi.    
% \end{equation*}
% \end{lem}

% \eqref{eq:disturbance_vector} is indeed bounded by the worst-case disturbance \eqref{eq:worst_case_disturbance}.
% \begin{lem}\label{lem:disturbance_bound}
% For all $\q, \qd, \qd_a, \qdd_a \in \mathbb{R}^{\nq}$ and $\Delta \in \intparams$
% \begin{equation*}
% \rho_i([\Phi]) \geq w_i(\q, \qd, \qd_a, \qdd_a, \Delta).    
% \end{equation*}
% \end{lem}

% \jon{\text{add $k$-dependence}}
% \fix{no longer required}
Plugging \eqref{eq:controller}, \eqref{eq:nominal_controller}, and \eqref{eq:disturbance_equation}  into \eqref{eq:manipulator_equation} one obtains 
% % worst-case disturbance
\begin{equation}
    \label{eq:modified_manipulator_equation}
\begin{split}
    \robv(\qA(t), \nomparams, \intparams) + \wdist(\qA(t), \nomparams,\Delta) = \bM(q(t),\trueparams)\robrdot(t) + \\ + \bC(q(t), \dot{q}(t), \trueparams)\robr(t)
    \end{split}
\end{equation}
where $\robr(t)$ is the \defemph{modified tracking error}
\begin{equation} \label{eq:modified_error}
    \begin{split}
    \robr(t) &= \errdot(t) + K_r \err(t)\\
    \robrdot(t) &= \errddot(t) + K_r \errdot(t).
    \end{split}
\end{equation}
Our goal is to obtain a uniform bound on the modified tracking error, $r: T \to \R^{\nq}$.
% As a consequence of Assumption \ref{assum:initial_condition_known}, the following is true.
% \begin{prop}
% \label{prop:initial_tracking_error}
% At the start of \methodname's first planning iteration, Assum. \ref{assum:initial_condition_known} states that the initial condition is known.
% By Def. \ref{def:traj_param}, the desired trajectory is designed to start from this initial condition
% Therefore, there is no initial error in position or velocity, i.e. $\err(0) = 0$ and $\errdot(0) = 0$.
% Observing \eqref{eq:modified_error}, this implies that $\robr(0) = 0$ as well.
% \end{prop}
% \noindent
% We will use this proposition shortly to obtain a uniform bound on the modified tracking error $r: T \to \R^{\nq}$.
% \begin{align}
%     \robr &= \errdot + K_r \err \\
%     \robrdot &= \errddot + K_r \errdot.
%     \label{eq:modified_error}
% \end{align}
% \begin{equation}\label{eq:disturbance_vector}
% \wdist \subseteq   [\bPhi] = \wdistinterval,
% \end{equation}
% where $[\Phi]$ is the interval disturbance vector.
% Using the following definition, we can now measure the size of the interval $[\bPhi]$.
% We use the previous lemma, along with the following proposition, to develop 
To do this, we make the following assumption about the mass matrix $\bM(q(t),\trueparams)$:
% In fact one can 
% One useful fact \shrey{this phrasing feels so casual but I can't figure out how to rewrite it...} about mass matrices for multi-link robots composed of only revolute joints is that the smallest eigenvalue of the robot can be bounded from below across all configurations \cite{Ghorbel1998uniform}.
% this is the reference proving that property https://www.ams.org/journals/proc/1965-016-01/S0002-9939-1965-0171902-8/S0002-9939-1965-0171902-8.pdf
\begin{assum}[Mass Matrix Eigenvalue Bound]
\label{ass:eigenvalue_bound}
% Let $\intparams$ be an interval of inertial parameters.
% and let $[\bM] = [\bM(\cdot,\underline{\trueparams}), \bM(\cdot,\overline{\trueparams})]$ be the interval of corresponding inertia matrices. 
% Define $\lambdamin(\bM(q,\trueparams))$ and $\lambdamax(\bM(q,\trueparams))$ to be the minimum and maximum eigenvalues of $\bM(q,\trueparams)$, respectively.
Let $\lambdamin(\bM(q(t),\trueparams))$ and $\lambdamax(\bM(q(t), \trueparams))$ be the minimum and maximum eigenvalues of $\bM(q(t),\trueparams)$.
% Let $\sigma_{m}(\trueparams)$ and $\sigma_{M}(\trueparams)$ be uniform bounds on the eigenvalues of $\bM(\cdot, \trueparams)$ \cite{Ghorbel1998uniform}.
% There exist ultimate bounds $\sigma_m$ and $\sigma_M$ on the eigenvalues of the mass matrix so that
There exist finite uniform bounds $\sigma_m > 0$ and $\sigma_M > 0$ on these eigenvalue so that
\begin{align}
    0 &< \sigm \leq \lambdamin(\bM(q(t),\trueparams)) \\
    0 &< \lambdamax(\bM(q(t),\trueparams)) \leq \sigM
\end{align}
holds for all $q(t) \in Q$ and $\trueparams \in \intparams$. 
% Let $\sigma_{m}(\trueparams)$ and $\sigma_{M}(\trueparams)$ be uniform bounds on the eigenvalues of $\bM(\q, \trueparams)$ for all $\q \in Q$ \cite{Ghorbel1998uniform}.
% Then for all $\trueparams \in \intparams$ and $\q \in Q$ we have
% \begin{align}
%     \sigma_{m}(\underline{\trueparams}) &\leq \lambdamin(\Mq)\\
%     \sigma_{M}(\overline{\trueparams}) &\geq \lambdamax(\Mq).
% \end{align}
\end{assum}
\noindent Using analysis, one can prove that if  $\bM(q(t),\trueparams)$ is positive definite for all $q(t)$ and $\trueparams$, then $\sigma_m$ is at least equal to zero. 
However proving that this lower bound is greater than zero requires more explicit structure on $\intparams$. 
Because this is not the emphasis of this paper, we make this assumption and verify that it is satisfied numerically during experiments.

We next state a theorem whose proof can be found in App. \ref{app:robust_controller_proof} that describes how to define
% \shrey{Probably instead of ``complete'' it would be better to say, how to incorporate the disturbance or something?}
% \fix{reworded}
the robust input $v$ in \eqref{eq:controller} by making use of \eqref{eq:worst_case_disturbance} and the previous assumption while ensuring a uniformly bounded tracking error for $r$. 
% % ultimate bound
% \begin{thm}\label{thm:ultimate_boundedness}
% % Suppose there is a known interval $\intparams$ of inertial parameters for a rigid serial manipulator.  
% % Suppose there is a known interval $\intparams$ of inertial parameters.  
% Let $\Delta \in \intparams$ be arbitrary and let the true dynamics be given by \eqref{eq:manipulator_equation}. 
% Let $\kappa (t)$ and $\varphi (t)$ be two positive increasing functions such that $\kappa_P = \min{\kappa (t)}$ and $\varphi_P = \min{\varphi (t)}$.
% If the robust input $v$  in \eqref{eq:controller} is defined as
% \begin{equation}\label{eq:robust_input}
%     v = (\kappa(t) + \|\rho([\Phi]) \| + \varphi(t)) r,
% \end{equation}
% then there exists a finite time $\tau$ such that for all $t \geq \tau$
% \begin{equation}\label{eq:ultimate_bound}
%     \|r(t)\| \leq \frac{1}{\kappa_P} \sqrt{\frac{\sigma_M(\overline{\Delta})}{\sigma_m(\underline{\Delta})}}.
% \end{equation}
% \end{thm}

\begin{thm}[Uniform Tracking Error Bound]
\label{thm:robust_controller}
    % \pat{Robust input gives ultimate bound $\norm{\robr} \leq \ultbound \, \forall t > t_1$. Need to rearrange so these terms are introduced before the proof}
    Suppose Ass. \ref{ass:eigenvalue_bound} is satisfied.
    Let $\roblevel > 0$ be a user-specified constant.
    Consider the following candidate Lyapunov Function
    \begin{equation}
        \label{eq:lyap_r}
        \roblyap = \tfrac{1}{2}\robr(t)\trans \bM(q(t),\trueparams) \robr(t).
    \end{equation}
    Let $\underline{h}$ be a function that is continuous in its first argument such that 
    % \begin{equation}
    %     \label{eq:CBF_defn}
    %     \robh = -\roblyap + \roblevel.
    % \end{equation}
    % Next, let 
    \begin{equation}
        \label{eq:CBF_lower_bound}
        \robhmin \leq 
        -\sup_{\trueparams^* \in \intparams}
                \left( V(\qA(t),\Delta^*) \right) + \roblevel,
    \end{equation}
    for all $\qA(t)$.
    % and therefore $\robhmin \leq \robh$ for all $\q \in Q$ and $\robr \in \R^{\nq}$.
    % \shrey{It seems bits of the proof have leaked into this theorem statement}
    % \fix{removed}
    % For brevity, we define $\normwmax = ||\rho([\wdistinterval])||$.
 Let $\robKinf:\R \to \R$ be some extended class $\mathcal{K}_\infty$ function (i.e., $\robKinf: \R \to \R$ is strictly increasing with $\robKinf(0) = 0$).
 Let $\wmax$ be any continuous function in its first argument such that
%  \begin{equation}
%  \label{eq:cond_wmax}
%     \norm{\wmax(\qA(t), \nomparams,\intparams)}\geq \norm{{\rho([\wdistinterval(\qA(t), \nomparams,\intparams)]}},
%  \end{equation}
 \begin{equation}
 \label{eq:cond_wmax}
    \wmax(\qA(t), \nomparams,\intparams)\geq {\rho([\wdistinterval(\qA(t), \nomparams,\intparams)]},
 \end{equation}
 for all $\qA(t)$, $\nomparams$, and $\intparams$, where the inequality is applied element-wise.
 Suppose that $e(0) = \dot{e}(0) = 0$ and let
\begin{equation}
    \label{eq:ultimate_bound_defn}
    \epsilon \coloneqq \ultbound.
\end{equation}
If the robust input is set equal to
\begin{equation}
    \label{eq:robust_input}
    \robv(\qA(t), \nomparams, \intparams ) = \begin{cases} 
    -\robcoeff(\qA(t),\nomparams, \intparams) \frac{\robr(t)}{\norm{\robr(t)}} & \norm{\robr(t)} \neq 0 \\ 
    \phantom{-} 0 & \norm{\robr(t)} = 0 
    \end{cases} 
\end{equation}
with
% \begin{equation}
%     \label{eq:robust_coeff}
%     \begin{split}
%     \robcoeff(\qA(t),\nomparams,\intparams) = \max\Big(0, \frac{-\robKinf(\robhmin)}{\norm{\robr(t)}} +  \\ \phantom{ \robcoeff(\qA(t),\nomparams,\intparams) =}  + \norm{\wmax(\qA(t),\nomparams,\intparams)}\Big),
%     \end{split}
% \end{equation}
\begin{equation}
    \label{eq:robust_coeff}
    \begin{split}
    \robcoeff(\qA(t),\nomparams,\intparams) = \max\Big(0, \frac{-\robKinf(\robhmin)}{\norm{\robr(t)}} +  \\ \phantom{ \robcoeff(\qA(t),\nomparams,\intparams) =}  + \frac{\abs{r(t)}\trans\wmax(\qA(t),\nomparams,\intparams)}{\norm{r(t)}}\Big),
    \end{split}
\end{equation}
then the modified tracking error trajectories $r: T \to \R^{\nq}$ are uniformly bounded by
\begin{equation}
    \label{eq:ultimate_bound}
    \norm{\robr(t)} \leq \epsilon, \qquad \forall t \in T
\end{equation} 
when tracking desired trajectories that satisfy Def. \ref{def:traj_param}.
\end{thm}

% \begin{proof}
% The proof is given in Appx. \ref{app:robust_controller_proof}.
% \Ram{maybe move this to an appendix?}
% \fix{done}
% \end{proof}

Theorem \ref{thm:robust_controller} provides a uniform bound on the tracking error trajectories $\robr: T \to \R^{\nq}$. 
This result can be used to bound the position and velocity tracking errors.
% We utilize these bounds to synthesize desired trajectories in Sec. \ref{sec:algorithm_theory}.
The following claim, whose proof can be found in App. \ref{app:tracking_error}, is inspired by \cite[Cor. 6]{giusti2016bound} and \cite[Proof of Theorem 2.3]{de2012theory}, but is distinct because it considers a uniform bound on $\robr: T \to \R^{\nq}$ \eqref{eq:ultimate_bound} that applies for all $T$, rather than an ultimate bound that is only valid after some finite time:
% as a corollary, the tracking error is uniformly bounded
\begin{cor}[Tracking Performance]
\label{cor:tracking_error}
Suppose Ass. \ref{ass:eigenvalue_bound} is satisfied and $e(0) = \dot{e}(0) = 0$.
If $\robr: T \to \R^{\nq}$ is uniformly bounded as in \eqref{eq:ultimate_bound}, then the controller \eqref{eq:controller} with robust input \eqref{eq:robust_input} reaches any desired tracking performance provided a large enough gain matrix $K_r$.
In particular, letting
\begin{align}
    % \pboundj &\coloneqq \frac{1}{K_{r, j}}\ultbound \quad \regtext{and}\label{eq:ultimate_bound_position}\\
    % \vbound &\coloneqq 2\ultbound\label{eq:ultimate_bound_velocity},
    \pboundj &\coloneqq \frac{\epsilon}{K_{r, j}} \quad \regtext{and}\label{eq:ultimate_bound_position}\\
    \vbound &\coloneqq 2\epsilon\label{eq:ultimate_bound_velocity},
\end{align}
where $\epsilon$ as in \eqref{eq:ultimate_bound_defn} and $K_{r, j}$ is the $j$\ts{th} element of the diagonal gain matrix $K_r$, then $|\errj(t)| \leq \pboundj$ and $|\errjdot(t)| \leq \vbound$ for all $t \in T$ and $j \in \Nq$.
\end{cor}
\noindent Though these bounds can be made arbitrarily small by either increasing $K_r$ or decreasing $V_m$, satisfying these bounds may require large torques, causing input saturation.
In Sec. \ref{sec:algorithm_theory}, we explain how to overapproximate the torques required to track a given desired trajectory including all possible tracking error.
% We use these overapproximations to design trajectories that are guaranteed to be realizable on the robot in Sec. \ref{sec:implementation}.

Before proceeding further, we make one final remark about the validity of these bounds for all time.
\begin{rem}[Error Bound for All Time]
\label{rem:combining_traj}
% Suppose the initial condition of the robot is known at the first planning iteration. 
By Def. \ref{def:start_and_goal_config}, the robot starts from configuration $\qstart$ with zero initial velocity and acceleration.
Suppose that the initial condition of the desired trajectory in the first planning iteration matches this (i.e., $\initq = \qstart$, $\initdq = \zeros$, $\initddq = \zeros$), and that the initial condition of all subsequent desired trajectories match the state of the previous desired trajectory at $t = t\plan$.
Then, one can satisfy the bounds described in Thm. \ref{thm:robust_controller} and Cor. \ref{cor:tracking_error} for all time.
% If the initial condition of the desired trajectory is equal to
% \begin{enumerate}
%     \item first planning iteration $\initq = \qstart$
%     \item the state of the previous desired trajectory at $t = t\plan$
% \end{enumerate}
% If the initial condition of the desired trajectory is equal to either (1) the initial condition of the robot at the first planning iteration or (2) the state of the previous desired trajectory at $t = t_p$ for all subsequent planning iterations, then one can satisfy the bounds described in Theorem \ref{thm:robust_controller} and Corollary \ref{cor:tracking_error} for all time. 
\end{rem}
\noindent This remark follows directly from the proofs of Thm. \ref{thm:robust_controller} and Cor. \ref{cor:tracking_error} by noticing that the arguments in their proofs hold for desired trajectories that are defined for all time rather than just for $t \in T$. 
One can create a valid desired trajectory defined for all time by applying the strategy described in Rem. \ref{rem:combining_traj}.
This is the approach we adopt while performing receding horizon planning as described in Sec. \ref{sec:algorithm_theory}. 
To apply Rem. \ref{rem:combining_traj}, one only needs to know the final state of the previous \defemph{desired trajectory} rather than the final state of the robot's \defemph{actual trajectory}.
% does not need to know the final state of the robot at each planning iteration. 
% Instead, one only needs to know the final state of the desired trajectory to apply Remark \ref{rem:combining_traj}

% \shrey{figure suggestion: show the tracking error and the relevant level set of the Lyapunov function.
% It would be even better to illustrate it as a comparison against the worse Giusti robust bound like Patrick did in that one meeting a while ago.}

% \begin{figure}[t]
%     \centering
%     \includegraphics[width=0.9\columnwidth]{figures/example_joint_trajectory.png}
%     \caption{Example parameterized trajectory (blue, corresponding to point $k$ in the parameter space on the left) for joint $i$.
%     The region of attraction for our robust controller is shown as the blue volume around the trajectory, and a particular executed trajectory is shown in black.}
%     \label{fig:example_environment}
% \end{figure}

% clear from the PROOF of the theorem 
% \jon{I think I need to modify this statement. I think both the initial position and velocity need to be bounded, e.g. the ultimate bound is satisfied}
% It is clear from Theorem \ref{thm:ultimate_boundedness} and Corollary \ref{thm:tracking_error} that if the norm of the initial tracking error $\|q_d(0) - q(0)\|$ is less than the ultimate bound \eqref{eq:ultimate_bound}, then $\|q_d(t) - q(0)\|$ remains bounded by \eqref{eq:ultimate_bound} for all time.  

% We will use this result in Section \ref{sec:online_planning} to guarantee safe motion planning.

\subsection{Controller Implementation}

\begin{algorithm}[t]
\small
\begin{algorithmic}[1]
\State Compute $\dot{q}_a(t)$ and $\ddot{q}_a(t)$ using \eqref{eq:modified_ref}.
\State Initialize base linear acceleration $a_0^0$.
% \State{\bf parfor} $t \in T$ // parallel for each time step

\State\hspace{0in}{\bf for} $j = 1:\nq$ // for each joint

    \State\hspace{0.2in} $\nom{R\jssmu}, p\jssmu \gets \nom{\homtrans\jssmu}(\nom{\qj})$ as in \eqref{eq:homogeneous_transform}
    % \Ram{need to define this}
    
    \State\hspace{0.2in} $\nom{R\jssm} \gets \texttt{transpose}(\nom{R\jssmu})$
    
\State\hspace{0in}{\bf end for}

\State\hspace{0in}{\bf for} $j = 1:\nq$ // for each joint (forward recursion)

    \State\hspace{0.2in} $\nom{\omega\jss} \gets \nom{R\jssm}  \nom{\omega\jssmm} + \nom{\qdj}\hat{z}_j$
    
    \State\hspace{0.2in} $\nom{\omega\jssa} \gets \nom{R\jssm}  \nom{\omega\jssmma} + \nom{\qajdot}\hat{z}_j$
    
    \State\hspace{0.2in} $\nom{\dot{\omega}\jss} \gets \nom{R\jssm}  \nom{\dot{\omega}\jssmm} + ((\nom{R\jssm}  \nom{\omega\jssmma}) \times (\nom{\qdj} \hat{z}_j)) + \nom{\ddot{q}_{a,j}(t)}\hat{z}_j$
    
    \State\hspace{0.2in} $\nom{a\jss} \gets (\nom{R\jssm}  \nom{a\jssmm}) + (\nom{\dot{\omega}\jss} \times \nom{p\jssmu}) + (\nom{\omega\jss} \times (\nom{\omega\jssa} \times \nom{p\jssmu}))$
    
    \State\hspace{0.2in} $\iv{a\jssc} \gets \nom{a\jss} + (\nom{\dot{\omega}\jss} \bigotimes \iv{c\jss}) + (\nom{\omega\jss} \bigotimes (\nom{\omega\jssa} \bigotimes \iv{c\jss})) $
    
    \State\hspace{0.2in} $\iv{F\jss} \gets \iv{m_i}  \iv{a\jssc}$
    
    \State\hspace{0.2in} $\iv{N\jss} \gets \iv{I\jss}  \nom{\dot{\omega}\jss} \bigoplus (\nom{\omega\jssa}  \bigotimes (\iv{I\jss}  \nom{\omega\jss})) $

\State\hspace{0in}{\bf end for}

% \State\hspace{0in}Initialize $\iv{f^{\nq+1}_{\nq+1}}, \iv{n^{\nq+1}_{\nq+1}}, \nom{R^{\nq}_{\nq+1}}$ \jon{i'll fix this.}

\State\hspace{0in} $\nom{R^{\nq}_{\nq+1}} \leftarrow I_{3 \times 3}$ 

\State\hspace{0in}$\iv{f^{\nq+1}_{\nq+1}} \leftarrow \zeros$ 
\State\hspace{0in}$\iv{n^{\nq+1}_{\nq+1}} \leftarrow \zeros$ 

\State\hspace{0in}{\bf for} $j = \nq:-1:1$ // for each joint (backward recursion)

\State\hspace{0.2in} $\iv{f\jss} \gets (\nom{R\jssp}  \iv{f^{j+1}_{j+1}}) \bigoplus \iv{F\jss} $

\State\hspace{0.2in} $\iv{n\jss} \gets (\nom{R\jssp}  \iv{n^{j+1}_{j+1}}) \bigoplus (\iv{c\jss} \bigotimes \iv{F\jss}) \bigoplus \newline \hspace*{7em} \bigoplus (\nom{p\jssmj} \bigotimes (\nom{R\jssp}  \iv{f^{j+1}_{j+1}})) \bigoplus \iv{N\jss} $ 
\State\hspace{0.2in} $\iv{u_j(t,\intparams)} \gets \hat{z}_j^\top\iv{n\jss}$

\State\hspace{0in}{\bf end for}

% \State{\bf end parfor}
\end{algorithmic}
\caption{\small \phantom{stupid stupid stupid stupid stupid stupid stupid} $\{\iv{u_j(t,\intparams)}\,:\, j \in \Nq\} = \texttt{IRNEA}(\qA(t), \iv{\Delta}, a_0^0)$}
\label{alg:IRNEA}
\end{algorithm}
The Interval Recursive Newton Euler Algorithm (IRNEA), introduced in \cite{giusti2017efficient}, can be applied to compute the the nominal and robust inputs.
IRNEA is described in Alg. \ref{alg:IRNEA}, where we have suppressed the dependence on $q_j(t)$ in $\nom{R\jssmu}$ and  $p\jssmu$  for convenience.
% in Algorithm \ref{alg:IRNEA}.
The primary distinction between the classical Recursive Newton Euler Algorithm (RNEA) and IRNEA is that the latter takes in an interval set of inertial parameters and performs interval arithmetic in place of standard arithmetic. 
% However, note that the first and third arguments to IRNEA are vectors and not intervals.
% The nominal input $\tau$ can be directly computed using a modified Recursive Newton-Euler Algorithm (RNEA) \cite{giusti2017efficient}, detailed in Algorithm \ref{alg:RNEA}:
One can apply IRNEA to compute $\tau$:
\begin{equation}
    \tau(\qA(t),\nomparams) = \texttt{IRNEA}(\qA(t), \nomparams, a_0^0)
\end{equation}
where $a_0^0 = (0, 0, 9.81)\trans$ accounts for the effect of gravity.
By passing the nominal parameters $\nomparams$ (i.e., a degenerate interval) to IRNEA, the interval arithmetic in Alg. \ref{alg:IRNEA} become standard arithmetic and IRNEA is identical to RNEA.
% Note that throughout Algorithm \ref{alg:RNEA}, we have suppressed the dependence on $q_j(t)$ in $\nom{R\jssmu}$ for convenience. 
Computing $\robv$ requires computing bounds on the worst-case disturbance $\wmax(\qA(t), \nomparams,\intparams)$  and on the function $\robhmin$.
% The Interval RNEA (IRNEA) Algorithm, introduced in \cite{giusti2017efficient}, provides a means to efficiently compute these terms as described in Algorithm \ref{alg:IRNEA}, where we have again suppressed the dependence on $q_j(t)$ in $\nom{R\jssmu}$ for convenience.
% % in Algorithm \ref{alg:IRNEA}.
% The primary distinction between RNEA and IRNEA is that the latter takes in an interval set of inertial parameters and performs interval arithmetic operations in place of standard arithmetic operations. 
% However, note that the first arguments to both algorithms are vectors and not intervals.
The following lemma, whose proof can be found in App. \ref{app:compute_input}, describes how to use Alg. \ref{alg:IRNEA} to compute $\wmax$ and a $\underline{h}$ to satisfy the requirements of Thm. \ref{thm:robust_controller} and thereby compute the robust passivity-based controller.
\begin{lem}[Computing the Robust Input]
\label{lem:compute_input}
 If the interval disturbance in \eqref{eq:disturbance_equation} is computed as:
 \begin{equation}
    \label{eq:interval_disturbance}
     \iv{w(\qA(t),\nomparams,\intparams)} = \texttt{IRNEA}(\qA(t), \intparams, a_0^0) - \tau(\qA(t), \nomparams),
 \end{equation}
 and we define $\wmax$ using the previous equation as:
%  \begin{equation}
%      \norm{\wmax(\qA(t), \nomparams,\intparams)} = \norm{{\rho([\wdistinterval(\qA(t), \nomparams,\intparams)]}},
%  \end{equation}
 \begin{equation}
    \label{eq:interval_w_max}
     \wmax(\qA(t), \nomparams,\intparams) := {\rho([\wdistinterval(\qA(t), \nomparams,\intparams)]},
 \end{equation}
 where $\rho$ is as in Def. \ref{def:worst_case_disturbance}, then $\wmax$ satisfies \eqref{eq:cond_wmax} and is continuous in its first argument.
 Similarly, let
\begin{equation}
    % \iv{V(\qA(t),\intparams)} = \frac{1}{2} r(t)\trans \texttt{IRNEA}(q_R(t), \intparams, \zeros),
    \iv{V(\qA(t),\intparams)} = \frac{1}{2} r(t)\trans \iv{M(q(t), \intparams) r(t)}.
\end{equation}
The term $\iv{M(q(t), \intparams) r(t)}$ can be computed as
\begin{equation}
\iv{M(q(t), \intparams) r(t) } = \texttt{IRNEA}(q_R(t), \intparams, \zeros)
\end{equation}
where $q_R(t)$ is defined as
\begin{equation}
q_R(t) = \begin{bmatrix}
    q(t)\trans & \zeros\trans & q(t)\trans & \zeros\trans & r(t)\trans,
\end{bmatrix}\trans
\end{equation}
% is a vector of zeros, except in its first and fourth element which are equal to $q(t)$ and in its last element which is equal to $r(t)$, 
and the base acceleration is set to $\zeros$.
If $\underline{h}$  is defined as:
\begin{equation}\label{eq:robhmin_compute}
    \robhmin = -(\setop{sup} (\iv{V(\qA(t,\intparams)})) + \roblevel,
\end{equation}
 then it satisfies \eqref{eq:CBF_lower_bound} and is continuous in its first argument. 
\end{lem}
\section{Planning Algorithm Formulation} 
\label{sec:algorithm_theory}

% \subsection{Polynomial Zonotope Constraint Theory}
% \label{sec:pz_constraint_theory}
One of the key ideas of this work is to generate polynomial zonotope overapproximations of the trajectory and all constraints in the optimization problem \eqref{eq:optcost}-\eqref{eq:optcolcon} at the start of the planning iteration, then use these constraint overapproximations to perform optimization within $t\plan$.
Our approach guarantees safety while also ensuring:
\begin{enumerate}
    \item \textbf{Speed:} the size of all polynomial zonotope constraints can be fixed \textit{a priori}, ensuring fast constraint evaluation, and analytical gradients are generated via polynomial differentiation for use in optimization.
    % \item \textbf{Differentiability:} analytical gradients are generated via polynomial differentiation for use in optimization.
    % \item \textbf{Dependencies:} dependencies between inertial parameters (such as the proportional coupling between a link's mass and its inertia matrix) are lost during interval arithmetic operations, but can be explicitly modeled using polynomial zonotopes.
    \item \textbf{Continuous time and tracking error intervals:} each constraint is satisfied over the entire trajectory (including tracking error) for all time $t \in T$ rather than just at discretized time instances.
    \item \textbf{Physical consistency:} dependencies between uncertain inertial parameters can be modeled explicitly using polynomial zonotopes.
\end{enumerate}
\noindent This section presents the theory to transform the optimization problem \eqref{eq:optcost}-\eqref{eq:optcolcon} into an implementable version whose solutions can be be followed by the robot in a dynamically feasible and collision free fashion. 

\subsection{Polynomial Zonotope Trajectory Representation}
\label{subsec:pz_traj_rep}

\begin{figure*}[t]
    \centering
    \includegraphics[width=0.97\textwidth]{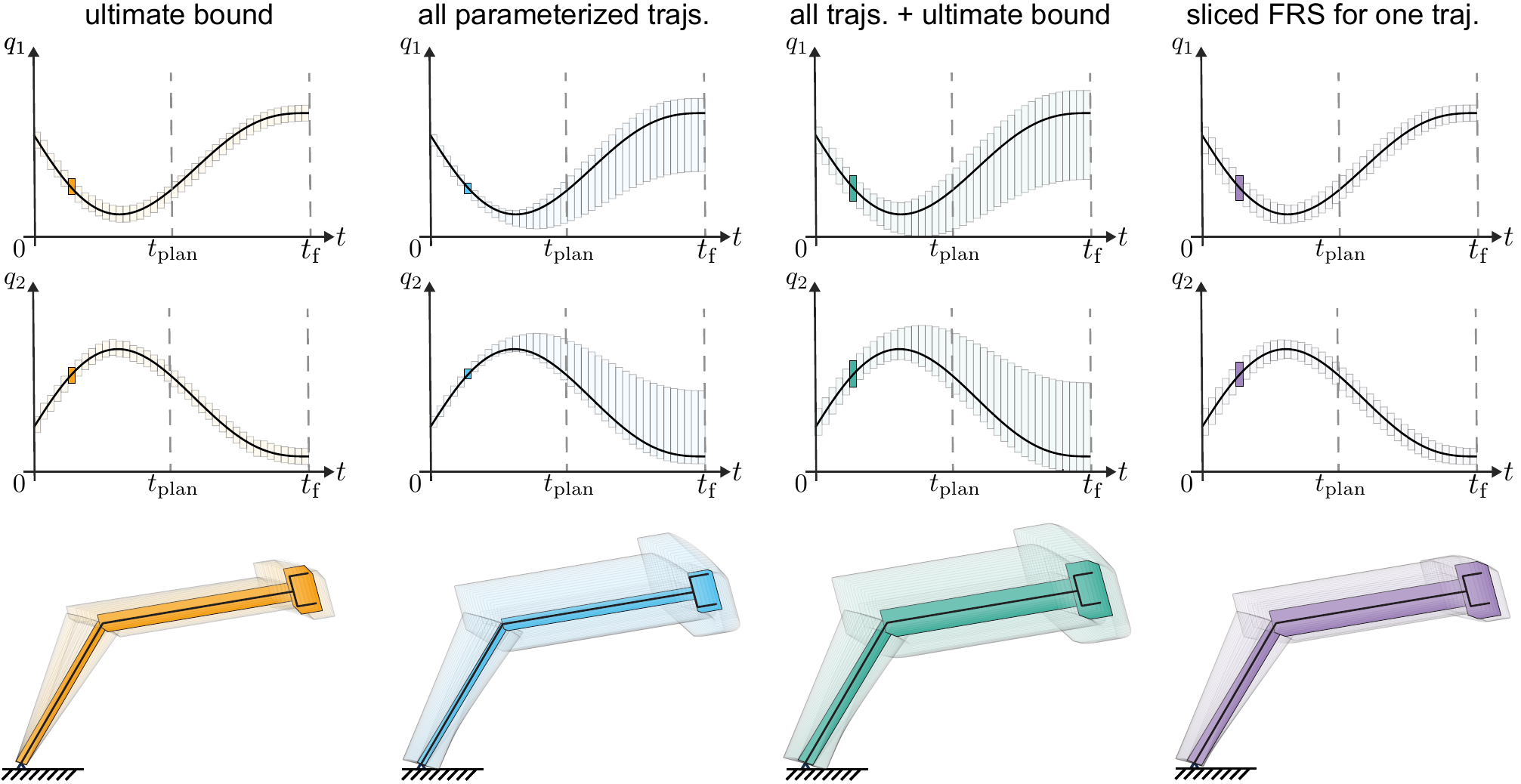}
    \caption{ 
    A visualization of how the polynomial zonotope trajectory representation is used to construct the polynomial zonotope forward occupancy for a two link arm in 2D. 
    A desired trajectory for each joint is shown in black in the top two rows.
    The same desired trajectory is depicted in each column.
    The planning time horizon is partitioned into a finite set of polynomial zonotopes (Sec. \ref{subsubsec:time_horizon_PZs}).
    In each column, a single time interval is highlighted in the top two rows and the corresponding time instance is highlighted in the visualization in the bottom row. 
    The top two rows of the first column depicts a finite set of polynomial zonotopes generated by uniformly buffering a desired trajectory by the ultimate bound (Lem. \ref{lem:pz_q_bound}).
    The bottom row of the first column depicts the forward occupancy of the robot over this entire set (Lem. \ref{lem:pz_fo}).
    The second column depicts a family of desired trajectories that are overapproximated by a finite set of polynomial zonotopes and the corresponding forward occupancy of the robot over this entire set.
    The third column depicts a family of desired trajectories buffered by the ultimate bound that are overapproximated by a finite set of polynomial zonotopes and the corresponding forward occupancy of the robot over this entire set.
    The fourth column depicts a subset of the sets illustrated in the third column generated by slicing in a specific trajectory parameter (Sec. \ref{subsubsec:pz_slice}). 
    Note that the desired trajectory always remains within the polynomial zonotope overapproximations.}
    \label{fig:example_polyzono}
\end{figure*}

% \shrey{figure suggestion: show the unsliced and sliced forward occupancy PZs for a given initial condition and trajectory overlaid on the arm (this could actually be merged with the suggestion for Fig. 1).
% Maybe it is even possible to show the safe and unsafe subsets of the parameter space, perhaps just projected into two of the joint dimensions?
% In other words, a figure to show the reachable sets and the constraints.}
% \Ram{we need a killer figure here that describes how PZ-ing works on time, $K$, desired trajectories and how it behaves with the epsilon from the uniform bound...}

As illustrated in Fig. \ref{fig:example_polyzono}, this subsection describes how \methodname overapproximates parameterized desired trajectories, as in Def. \ref{def:traj_param}, to conservatively account for continuous time and tracking error.
% Both time and tracking error are drawn from continuous, bounded intervals.
% Our approach to overapproximate trajectories of the robot applies polynomial zonotope arithmetic.
Because of Rem. \ref{rem:combining_traj}, the initial condition of the desired trajectories at each planning iteration is known, so the desired trajectories are functions of time and the trajectory parameter.
% before any optimization problem needs to be solved.
% As a result, 
We create polynomial zonotope versions of $T$ and $K$ that are then plugged into the desired trajectory to create polynomial zonotopes representing the desired trajectories.

\subsubsection{Time Horizon PZs}
\label{subsubsec:time_horizon_PZs}
We first create polynomial zonotopes representing the time horizon $T$.
Choose a timestep $\timestep$ so that $\nt \coloneqq \frac{T}{\timestep} \in \N$.
Let $N_t := \{1,\ldots,\nt\}$.
Divide the compact time horizon $T \subset \R$ into $\nt$ time subintervals.
% For example, if $T = [0, 1]$ and $\timestep = 0.01$, then the time horizon is divided into $\nt = 100$ subintervals.
Consider the $i$\ts{th} time subinterval corresponding to $t \in \iv{(i-1)\timestep, i\timestep}$.
We represent this subinterval as a polynomial zonotope $\pz{T_i}$, where 
\begin{equation}
    \label{eq:time_pz}
    \pz{T_i} =
        \left\{t \in T \mid 
            t = \tfrac{(i-1) + i}{2}\timestep + \tfrac{1}{2} \timestep \tvari,\ \tvari \in [-1,1]
        \right\}
\end{equation}
with indeterminate $\tvari \in \iv{-1, 1}$.

\subsubsection{Trajectory Parameter PZs}
\label{subsubsec:trajectory_PZs}
Next, we describe how to create polynomial zonotopes representing the set of trajectory parameters $K$.
This work chooses $K = \bigtimes_{i=1}^{n_q} K_i$, where each $\Kj$ is a compact one-dimensional interval.
For simplicity, we let each $\Kj = \iv{-1, 1}$.
% and use $\kjscale$ to scale the range of accelerations that this interval corresponds to.
% In Sec. \ref{sec:algorithm_theory} \pat{check section is correct}
% To create polynomial zonotopes representing trajectories, we consider the polynomial zonotope representation $\pz{\Kj}$, which we now define.
% \shrey{I think this definition needs to be more clearly motivated by the use case that we create a set of trajectory parameters, and then need to optimize over that set. Magically (as will be explained later), evaluating cost/constraints on the decision variable $k$ is equivalent to slicing.}
% \Ram{also this isn't really a definition, but rather a modeling choice? Unless the definition is really the slice function, but that was defined earlier? }
% \pat{I removed the definition environment, and wrapped this more in this section's text. Before this was in the Sec. \ref{sec:trajectory}, but I think it fits better here.}
% \begin{defn}
% \label{def:pz_Kj}
We represent the interval $\Kj$ as a polynomial zonotope $\pz{\Kj} = \kjvar$ where $\kjvar \in \iv{-1, 1}$ is an indeterminate.
We let $\kvar \in \iv{-1,1}^{n_q}$ denote the vector of indeterminates where the $j$\ts{th} component of $\kvar$ is $\kjvar$.
With this choice of $\Kj$, any particular $k_j$ yields $k_j = \setop{slice}(\pz{\Kj}, \kjvar, k_j)$ (see \eqref{eq:pz_slice}).
% Plugging in $k_j$ to the indeterminate $\kjvar$ yields the subset of $\pz{\Kj}$, which is $k_j$ itself.
% The process of ``slicing'' a polynomial zonotope by a trajectory parameter is critical to our formulation of the constraints within \methodname.
% For this reason, we give this process the following special notation.
% Letting $\kvar = (x_{k_1}, x_{k_2}, ..., x_{k_{\nq}})^\top$, and $\pz{P}$ some polynomial zonotope, we use the shorthand $\pzk{P}= \setop{slice}(\pz{P}, \kvar, k)$.
% This allows us to obtain subsets of polynomial zonotopes corresponding to a choice of $k$ by slicing the corresponding $\kvar$ indeterminates.
% \begin{defn}
% Letting $\kvar = (x_{k_1}, x_{k_2}, ..., x_{k_{\nq}})^\top$, and $\pz{P}$ some polynomial zonotope, we introduce the shorter notation $\pzk{P}= \setop{slice}(\pz{P}, \kvar, k)$.
% This allows us to obtain subsets of polynomial zonotopes corresponding to a choice of $k$ by slicing the corresponding $\kvar$ indeterminates.
% \end{defn}

% \pat{i'm going to describe the general idea for how to get these objects here... we can speak about the actual implementation details (offline computation, slicing by initial velocity, rotating by actual initial condition, etc.) in an "implementation details" section!}
% \jon{this paragraph is now out of order}
% \fix{fixed}
\subsubsection{Desired Trajectory PZs}
% The desired position, velocity, and acceleration trajectories of the robot, defined in Def. \ref{def:traj_param}, are functions of both time $t$ and the trajectory parameter $k$.
Next, we create polynomial zonotopes $\pzqdesji$ that overapproximate $\qdeskj$ for all $t$ in the $i$\ts{th} time subinterval and $k \in K$ by plugging the polynomial zonotopes $\pz{T_i}$ and $\pz{K}$ into $\qdeskj$ (see second column of Fig. \ref{fig:example_polyzono}).

% \shrey{Yeah the way this is currently written really confuses me.
% For example, I am confused by the phrase ``the function's arguments,'' since none of the previous objects that were discussed seem to be functions.
% I think this is trying to say that (1) we can consider the function $q_{d,j}: T\times K \to Q$ as taking set-valued inputs, (2) we represent the inputs sets $T$ and $K$ as polynomial zonotopes, and (3) the evaluation $\pzi{Q_{d,j}} = q_{d,j}(\pzi{T},\pz{K})$ is such that, for any $(t,k) \in T\times K$, $q_{d,j}(t;k) \in \pzi{Q_{d,j}}$ where I made up notation for this last bit.}
% \shrey{Also, I recommend avoiding macros like ``\textbackslash qdeskj'' for $\qdeskj$, where the function $q_{\regtext{whatever}}$ has already been evaluated at the point $(t;k)$ inside the macro, making the LaTeX code unreadable.\newcommand{\qdesexample}[2]{q_{d,#1}\left(#2\right)} Consider instead the example macro I put in the code right here for $\qdesexample{j}{t;k}$.}

\begin{lem}[Desired Trajectory PZs]
\label{lem:pz_desired_trajectory}
The desired trajectory polynomial zonotopes $\pzqdesji$ are overapproximative, i.e., for each $j \in \Nq$,
\begin{equation}
    \qdeskj \in \pzqdesji, \quad \forall t \in \pz{T_i}, \, k \in \pz{K}.
\end{equation}
One can similarly define $\pzqddesji$, and $\pzqdddesji$ that are overapproximative.
\end{lem}
\begin{proof}
$\qdesj$ is an analytic function that depends only on $t$ and $k$, which are included in $\pz{T_i}$ and $\pz{K}$.
All operations involving polynomial zonotopes are either exact or overapproximative for analytic functions \cite[Sec. 2.D]{kochdumper2020sparse}, so $\pz{T_i}$ and $\pz{K}$ in place of $t$ and $k$ yields an overapproximative polynomial zonotope.
\end{proof}

To illustrate how the polynomial zonotope representation works, consider the following example:
\begin{ex}
\label{ex:bernstein}
Suppose we parameterize the desired trajectories using degree 5 Bernstein polynomials.
Letting $T = [0, 1]$, the Bernstein polynomials take the form
\begin{equation}
    \label{eq:q_bernstein_polynomial}
    \qdeskj = \sum_{l= 0}^{5} \beta_{j, l}(k) b_{j, l}(t),
\end{equation}
where $\beta_{j, l}(k) \in \R$ are the Bernstein Coefficients and $b_{j, l}: T \to \R$ are the Bernstein Basis Polynomials of degree 5 given by
\begin{equation}
    \label{eq:bernstein_basis_polynomial}
    b_{j, l}(t) = {5\choose l}t^{l}(1 - t)^{5 -l},
\end{equation}
for each $l \in \{0, \ldots, 5\}$.
From Def. \ref{def:traj_param} the first and second properties of trajectory parameters constrain the initial position, velocity, and acceleration of the trajectory to match the initial condition, and require the final velocity and acceleration to be $0$.
These properties fix five of the six Bernstein coefficients $\beta_{j, \nu}$ as $\beta_{j, 0}(k) = \initqj$, $\beta_{j, 1}(k) = \tfrac{1}{5}(\initdqj + 5\beta_{j, 0})$,$\beta_{j, 2}(k) = \tfrac{1}{20}(\initddqj + 40\beta_{j, 1} - 20\beta_{j, 0})$, $\beta_{j, 3}(k) = \tfrac{1}{20}(0 + 40\beta_{j, 4} - 20\beta_{j, 5})$, and $\beta_{j, 4}(k) = \tfrac{1}{5}(0 + 5\beta_{j, 5})$.
% \begin{align}
%     % \label{eq:bernstein_coeff}
%     \beta_{j, 0}(k) &= \initqj \\
%     \beta_{j, 1}(k) &= \tfrac{1}{5}(\initdqj + 5\beta_{j, 0}) \\
%     \beta_{j, 2}(k) &= \tfrac{1}{20}(\initddqj + 40\beta_{j, 1} - 20\beta_{j, 0})\\
%     \beta_{j, 3}(k) &= \tfrac{1}{20}(0 + 40\beta_{j, 4} - 20\beta_{j, 5})\\
%     \beta_{j, 4}(k) &= \tfrac{1}{5}(0 + 5\beta_{j, 5}).
% \end{align}
Because the first five Bernstein coefficients are fixed, we drop the dependence on $k$ for these coefficients.
We choose the last coefficient as
\begin{equation}
    \label{eq:bernstein_k_coeff}
    \beta_{j, 5}(k) = \kjscale \kj + \kjoffset
\end{equation}
where $\kjscale$ and $\kjoffset \in \R$ are user-specified constants.
The coefficient $\beta_{j, 5}$ equals the final position $q_{d,j}(1; k)$ at time $t = 1$, so the choice of trajectory parameter $k_j$ determines this value.
% Note that with this formulation, $\beta_{j, 3} = \beta_{j, 4} = \beta_{j, 5}$.

% Similarly, we may write the interval $\Kj = \iv{-1, 1}$ as a polynomial zonotope $\pz{\Kj} = k_j$ with $k_j \in \iv{-1, 1}$.
Construct the polynomial zonotope representation $\pzqdesji$ by plugging in $\pz{T_i}$ from \eqref{eq:time_pz} and $\pz{\Kj}$.
Letting $\pzgreek{\beta}_\pz{j, 5}(\pz{K}) = \kjscale \pz{\Kj} + \kjoffset$, and plugging $\pz{T_i}$ into the basis polynomials \eqref{eq:bernstein_basis_polynomial}, we obtain
\begin{equation}
    \pzqdesji = \left( \sum_{l = 0}^{4} \beta_{j, l} \pz{b_{j, l}}(\pz{T_i}) \right) \oplus \bm{\beta}_\mathbf{j, 5}(\pz{K})\pz{b_{j, 5}}(\pz{T_i}),
\end{equation}
The desired velocity and acceleration polynomial zonotopes $\pzqddesji$ and $\pzqdddesji$ can be similarly computed.
% These desired trajectory polynomial zonotopes are combined with the tracking error polynomial zonotopes to get $\pzqi$ and $\pzqdi$ as in \eqref{eq:pz_pos} and \eqref{eq:pz_vel}, as well as $\pzqdai$ and $\pzqddai$ as in \eqref{eq:pz_mod_traj}.
% To summarize, by specifying the initial condition of the desired trajectory, i.e., $(\initq, \initdq, \initddq)$, one can construct the polynomial zonotope representation of the desired trajectory and all of its variants.
\end{ex}

% \noindent
% Details on the creation of these objects are provided in Sec. \ref{sec:implementation_trajectories}.
% \shrey{``these objects'' is vague}
% These can be acquired using polynomial zonotope arithmetic.

\subsubsection{Slicing Yields Desired Trajectories}
\label{subsubsec:pz_slice}
$\pz{T_i}$ and $\pz{K_j}$ have indeterminates $\tvari$ and $\kjvar$ respectively.
Because the desired trajectories only depend on $t$ and $k$, the polynomial zonotopes $\pzqdesji$, $\pzqddesji$ and $\pzqdddesji$ depend only on the indeterminates $\tvari$ and $\kvar$.
By plugging in a given $k$ for $\kvar$ via the $\setop{slice}$ operation, we obtain a polynomial zonotope where $\tvari$ is the only remaining indeterminate. 
Because we perform this slicing operation throughout this document, we use the shorthand $\pzqdesjki = \setop{slice}(\pzqdesji, \kvar, k)$ for a given polynomial zonotope, $\pzqdesji$.
Because of our previous observation, $\qdeskj \in \pzqdesjki$ for all $t \in \pz{T_i}$ (and similarly for $\pzqddesjki$ and $\pzqdddesjki$).
% Note that $\qdeskj \in \setop{slice}{\pzqdesji, $
% is well-defined.
% Setting $\pzg_0 = \initqj + \initdqj \frac{1}{2} \timestep$, $\pzg_1 = \frac{1}{2} \frac{1}{4} \timestep ^2$ with $\pzi_1 = k_j$, $\pzg_2 = \initdqj \frac{1}{2} \timestep$, and $\pzg_3 = \frac{1}{2} 

\subsubsection{Incorporating Tracking Error}
Recall from Sec. \ref{sec:controller_design} that desired trajectories may not be tracked perfectly.
% However, by Thm. \ref{thm:robust_controller} and Cor. \ref{cor:tracking_error}, the controller \eqref{eq:controller} ensures that the tracking error is uniformly ultimately bounded when applying the robust input given by \eqref{eq:robust_input}.
% From Cor. \ref{cor:tracking_error}, at each time $t$ the position error of each joint $\errj(t)$ is bounded by $\pboundj$, and the velocity error $\errjdot(t)$ is bounded by $\vbound$.
% We use this to overapproximate any trajectory followed by the robot.
As depicted in the first and third column of Fig. \ref{fig:example_polyzono}, by applying Cor. \ref{cor:tracking_error} and Lem. \ref{lem:pz_desired_trajectory}, we can overapproximate any trajectory followed by the robot:
\begin{lem}[Error in Config. Space]
\label{lem:pz_q_bound}
Let $\pzpboundj = \pboundj\epvarj$ and $\pzvboundj = \vbound\evvarj$, with indeterminates $\epvarj \in \iv{-1, 1}$ and $\evvarj \in \iv{-1, 1}$.
Then, for each $i \in \Nt$ let
\begin{align}
    \pzqji &= \pzqdesji \oplus \pzpboundj \label{eq:pz_pos} \\
    \pzqdji &= \pzqddesji \oplus \pzvboundj \label{eq:pz_vel}.
\end{align}
Given the set of trajectory parameters $\pz{K}$, the polynomial zonotopes $\pzqji$ and $\pzqdji$ overapproximate the set of all joint trajectories that can possibly be executed by the robot, i.e., for each $j \in \Nq$,  $k \in \pz{K}$ we have
\begin{align}
    q_j(t; k) \in \pzqjki, \quad \forall t \in \pz{T_i} \\
    \dot{q}_j(t; k) \in \pzqdjki, \quad \forall t \in \pz{T_i}.
\end{align}
\end{lem}
% \begin{proof}
% By Lem. \ref{lem:pz_desired_trajectory}, the desired trajectory polynomial zonotopes contain all desired trajectories.
% By Cor. \ref{cor:tracking_error}, the position and velocity trajectories differ from the desired trajectories by at most $\pm \pbound$ and $\pm \vbound$, respectively.
% Therefore the Minkowski sum of $\pzqdesji \oplus \pz{\pboundj}$ must contain the actual configuration $q_j(t; k)$, and similarly for velocities.
% \end{proof}
\noindent Moving forward, let $\pzqdesi = \bigtimes_{j=1}^{n_q} \pzi{q_{d, j}},$ and similarly define $\pzqi, \pzqddesi$, $\pzqdi$, $\pzqdddesi$, and $\pzqddi$.
Furthermore, let $\pz{\pboundvec} = \bigtimes_{j=1}^{n_q} \pzpboundj,$ and similarly define $\pz{\vboundvec}$ using $\pzvboundj$.

\subsection{Reachable Set Construction}
\label{sec:reachset_construction}
Next, \methodname bounds the kinematic behavior of the arm and the possible torques required to track a trajectory.
% We formulate algorithms which generate polynomial zonotopes (parameterized by $k$) describing the volume occupied by the arm and the motor torques required to track a given trajectory.
These bounds are then used to formulate constraints on $k$ that overapproximate those in \eqref{eq:optpos} - \eqref{eq:optcolcon}.
% In the remainder of this subsection, we describe how to compute reachable sets for a single joint/link over a single interval of time.

\subsubsection{Forward Kinematics Reachable Set}
% When planning certain trajectories (e.g. manipulating a coffee mug), we must constrain the pose of one or more frames on the robot during each planning iteration.
% To accomplish this, we 
We represent the robot's forward kinematics \eqref{eq:fk_j} using polynomial zonotope overapproximations of the joint position trajectories.
Using $\pzqji$, we compute $\pz{p\jssmu}(\pzqji)$ and $\pz{R\jssmu}(\pzqji)$, which represent overapproximations of the position and orientation of the $j$\ts{th} frame with respect to frame $(j-1)$ at the $i$\ts{th} time step.
The rotation matrix $R\jssmu(q_j(t;k))$ depends on $\cos{(q_j(t;k))}$ and $\sin{(q_j(t;k))}$.
As a result, we must compute $\cos(\pzqji)$ and $\sin(\pzqji)$ to compute $\pz{R\jssmu}(\pzqji)$.
Cosine and sine are analytic functions, so we compute  $\cos{(\pzqji)}$ and  $\sin{(\pzqji)}$ using Taylor expansions as in  Sec. \ref{subsec:polyzono}.
% In practice, only a finite degree Taylor expansion can be constructed and the Lagrange Remainder as described in Sec. \ref{subsec:polyzono} is used to conservatively bound the truncation error. 
% can be used as we describe in Sec. \ref{sec:implementation}.
Because all operations involving polynomial zonotopes are either exact or overapproximative, the polynomial zonotope forward kinematics can be computed similarly to the classical formulation \eqref{eq:fk_j} and proven to be overapproximative:
\begin{lem}[PZ Forward Kinematics]
\label{lem:PZFK}
Let the polynomial zonotope forward kinematics for the $j$\ts{th} frame at the $i$\ts{th} time step be defined as
\begin{equation}\label{eq:pzfk_j}
    \pzFKjKi = 
    \begin{bmatrix} \pz{R_j}(\pzqi) & \pz{p_j}(\pzqi)\\
    \mathbf{0}  & 1 \\
    \end{bmatrix},
\end{equation}
where
\begin{align}
    \pz{R_j}(\pzqi) &= \prod_{l=1}^{j} \pz{R_{l}^{l-1}}(\pzqli), \\
    \pz{p_j}(\pzqi) &= \sum_{l=1}^{j} \pz{R_l}(\pzqi) \pz{p_{l}^{l-1}},
\end{align}
then for each  $j \in \Nq$,  $k \in \pz{K}$,  $\FK_j(q(t;k)) \in  \pzFKjki$ for all $t \in \pz{T_i}$.
\end{lem}
% Finally, the forward kinematics reachable set for a single planning iteration is given by the union
% \begin{equation}\label{eq:pzfk}
%     \mathbf{FK} \subseteq \bigcup_{i=1}^{\nt} \bigcup_{j=1}^{\nq} \pzi{\FK_j}.
% \end{equation}
\noindent The Polynomial Zonotope Forward Kinematics Algorithm (PZFK), detailed in Alg. \ref{alg:compose_fk}, is used to compute the objects in Lem. \ref{lem:PZFK} from the set of possible configurations $\pzqi$.
%          %~~~~~~~~~~~~ Compose FK
% \begin{algorithm}[t]
% \small
% \jon{define pzHomTrans}
% \begin{algorithmic}[1]

% \State{\bf parfor} $i = 1:\nt$ %// parallel for each time step

% \State\hspace{0.2in}{\bf for} $j = 1:\nq$ %// for each joint
%     \State\hspace{0.4in}$\pz{p_j}(\pzqi) \leftarrow \pz{\mathbf{0}}$ %// initialize frame location 
        
%     \State\hspace{0.4in}$ \pz{R_j}(\pzqi) \leftarrow \pz{I_{3 \times 3}}$ %// initialize frame orientation
    
%      \State\hspace{0.4in}{\bf for} $l = 1:j$

%         \State\hspace{0.6in}$\pz{R_{l}^{l-1}}(\pzqli), \pz{\nom{p\lssmu}} \leftarrow \texttt{pzTransMat}(\pzi{q_j})$ 
    
%         \State\hspace{0.6in}$\pz{p_j}(\pzqi) \leftarrow \pz{p_j}(\pzqi) \oplus \pz{R_j}(\pzqi)  \odot \pz{p_{l}^{l-1}}$
        
%         \State\hspace{0.6in}$ \pz{R_j}(\pzqi) \leftarrow  \pz{R_j}(\pzqi) \odot  \pz{R_{l}^{l-1}}(\pzqli)$ 

%     \State\hspace{0.4in}{\bf end for}
    
%     \State\hspace{0.4in}$\pz{\FK_j}(\pzqi) \leftarrow \{ \pz{R_j}(\pzqi) ,  \pz{p_j}(\pzqi) \}$ 

% \State\hspace{0.2in}{\bf end for}

% \State{\bf end parfor}

% \end{algorithmic}
% \caption{\small $\{\pz{\FK_j}(\pzqi)\,:\, j \in \Nq,\ i \in \Nt \} = \texttt{PZFK}(\{ \pzi{q_j} \,:\, j \in \Nq,\ i \in \Nt \})$}
% \label{alg:compose_fk}
% \end{algorithm}

%~~~~~~~~~~~~ Compose FK
\begin{algorithm}[t]
\small
\begin{algorithmic}[1]

% \State{\bf parfor} $i = 1:\nt$ %// parallel for each time step

\State $\pz{p_0}(\pzqi) \leftarrow \zeros$ %// initialize frame location 

\State$ \pz{R_0}(\pzqi) \leftarrow I_{3 \times 3}$ %// initialize frame orientation

\State\hspace{0in}{\bf for} $j = 1:\nq$ %// for each joint

    % \State\hspace{0.2in}$\pz{R_{j}^{j-1}}(\pzqji), \pz{\nom{p\jssmu}} \leftarrow \texttt{pzTransMat}(\pzi{q_j})$ 
    
    \State\hspace{0.2in}$\pz{R_{j}^{j-1}}(\pzqji), \pz{\nom{p\jssmu}} \leftarrow \pz{\homtrans_{j}^{j-1}}(\pzqji)$ as in \eqref{eq:homogeneous_transform}
    
    \State\hspace{0.2in}$\pz{p_j}(\pzqi) \leftarrow \pz{p_{j-1}}(\pzqi) \oplus \pz{R_{j-1}}(\pzqi) \pz{p_{j}^{j-1}}$
        
    \State\hspace{0.2in}$ \pz{R_j}(\pzqi) \leftarrow  \pz{R_{j-1}}(\pzqi)  \pz{R_{j}^{j-1}}(\pzqji)$ 
    
    \State\hspace{0.2in}$\pz{\FK_j}(\pzqi) \leftarrow \{ \pz{R_j}(\pzqi) ,  \pz{p_j}(\pzqi) \}$ 
    
\State\hspace{0in}{\bf end for}

% \State{\bf end parfor}

\end{algorithmic}
\caption{\small $\{\pz{\FK_j}(\pzqi)\,:\, j \in \Nq \} = \texttt{PZFK}(\pzqi)$}
\label{alg:compose_fk}
\end{algorithm}

\subsubsection{Forward Occupancy Reachable Set}
To ensure that the robot never collides with any obstacles, we overapproximate the forward occupancy \eqref{eq:forward_occupancy_j} of each link over every time interval.
By applying the polynomial zonotope forward kinematics set and the fact that all operations involving polynomial zonotopes are either exact or overapproximative, the polynomial zonotope forward occupancy reachable set can be computed and proven to be overapproximative: 
\begin{lem}[PZ Forward Occupancy]
\label{lem:pz_fo}
Let the $j$\ts{th} link of the robot be overapproximated by a polynomial zonotope denoted by $\pz{L_j}$ and the polynomial zonotope forward occupancy reachable set for the $j$\ts{th} link at the $i$\ts{th} time step be defined as:
\begin{align}\label{eq:pz_forward_occupancy_j}
     \pzFOjKi = \pz{p_j}(\pzqi) \oplus \pz{R_j}(\pzqi)\pz{L_j},
\end{align}
then for each $j \in \Nq$,  $k \in \pz{K}$, $\FO_j(q(t;k)) \in  \pzFOjki$ for all $t \in \pz{T_i}$.
\end{lem}
\noindent Sec. \ref{subsec:constraint_generation} describes how the forward occupancy reachable set is used to compute collision-avoidance constraints.
% The Polynomial Zonotope Forward Occupancy Algorithm (PZFO), detailed in Alg. \ref{alg:compose_fo}, is identical to the PZFK algorithm but is applied to the volume of each link rather than joint positions.
Let
\begin{equation}
    \label{eq:pz_forward_occupancy}
    \pz{FO}(\pzqi) = \bigcup_{j = 1}^{\nq} \pzFOjKi.
\end{equation}
% \input{algorithms/pzfo}

% We start by overapproximating the $j$\ts{th} link of the robot by a polynomial zonotope denoted by $\pz{L_j}$.
% Next, building off the forward kinematics reachable set \eqref{eq:pzfk_j}, the forward occupancy reachable set for the $j$\ts{th} link at the $i$\ts{th} time step is
% \begin{align}\label{eq:pz_forward_occupancy_j}
%      \pzi{\FO_j} = \pzi{p_j} \oplus \pzi{R_j}\pz{L_j}.
% \end{align}
% In Sec. \ref{sec:collision_avoidance_implementation}, we describe how the forward occupancy reachable set is used to compute collision-avoidance constraints.
% \begin{equation}\label{eq:pz_forward_occupancy}
%     \mathbb{FO} \subseteq \bigcup_{i=1}^{\nt} \bigcup_{j=1}^{\nq} \pzi{\FO_j}.
% \end{equation}
% \jon{do we want to distinguish between the actual reachable set and the overapproximation of the reachable set using polynomial zonotopes?}

\subsubsection{Input Reachable Set}
\label{subsubsec:theory_input_reach_set}
% As described in the previous section, the forward reachable set gives a continuum of trajectories that can be evaluated for satisfying safety constraints.
% But in order to execute a motion plan safely, the robot must be able to track a given trajectory without violating its torque limits. 
Mobile manipulators must also avoid saturating their available motor torques.
We describe how to overapproximate the set of torques that may be required for the robust passivity-based control input \eqref{eq:controller} to track any parameterized trajectory.
Just as we did in Sec. \ref{sec:controller_design}, we define a total feedback trajectory polynomial zonotope $\pzi{\qA}$ by computing a polynomial zonotope representation of each of the components of $\qA$.
% \begin{equation}
% \label{eq:pz_total_feedback}
% \pzi{\qA} = \\
% \begin{bmatrix} \pzqi \\ \pzqdi \\ \pzqdesi \\ \pzqddesi \\ \pzqdddesi \end{bmatrix}.
% \end{equation}
% by applying polynomial zonotope arithmetic to \eqref{eq:total_feedback_traj}. 
We similarly define modified desired velocity and acceleration trajectory polynomial zonotopes:
% Similar to 
\begin{equation} \label{eq:pz_mod_traj}
    \begin{split}
    \pzqdai &= \pzqdesi \oplus K_r \pz{\pboundvec}, \\
    \pzqddai &= \pzqddesi \oplus K_r \pz{\vboundvec}.
    \end{split}
\end{equation}
% \begin{equation}
%     \pzi{\tau} = \bM(\pzi{q}, \pz{\Delta}) \pzi{\ddot{q}_a} + \bC(\pzi{q}, \pzi{\dot{q}}, \pz{\Delta}) \pzi{\dot{q}_a} + \bG(\pzi{q}, \pz{\Delta})
% \end{equation}
Using these definitions and plugging into \eqref{eq:nominal_controller}, we obtain
\begin{equation}
    \label{eq:pz_nominal_input}
    \begin{split}
    % \pzi{\tau} = \pzi{\bM} \pzi{\ddot{q}_a} + \pzi{\bC} \pzi{\dot{q}_a} + \pzi{\bG}
    \bm{\tau}( \pzqAi, &\nomparams) = \pz{\bM}(\pzi{q}, \nomparams) \pzi{\ddot{q}_a} + \\ &+ \pz{\bC}(\pzi{q}, \pzi{\dot{q}},\nomparams) \pzi{\dot{q}_a} + \\ &+ \pz{\bG}(\pzi{q}, \nomparams).
    \end{split}
\end{equation}
% As written in \eqref{eq:pz_nominal_input}, the computation of the nominal input $\bm{\tau}(\pzqAi,\Delta_0)$ requires finding the mass matrix and Coriolis and gravitational terms.
% Instead, a more direct approach is to use a modified Recursive Newton-Euler Algorithm (RNEA) 
To compute $\bm{\tau}(\pzqAi,\Delta_0)$, we introduce the Polynomial Zonotope RNEA (PZRNEA), detailed in Alg. \ref{alg:PZRNEA}.
Note just as in the original RNEA and IRNEA Algorithms, we have suppressed the dependence on $\pzi{q_j}$ in $\pz{R_{j-1}^{j}}$.
PZRNEA takes as inputs $\pzqAi$, and the nominal inertial parameters $\nomparams$ to find the nominal input:
% \Ram{should this be PZ-ed?} \pat{I think we can go either way... The nominal params are just a vector, so converting that to a pz is trivial. For the interval params, I added \eqref{eq:int_to_pz} to show how to turn an interval to a PZ... so the conversions are there implicitly but idk if we should make it explicit} 
% Here we use polynomial zonotopes and a modified RNEA algorithm, called Polynomial Zonotope RNEA (PZRNEA), to construct input constraints that are satisfied over the entire duration of a parameterized trajectory.
\begin{align}
    \label{eq:pz_nominal_pzrnea}
   \bm{\tau}(\pzqAi,\Delta_0)= \setop{PZRNEA}(\pzqAi, \nomparams, a_0^0),
%   &+ \pzi{-\robcoeff \frac{\robr}{\norm{\robr}}} \\
%   &\jon{\text{how to express robust input?}} \\
\end{align}
where $a_0^0 = [0, 0, 9.81]^\top$.

\begin{algorithm}[t]
\small
\begin{algorithmic}[1]
\State Compute $\pzqdai$ and $\pzqddai$ using \eqref{eq:pz_mod_traj}
\State Initialize base acceleration $a_0^0$.
% \State{\bf parfor} $i = 1:\nt$ // parallel for each time step

% \State\hspace{0.2in}Compute $\pzi{\dot{q}_a}$ and $\pzi{\ddot{q}_a}$ using \eqref{eq:modified_ref}.
\State\hspace{0in}{\bf for} $j = 1:\nq$ %// for each joint
    
    \State\hspace{0.2in}$\pz{R\jssmu}, \pz{\nom{p\jssmu}} \leftarrow \pz{\homtrans_{j}^{j-1}}(\pzqji)$ 
    
    \State\hspace{0.2in} $\pz{R\jssm} \gets \texttt{pzTranspose}(\pz{R\jssmu})$
    
\State\hspace{0in}{\bf end for}

\State\hspace{0in}{\bf for} $j = 1:\nq$ %// for each joint (start of forward recursion)

    \State\hspace{0.2in} $\pz{\omega\jss} \gets \pz{R\jssm}  \pz{\omega\jssmm} \bigoplus \pz{\qdj}\hat{z}_j$
    
    \State\hspace{0.2in} $\pz{\omega\jssa} \gets \pz{R\jssm}  \pz{\omega\jssmma} \bigoplus \pz{\qajdot}\hat{z}_j$
    
    \State\hspace{0.2in} $\pz{\dot{\omega}\jss} \gets \pz{R\jssm}  \pz{\dot{\omega}\jssmm} \bigoplus ((\pz{R\jssm}  \pz{\omega\jssmma}) \bigotimes (\pz{\qdj} \hat{z}_j)) \bigoplus \pz{\ddot{q}_{a,j}}\hat{z}_j$
    
    % \State\hspace{0.2in} $\pz{a\jss} \gets (\pz{R\jssm}  \pz{a\jssmm}) \bigoplus (\pz{\dot{\omega}\jss} \bigotimes \pz{p\jssmj}) \bigoplus (\pz{\omega\jss} \bigotimes (\pz{\omega\jssa} \times \pz{p\jssmj}))$
    
        \State\hspace{0.2in} $\pz{a\jss} \gets (\pz{R\jssm}  \pz{a\jssmm}) \bigoplus (\pz{\dot{\omega}\jss} \bigotimes \pz{p\jssmu}) \bigoplus (\pz{\omega\jss} \bigotimes (\pz{\omega\jssa} \times \pz{p\jssmu}))$

    \State\hspace{0.2in} $\pz{a\jssc} \gets \pz{a\jss} \bigoplus (\pz{\dot{\omega}\jss} \bigotimes \pz{c\jss}) \bigoplus (\pz{\omega\jss} \bigotimes (\pz{\omega\jssa} \bigotimes \pz{c\jss})) $
    
    \State\hspace{0.2in} $\pz{F\jss} \gets \pz{m_i}  \pz{a\jssc}$
    
    \State\hspace{0.2in} $\pz{N\jss} \gets \pz{I\jss}  \pz{\dot{\omega}\jss} \bigoplus (\pz{\omega\jssa}  \bigotimes (\pz{I\jss}  \pz{\omega\jss})) $

\State\hspace{0in}{\bf end for}

% \State\hspace{0in}Initialize $\pz{R^{\nq}_{\nq+1}}$ \Ram{more details}
\State\hspace{0in} $\pz{R^{\nq}_{\nq+1}} \leftarrow \pz{I_{3 \times 3}}$
\State\hspace{0in}$\pz{f^{\nq+1}_{\nq+1}} \leftarrow \zeros$ 
\State\hspace{0in}$\pz{n^{\nq+1}_{\nq+1}} \leftarrow \zeros$ 

\State\hspace{0in}{\bf for} $j = \nq:-1:1$ %// for each joint (start of backward recursion)

\State\hspace{0.2in} $\pz{f\jss} \gets (\pz{R\jssp}  \pz{f^{j+1}_{j+1}}) \bigoplus \pz{F\jss} $

\State\hspace{0.2in} $\pz{n\jss} \gets (\pz{R\jssp}  \pz{n^{j+1}_{j+1}}) \bigoplus (\pz{c\jss} \bigotimes \pz{F\jss}) \bigoplus  \newline \hspace*{7em}  \bigoplus  (\pz{p\jssmj} \bigotimes (\pz{R\jssp}  \pz{f^{j+1}_{j+1}})) \bigoplus \pz{N\jss} $ 

\State\hspace{0.2in} $\pz{u}_j(\pzqAi, \intparams) \gets \hat{z}_j^\top \pz{n\jss}$

\State\hspace{0in}{\bf end for}

% \State{\bf end parfor}
\end{algorithmic}
\caption{\small \phantom{stupid stupid stupid stupid stupid stupid stupid} $\{ \pz{u}_j(\pzqAi, \intparams) \,:\, j \in \Nq\} = \texttt{PZRNEA}(\pzqAi, \intparams , a_0^0)$}
\label{alg:PZRNEA}
\end{algorithm}

Next, we describe a bound on the robust control input \eqref{eq:robust_input} at each time step whose proof can be found in App. \ref{app:robust_input_bound}.
\begin{thm}[Robust Input Bound]
\label{thm:robust_input_bound}
% Let $\pz{h}$ be defined as 
% \begin{equation}
% \begin{split}
%     \pz{h}(\pzi{\qA},\pzparams) = -\frac{1}{2} \pzi{r}^T &\pz{M}(\pzi{q},\pzparams) \pzi{r} + \\ &+ V_M
%     \end{split}
% \end{equation}
Suppose $\alpha: \R \to \R$ in \eqref{eq:robust_coeff} is a linear function, i.e., $\alpha(x) = \alpha_c x$,
% \begin{equation}
%     \label{eq:alpha_linear}
%     \alpha(x) = \alpha_c x
% \end{equation}
where $\alpha_c > 0$ is a user-specified constant.
% and note that $\alpha$ is an extended class $\mathcal{K}_\infty$ as required by Thm. \ref{thm:robust_controller}.
Suppose as in \eqref{eq:interval_disturbance}, the disturbance \eqref{eq:disturbance_equation} is overapproximated using
% \begin{equation}
% \begin{split}
%     \pzgreek{\tau}_\regtext{int}&(\pzi{\qA},\pzparams) = \pz{M}(\pzi{q}, \pzparams) \pzi{\ddot{q}_a} + \\ &+ \pz{C}(\pzi{q}, \pzi{\dot{q}},\pzparams) \pzi{\dot{q}_a} + \pz{G}(\pzi{q}, \pzparams),
%     \end{split}
% \end{equation}
% and using \eqref{eq:disturbance_equation} we obtain
\begin{equation}
    \label{eq:pz_disturbance}
\begin{split}
    % \pz{w}(\pzi{\qA},\nomparams,\pzparams) = \pzgreek{\tau}_{\regtext{int}}(\pzi{\qA},\pzparams) - \pzgreek{\tau}(\pzi{\qA},\nomparams).
    \pz{w}(\pzi{\qA},\nomparams,\intparams) = \setop{PZRNEA}(\pzi{\qA},\intparams, a_0^0) + \\ - \pzgreek{\tau}(\pzi{\qA},\nomparams)
\end{split}
\end{equation}
with $\setop{PZRNEA}$ as in Alg. \ref{alg:PZRNEA} and $a_0^0 = [0, 0, 9.81]^\top$.
Using Def. \ref{def:worst_case_disturbance}, let
\begin{equation}
\begin{split}
    \rho(\pz{w}(\pzi{\qA},\nomparams,\intparams)) &=
    \max\Big(
        \abs{\setop{inf}(\pz{w}(\pzi{\qA},\nomparams,\intparams))}, \\ 
    &\abs{
        \setop{sup}(\pz{w}(\pzi{\qA},\nomparams,\intparams))
        }
    \Big), 
    \end{split}
\end{equation}
and as in \eqref{eq:interval_w_max} let
\begin{equation}
\label{eq:pz_w_max}
\wmax(\pzgreek{*}) = \rho(\pz{w}(\pzi{\qA},\nomparams,\intparams))
\end{equation}
where $\wmax(\pzgreek{*}) := \wmax(\pzi{\qA},\nomparams,\intparams)$.
Then for all $\qA(t;k)$ that satisfies the uniform bound \eqref{eq:ultimate_bound}, the following bound is satisfied for each $j \in \Nq$ 
%     \begin{multline}
%     \label{eq:pz_robust_input_bound}
%   \abs{\robv(\qA(t;k), \nomparams, \pzparams )_j} \leq \max \Big\{ 0, \norm{\rho(*)}(\frac{1 + \rho(*)_j}{2}) + \\ + \frac{-\robKinf(\setop{inf} (\pz{h}(\pzi{\qA},\pzparams)))}{\ultbound} \Big\},
%   \end{multline}
\begin{equation}
\label{eq:pz_robust_input_bound}
\begin{split}
    \abs{\robv(\qA(t;k), \nomparams, \intparams )_j} \leq &\frac{\alpha_c \epsilon(\sigma_M - \sigma_m)}{2} + \\ &+ \frac{\norm{\wmax(\pzgreek{*})} + \wmax(\pzgreek{*})_j}{2},
    \end{split}
\end{equation}
where $\abs{\robv(\qA(t;k), \nomparams, \intparams )_j}$ is the $j$\ts{th} component of the robust input.
% where $\rho(*) = \rho(\pz{w}(\pzi{\qA},\nomparams,\pzparams))$ and $\rho(*)_j$ is its $j$-th component for brevity.
\end{thm}
\noindent This theorem requires that $\alpha$ is linear with positive slope rather than an arbitrary extended class $\mathcal{K}_\infty$ function. 
The proof could be extended to any $\mathcal{K}_\infty$ function, but would generate a bound that is more difficult to compute.
As we show in Sec. \ref{sec:experiments}, a linear $\alpha$ works well. 
Using \eqref{eq:pz_robust_input_bound}, we bound the robust input by a constant in each dimension.
That is, we let
% \begin{multline}
%     \label{eq:pz_robust_coeff}
%     \pzi{v_j} = \max \Big\{ 0, \norm{\rho(*)}(\frac{1 + \rho(*)_j}{2}) + \\ + \frac{-\robKinf(\setop{inf} (\pz{h}(\pzi{\qA},\pzparams)))}{\ultbound} \Big\} x_{v_{j}},
% \end{multline} 
\begin{equation}
    \label{eq:pz_robust_coeff}
    \begin{split}
    \pz{ v }_j(\pzqAi,\Delta_0,[\Delta]) = 
        \Big(&\frac{\alpha_c \epsilon(\sigma_M - \sigma_m)}{2} + \\
        &+ \frac{\norm{\wmax(\pzgreek{*})} + \wmax(\pzgreek{*})_j}{2}\Big)x_{v_{j}}
    \end{split}
\end{equation}
with $\wmax(\pzgreek{*})$ as in Thm. \ref{thm:robust_input_bound} and $x_{v_{j}}$ an indeterminate in $[-1, 1]$.
Using this we let
\begin{equation}
    \label{eq:pz_robust_input}
     \pz{ v }(\pzqAi,\Delta_0,[\Delta]) = \bigtimes_{j=1}^{n_q} \pz{ v }_j(\pzqAi,\Delta_0,[\Delta]).
\end{equation}

We define the \emph{input reachable set} for the $i$\ts{th} timestep as
\begin{equation}
\label{eq:pz_input}
\begin{split}
\pzujKi = \bm{\tau}(&\pzqAi,\Delta_0) + \\ &- \pz{ v }(\pzqAi,\Delta_0,[\Delta]).
\end{split}
\end{equation}
Because the robust input bound is treated as constant in each dimension in \eqref{eq:pz_robust_coeff}, when we slice the input reachable set by $k$, we only slice the polynomial zonotope representation of the nominal input by $k$, \emph{i.e.},
\begin{equation}
\begin{split}
\pzujki = \bm{\tau}(&\pzqAki,\Delta_0) + \\ &- \pz{ v }(\pzqAi,\Delta_0,[\Delta]).
\end{split}
\end{equation}
Using these definitions and the fact that all relevant operations involving polynomial zonotopes are either exact or overapproximate, one can prove that the input reachable set contains the robust passivity-based control input:
\begin{lem}[Input PZ Conservativeness]
\label{lem:pz_input_reach_set}
The input reachable set is over approximative, i.e., for each $k \in \pz{K}$
\begin{equation}
u(\qA(t;k), \nomparams, \intparams) \in \pzujki, \quad \forall t \in \pz{T_i}.
\end{equation}
\end{lem}
% \begin{proof}
% The polynomial zonotope bound of the nominal input $\bm{\tau}(\pzqAki,\Delta_0)$ is constructed using Alg. \ref{alg:PZRNEA}. 
% Because all operations in Alg. \ref{alg:PZRNEA} involve polynomial zonotopes with an input that is an overapproximation to $\qA(t;k)$, the output of the algorithm is an overapproximation to $\tau(\qA(t;k),\Delta_0)$. 
% The result then follows by applying Theorem \ref{thm:robust_input_bound} and \eqref{eq:pz_input}.
% \end{proof}

%~~~~~~~~ Constraint generation
\subsection{Constraint Generation} 
\label{subsec:constraint_generation}
We now have all the components to represent the trajectory optimization constraints using polynomial zonotopes.
Recall, that one can compute a conservative bound on the maximum or minimum of a polynomial zonotope by applying the $\setop{sup}$ and $\setop{inf}$ operators as in \eqref{eq:pz_sup} and \eqref{eq:pz_inf}, respectively
% More details on the implementation of these constraints is provided in Sec. \ref{sec:implementation}.
% \pat{I'm not sure we actually have to say a lot of what I wrote here... think we can just introduce the PZ optimization}

% \jon{\begin{itemize}
%     \item Recall that we are overapproximating a continuum of trajectories by a discretized set of polynomial zonotopes
%     \item In contrast to other methods that can only satisfy constraints at discretized time points, formulating our constraints in terms of polynomial zonotopes allows us to satisfy the constraints over time interval. In this section, we show how to derive constraints for $\dots$
% \end{itemize}}
\subsubsection{Joint Limit Constraints} 
\label{subsec:implementation_joint_constraints}
% \thought{
% \begin{itemize}
%     \item position constraints are simply box constraints
%     \item orientation constraints constrain the direction of the $x$ and $z$ axes of each frame.
%     \item note that using the forward kinematics we could constrain position and/or orientation of any point on the robot
% \end{itemize}
% }
% The position of the $j$\ts{th} frame is thus constrained to lie in a set \eqref{eq:pz_optposcon}.
% This formulation also allows us to constrain the orientation of the $j$\ts{th} frame by constraining its $\hat{x}_j$ \eqref{eq:pz_optorncon1} and $\hat{z}_j$ \eqref{eq:pz_optorncon2} axes.
The polynomial zonotopes $\pzqji$ and $\pzqdji$ incorporate tracking error and overapproximate all reachable joint angles and velocities over each time step.
Choosing $k$ such that $\pzqjki$ and $\pzqdjki$ are completely contained within $[\qlim^-, \qlim^+]$ and $[\dqlim^-, \dqlim^+]$, respectively, ensures that position and velocity limits are always satisfied.
 % One can compute a conservative bound on the maximum or minimum of a polynomial zonotope by applying the $\setop{sup}$ and $\setop{inf}$ operators as in \eqref{eq:pz_sup} and \eqref{eq:pz_inf}, respectively.

\subsubsection{Collision Avoidance Constraints}
\label{subsec:implementation_collision_constraints}
The forward occupancy of each of the robot's links is overapproximated by $\pzFOjKi$.
The robot must never collide with obstacles, which is guaranteed by choosing $k$ such that $\pzFOjki \bigcap \obsset = \emptyset$.
Because the obstacles are convex polytopes (Assum. \ref{assum:obstacles}), one can compute a halfspace representation of obstacle $O$ given by $\Aobs \in \R^{\nf \times 3}$ and $\bobs \in \R^{\nf}$, where $\nf$ is the number of faces of $O$.
A point $p_1 \in \R^{3}$ is contained within the obstacle if and only if $\Aobs p_1 - \bobs \leq 0$, where the inequality is taken element-wise.
Therefore, $p_1$ lies outside the obstacle if any of the inequalities do not hold.
More succinctly, $p_1 \not \in O \iff \max(\Aobs p_1 - \bobs) > 0$, where the max is taken element-wise.
This implies that $\pzFOjki \cap O = \emptyset \iff \max(\Aobs p_1 - \bobs) > 0, \; \forall p_1 \in \pzFOjki$.
The expression $\Aobs\pzFOjki - \bobs$ yields an $\nf \times 1$ polynomial zonotope.
The constraint that $\pzFOjki$ does not intersect $O$ can be written as $-\max{(\setop{inf}(\Aobs\pzFOjki - \bobs))} < 0$, where 
the $\max$ is taken element-wise.

\subsubsection{Input Constraints}
\label{subsec:implementation_input_constraints}
The input polynomial zonotope $\pzujKi$ is formed by applying \eqref{eq:pz_input}.
Importantly, the disturbances (caused by inertial parameter mismatch) and tracking error are not known at the time of planning.
The input polynomial zonotopes deal with these conservatively by assuming the worst-case disturbances and tracking error.
Choosing $k$ such that $\pzujki \subseteq [\ulim^-, \ulim^+]$ ensures that torque limits are not violated.

\subsection{Trajectory Optimization}
\label{sec:algorithm_theory:optimization}
We combine these polynomial zonotope constraints within a trajectory optimization program.
We seek to minimize a user-specified cost (such as reaching a desired intermediate waypoint) while satisfying each safety constraint:
\begin{align}
    \label{eq:pz_optcost}
    &\underset{k \in K}{\min} &&\numop{cost}(k) \\
    \label{eq:pz_optpos}
    &&& \hspace*{-0.75cm} \pzqjki \subseteq [\qlim^-, \qlim^+]  &\forall i \in N_t, j \in N_q \\
    \label{eq:pz_optvel}
    &&& \hspace*{-0.75cm}\pzqdjki \subseteq [\dqlim^-, \dqlim^+]  &\forall i \in N_t, j \in N_q \\
    \label{eq:pz_opttorque}
    &&& \hspace*{-0.75cm}\pzujki \subseteq [\ulim^-, \ulim^+]  &\forall i \in N_t, j \in N_q \\
    % \label{eq:pz_optposcon}
    % &&& \pzki{p_j} \subseteq [p_j^-, p_j^+] \qquad &\forall i, j \\
    % \label{eq:pz_optorncon1}
    % &&& \pzi{\hat{z}_{j}; k} \odot \pzi{\hat{z}_{d,j}}  \geq z_{d, j}^- \qquad &\forall i, j \\
    % \label{eq:pz_optorncon2}
    % &&& \pzk{\hat{x}_{j}} \odot \pz{\hat{x}_{d,j}}  \geq x_{d, j}^- \qquad &\forall i, j \\
    \label{eq:pz_optcolcon}
    &&& \hspace*{-0.75cm}\pzFOjki \bigcap \obsset = \emptyset  &\forall i \in N_t, j \in N_q. 
\end{align}
By applying Lem. \ref{lem:pz_q_bound}, \ref{lem:PZFK}, and \ref{lem:pz_input_reach_set} one can prove that any feasible solution to this optimization problem can be tracked safely by the robot:
\begin{lem}[Traj. Opt. Safety]
\label{lem:pz_opt_safe}
Suppose $k \in K$ satisfies the constraints in \eqref{eq:pz_optpos}--\eqref{eq:pz_optcolcon}, then $k$ can be tracked for all $t$ in $T$ by the robot while satisfying joint and input limits and without colliding into any obstacles.
\end{lem}

\subsection{\methodname{}'s Online Operation}

\begin{algorithm}[t]
\small
\begin{algorithmic}[1]
    % \State {\bf Require:} $q(0;k)$, $\dot{q}(0;k)$, $t\plan$, cost function $J: K \to \R$, $\{ \skai \}_{i=1}^{n_q}$, previous plan $q_{\regtext{prev}}: \T \to Q$,  $A \leftarrow \emptyset,\ b \leftarrow \emptyset$
    
    % \State $\{\Vit\} \leftarrow \numop{composeRS}(\initq,\initdq)$ // Sec. \ref{sec:online_RS_implementation} \label{lin:construct_frs}
    
    % \State $(h\obs,h\jlim) \leftarrow \numop{makeCons}(\initq,\initdq,\obsset,\{\Vit\})$ // Sec. \ref{sec:online_constraint_generation}
    
    % \State // solve \eqref{prog:trajopt} within $t\plan$ or else return $q\prev$
    
    % \State $q\plan \leftarrow \numop{optTraj}\left(\costfunc, h\obs, h\jlim, t\plan, q\prev\right)$ // Sec. \ref{sec:online_trajopt} \label{lin:trajopt}
    % %\pat{now $f$ not $\phi$, h lim formatting}

    % \State {\bf Require:} $\Nt \in \N, [\Delta],$ and $\Delta_0 \in [\Delta]$

    \State{\bf Parfor} $i = 1:\Nt$ %// parallel for each time step

    % \State\hspace{0.2 in}{\bf parfor} $j = 1:\nq$ %// for each joint

    \State\hspace{0.1in} $\{ \pzqi, \ldots, \pzqdddesi  \} \leftarrow \setop{PZ}(\initq,\initdq,\initddq)$ // Sec. \ref{subsec:pz_traj_rep}

    \State\hspace{0.1in} $\pzqAi \gets \{ \pzqi, \ldots, \pzqdddesi\}$ // Sec. \ref{subsubsec:theory_input_reach_set}

    \State\hspace{0.1in} // create forward occupancy reachable set //

    \State\hspace{0.1in} $\pz{FK}(\pzqi) \gets \setop{PZFK}(\pzqi)$ // Alg. \ref{alg:compose_fk}
    
    \State\hspace{0.1in} $\pz{FO}(\pzqi) \gets$ \eqref{eq:pz_forward_occupancy}
    
    \State\hspace{0.1in} // create input reachable set //

    % \State\hspace{0.2in} $\pzi{\qA} \gets$ \eqref{eq:pz_total_feedback}

    \State\hspace{0.1in} $\pzgreek{\tau}(\pzqAi,\Delta_0) \gets \setop{PZRNEA}(\pzqAi, \nomparams)$ // Alg. \ref{alg:PZRNEA}

    \State\hspace{0.1in} $\pz{ v }(\pzqAi,\Delta_0,[\Delta]) \gets$ \eqref{eq:pz_robust_input}

    \State\hspace{0.1in} $\pzujKi \gets$ \eqref{eq:pz_input}

    \State\hspace{0.1in} // create constraints //

    \State\hspace{0.1in} $\numop{jointCons}(\pzqi, \pzqdi)$ // Sec. \ref{subsec:implementation_joint_constraints}

    \State\hspace{0.1in} $\numop{collisionCons}(\pz{FO}(\pzqi), \obsset)$ // Sec. \ref{subsec:implementation_collision_constraints}

    \State\hspace{0.1in} $\numop{inputCons}(\pzi{u})$ // Sec. \ref{subsec:implementation_input_constraints}

    \State\hspace{0in}{\bf End Parfor}

    \State{\bf Try:} $k^* \gets$ solve \eqref{eq:pz_optcost} -- \eqref{eq:pz_optcolcon}
    \State{\bf Catch:} (if $t\elapsed > t\plan$), {\bf then}  $k^* = \texttt{NaN}$ // $t\elapsed$ measures the amount \phantom{dumb} of time since $\numop{Opt}$ was called //
    % \State $k^* \gets$ solve \eqref{eq:pz_imp_opt} within $t\plan$, else return $\texttt{NaN}$

    % \State {\bf If} $k^* \not = \texttt{NaN}$ {\bf return} $\{q_d(\cdot; k^*), \dot{q}_d(\cdot; k^*), \ddot{q}_d(\cdot; k^*) \}$
    % \State {\bf Else} {\bf return} previous plan
    
    \end{algorithmic}
\caption{\small \phantom{stupid stupid stupid stupid stupid stupid stup} ${\{ k^* \} = \numop{Opt}(\initq,\initdq,\initddq,\obsset,\numop{cost},t\plan, \Nt, \Delta_0, [\Delta])}$}
\label{alg:opt}
\end{algorithm}
\methodname's planning algorithm is summarized in Alg. \ref{alg:opt}.
Note in particular, that the construction and solving of the optimization problem described in lines \eqref{eq:pz_optpos}--\eqref{eq:pz_optcolcon} is given $t\plan$ time. 
If a solution is not found within that time, then the output of Alg. \ref{alg:opt} is set to $\texttt{NaN}$.
To facilitate real-time motion planning, we include the analytical constraint gradients when numerically solving the optimization problem.
As described above, each constraint contains a polynomial zonotope whose coefficients correspond to the trajectory parameters and are also the decision variables of the optimization problem.
Because each polynomial zonotope can be viewed as a polynomial function solely of the trajectory parameters, we can compute each constraint gradient with respect to the trajectory parameter $k$ using the standard rules of differential calculus, \emph{i.e.}, the power, product, and chain rules.

\begin{algorithm}[t]
\small
\begin{algorithmic}[1]
    \State {\bf Require:} $t\plan > 0$, $\Nt \in \N, [\Delta],\Delta_0 \in [\Delta],\obsset,$ and $\texttt{cost}: K \to \R,$.
    \State {\bf Initialize:} $j = 0$, $t_j = 0$, and \newline
    \phantom{\bf Initialize:} $\{ k^*_j \} = \texttt{Opt}(\qstart,\zeros,\zeros,\obsset,\texttt{cost}, t\plan, \Nt, \Delta_0, [\Delta])$
    \State {\bf If} $k^*_{j} = \texttt{NaN}$, {\bf then} break
    \State{\bf Loop:}

        \State \hspace{0.05in} // Line \ref{lin:apply} executes simultaneously with Lines \ref{lin:opt} -- \ref{lin:else} //

        \State \hspace{0.05in}  // Use Thm. \ref{thm:robust_controller}, Lem. \ref{lem:compute_input}, and Alg. \ref{alg:IRNEA}// \newline  \hspace{0.05in}  {\bf Apply} $u(\qA^j(t;k^*_j), \nomparams, \intparams)$  to robot for $t \in [t_j,t_j + t\plan]$ \label{lin:apply} 

        \State \hspace{0.05in}  $\{ k^*_{j+1} \} = \texttt{Opt}(q_d(t\plan;k^*_j), \dot{q}^{j}_d(t\plan;k^*_j), \ddot{q}^{j}_d(t\plan;k^*_j),\obsset,$ \phantom{stupid stupid stupi}$\texttt{cost},t\plan, \Nt, \Delta_0, [\Delta])$ // Alg. \ref{alg:opt} \label{lin:opt}

        \State \hspace{0.05in} {\bf If} $k^*_{j+1} = \texttt{NaN}$, {\bf then} break
        \State \hspace{0.05in} {\bf Else}  $t_{j+1} \leftarrow t_j + t\plan$ and $j \leftarrow j + 1$ \label{lin:else}

    \State{\bf End}
    \State {\bf Apply} $u(\qA^j(t;k^*_j), \nomparams, \intparams)$  to robot for $t \in [t_j+t\plan,t_j + \tfin]$
\end{algorithmic}
\caption{\small \methodname{} Online Planning and Control}
\label{alg:armourPlan}
\end{algorithm}

Algorithm \ref{alg:armourPlan} summarizes the online operation of \methodname{}. 
\methodname{} uses the robust passivity-based controller \eqref{eq:controller} to track the trajectory parameter computed at the previous planning iteration on Line \ref{lin:apply}. 
While this input is being applied, \methodname{} computes the trajectory to be tracked at the next planning iteration on Line \ref{lin:opt}.
By applying Cor. \ref{cor:tracking_error} and Lem. \ref{lem:pz_opt_safe}, one can prove that \methodname{} generates dynamically feasible, collision free behavior:
\begin{lem}[\methodname{} is Safe]
If $\qstart$ is collision free and one applies \methodname{}, as described in Algorithm \ref{alg:armourPlan}, then the robot is collision free. 
\end{lem}

% We conclude by making a remark about applying \methodname{} to real-world systems. 
In practice, the robust passivity-based controller on Line \ref{lin:apply} is computed at discrete time instances at the input sampling rate requested by the robot.
As we describe in Sec. \ref{sec:experiments}, the input sampling rate of manipulator robots are typically a kilohertz or more \cite{kinova-user-guide,panda-user-guide}. 
Fortunately, the robust passivity-based controller can be computed at that rate. 
However, the input applied into the robot is usually zero-order held between sampling instances. 
To apply Cor. \ref{cor:tracking_error} in this instance, one would need to show that the tracking error bound was still satisfied if the input was applied in this zero-order held fashion.
Note, one could extend the results in this paper to address this additional requirement by extending the controller presented in this paper to deal with arbitrary bounded input disturbances \cite[(1)]{giusti2016bound}. 
However, this extension falls out of the scope of this paper. 

\section{Demonstrations}
\label{sec:experiments}
We demonstrate \methodname in simulation and on hardware using the Kinova Gen3 7 DOF robotic arm.
Our code can be found online\footnote{\href{https://github.com/roahmlab/armour}{https://github.com/roahmlab/armour}}.
% \bohao{click "here"! Not sure if I did it properly}
% \pat{insert figure!}
% \pat{add link to github repo here?}
% \shrey{Add a one-sentence ``why is our method awesome'' takeaway here.}

\subsection{Implementation Details}
\label{sec:experiments:implementation_details}
\methodname is implemented in MATLAB, C++, and CUDA on a desktop with a 16 core 32 thread processor, 128 GB RAM, and Nvidia Quadro RTX 8000 GPU.
Polynomial zonotopes and their arithmetic in C++/CUDA based on the CORA toolbox \cite{althoff_cora}.

\subsubsection{Kinova Robot}
The Kinova Gen3 is composed of 7 revolute DOFs \cite{kinova-user-guide}.
During the simulation evaluation, we use the Kinova arm without its gripper.
We sampled a million configurations and found that the minimum and maximum eigenvalues of its mass matrix were uniformly bounded from above and below by$\sigma_M = 15.79636$ and $\sigma_m = 5.09562$, respectively.
During the real-world evaluation, we use the Kinova arm with its gripper and a $2$ lb dumbbell rigidly attached to the gripper.
The minimum and maximum eigenvalues of its mass matrix were uniformly bounded from above and below by $\sigma_M = 18.2726$ and $\sigma_m = 8.29939$, respectively.

We consider two setups for adding uncertainty to the arm.
The first case is used in simulation evaluation, which allows the mass of each link of the robot to vary by $\pm3\%$ of its nominal value.
The inertia matrices of each link vary accordingly.
We refer to this setup as the ``standard'' set of inertial parameters, denoted by $\intparams$.
The second case is used in real-world evaluation, which rigidly attaches a $2$ lb dumbbell to the arm's end effector.
The geometry of the dumbbell is considered when checking for collisions.
We treat the dumbbell's inertia as a point mass located at its centroid, and allow its mass to vary by $\pm3\%$, without varying the rest of the arm's parameters. 
% \pat{how are we actually dealing with dumbbell inertia in hardware?} \jon{I'm waiting to here back from Baiyue to see if he changed the inertia tensor.}
We call this the ``dumbbell'' set of inertial parameters, denoted by $\intparamsdb$. 

\subsubsection{Planning Times and Trajectories}
We let $t\plan = 0.5$s, and $\tfin = 1$s.
As in Ex. \ref{ex:bernstein}, we represent our desired trajectories using a degree 5 Bernstein polynomial and let $\kjoffset = \initqj$ and $\kjscale = \frac{\pi}{48}$.
% As a result, after $\tfin = 1$s the final position of any joint's trajectory can differ from its initial position by up to $\pm\frac{\pi}{48}$ radians for the ``standard'' inertial parameters, and $\pm\frac{\pi}{50}$ radians for the ``dumbbell'' inertial parameters.
% \pat{these can change.} \bohao{$\pm\frac{\pi}{24}$ radians for kinova without gripper, $\pm\frac{\pi}{50}$ radians for kinova with gripper and dumbbell on the real robot}

\subsubsection{Controller}
For the nominal passivity-based controller we set $K_r = 5 I_7$, where $I_7$ is a $7 \times 7$ identity matrix.
During our experiments, we were able to compute the passivity-based control input at approximately $15 kHz$.
However, the software for the arm only required a control update at $1 kHz$.
% \pat{I think we might actually want to try $K_r = 1 I_7$, which would give us a little worse position error but I think a much smaller max disturbance when using \texttt{PZRNEA}.} \bohao{Kr is 10 right now. looks good}

For the robust controller, several additional parameters must be specified.
As in Thm. \ref{thm:robust_input_bound}, we set $\alpha(y) = y$.
% , which satisfies the requirement that $\alpha: \R \to \R$ is an extended class $\mathcal{K}_\infty$ function (\emph{i.e.}, $\robKinf: \R \to \R$ is strictly increasing with $\robKinf(0) = 0$ and is defined on the entire real line $\R = (-\infty, \infty)$.
We let the Lyapunov function threshold be $\roblevel = 1\times 10^{-2}$.
% \bohao{Do I have permission to change this? cc reported that this is actually 1e-4}
% $\roblevel = 3.1 \times 10^{-7}$
% Together with the bound on the smallest eigenvalue $\sigma_m$, per \eqref{eq:ultimate_bound_defn}, this yields the uniform bound of 
% % \jon{Are we using the same uniform bound with and without the gripper?}
% % \bohao{Yes it should be the same. But the eigenvalues are different.}
% \begin{equation}
%     \epsilon = \ultbound = \sqrt{\frac{2 \times 1 \times 10 ^ {-2}}{8.29938}} \approx 0.0490899
% \end{equation}
% when we run the experiments with the gripper as we do in the real-world.
When we run the experiment in simulation, we do it without the gripper and the uniform bound is $\epsilon = \ultbound = \sqrt{\frac{2 \times 1 \times 10 ^ {-2}}{5.09562}} \approx 0.062649$.
Plugging into the position \eqref{eq:ultimate_bound_position} and velocity \eqref{eq:ultimate_bound_velocity} bounds yields $\pboundj \approx 0.01253 \text{ rad}$ and $\vbound \approx 0.02506 \text{ rad/s}$. 

% \subsubsection{Polynomial Zonotopes}
% We implement polynomial zonotopes and their arithmetic in C++/CUDA based on the CORA toolbox \cite{althoff_cora}.
% In this work, we reduce a generator if its norm is smaller than 5e-4.
% In other words, any generator whose norm is smaller than 5e-4 will be over-approximated as an interval.
% This allows the number of generators in one polynomial zonotope to be decreased for faster evaluation, while also creating less over-approximation.
%And we use a maximum zonotope order of 1 for independent generators, which means for 3-%dimensional polynomial zonotopes a maximum of 3 independent generators are kept after each %operation.
% The number of generators is managed by frequently reducing the polynomial zonotopes, as discussed in \cite[Prop. 16]{kochdumper2020sparse}. 
% \bohao{not sure how to describe this for C++ implementation... It's much more complicated...} \bohao{Follow up: The current version should be exactly the same.}

\subsubsection{High-level Planners}
In each planning iteration, \methodname minimizes a user-specified cost subject to the constraints detailed in Sec. \ref{sec:algorithm_theory}.
This work formulates the cost as minimizing the distance between $q_d(\tfin; k)$ and an intermediate waypoint $q_\regtext{des}$.
We use high-level planners to construct these intermediate waypoints, which form a rough path to the global goal.
\methodname is independent of the high-level planner, which is only used for the cost function.
\methodname enforces safety even if the high-level planner's waypoints are in collision or not dynamically feasible.
In the majority of simulations presented here, we use a straight line high-level planner that generates waypoints along a straight line (in configuration space) between the start and goal configurations.
For some scenarios (detailed below), we also test an RRT\ts{*} \cite{karaman2011sampling} that only ensures the robot's end effector is collision free.

\subsubsection{Comparisons}
We compare \methodname to a previous version of the method ARMTD \cite{holmes2020armtd} by running the algorithms on identical random worlds with the same high-level planner.
We also compare \methodname to CHOMP \cite{zucker2013chomp} (run via MoveIt \cite{moveit}) as was done previously with ARMTD.
Although CHOMP is not a receding-horizon planner, it provides a useful baseline for measuring the difficulty of the scenarios encountered by \methodname.
{\bf ARMTD and CHOMP are given the true inertial parameters, but \methodname is not given the true inertial parameters}.
This means that ARMTD and CHOMP can perfectly track trajectories that they compute while \methodname may not be able to, which should give ARMTD and CHOMP a considerable advantage when compared to \methodname.

\subsubsection{Algorithm Implementation}
\label{sec:experiments:alg_implementation}
% \pat{Provide details about solve times for GPU version, and which elements of the algorithm take the longest.} \bohao{Input constraints take the longest.}
We solve \methodname's trajectory optimization using Ipopt \cite{ipopt-cite}.
%using MATLAB's $\texttt{fmincon}$. \pat{cite?}
% The cost function is $\numop{cost}(k) = \norm{q_d(\tfin; k) - q_{\regtext{des}}}$, where $q_{\regtext{des}}$ is the intermediate waypoint provided by the high-level planner.
We provide analytic gradients to the optimization solver to speed computation.
% \bohao{Check this out}

\begin{figure}[t]
    \centering
    \includegraphics[width=0.9\columnwidth]{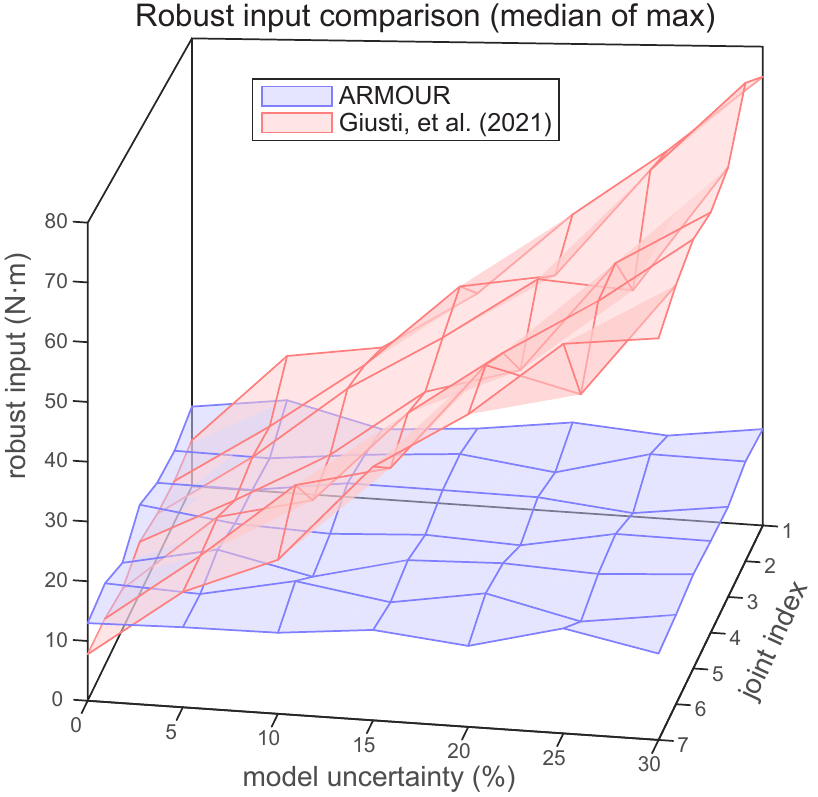}
    \caption{Robust input comparison. This figure compares \methodname{}'s robust input \eqref{eq:robust_input} to that of \cite[Theorem 2]{giusti2021interval} across each of the robot's joints. The uncertainty in the mass and the inertia of each link was varied between 0-30\%. Compared to \cite{giusti2016bound}, \methodname{}'s robust input is far less sensitive to increasing model uncertainty.}
    \label{fig:controller_comparison}
    \vspace*{-0.2cm}
\end{figure}

\subsection{Robust Controller Comparison}
\label{section:controller_comparison}
% objective
We compare the performance of our controller's robust input \eqref{eq:robust_input}  to that of \cite{giusti2016bound} under different model uncertainties\footnote{Note for a comparison between the input bounds derived from our robust input and that of \cite{giusti2016bound}, please see  \href{https://github.com/roahmlab/armour/blob/main/assets/TRO_Armour_Appendix_F.pdf}{https://github.com/roahmlab/armour/blob/main/assets/TRO-Armour-Appendix-F.pdf}}.
% motivation
In Def. \ref{def:traj_param} and Thm. \ref{thm:robust_controller}, we construct the desired trajectories such that the tracking error is always bounded.
Within our planning framework, this ensures that the robot stays within the reachable set despite model uncertainty.
However, our controller can be applied more generally without assumptions on the initial state of the robot.
% We demonstrate that our robust controller outperforms that of \cite{giusti2016bound}.

% experiment overview
In this experiment, we perturb the initial state of the desired trajectory such that the tracking error is not always guaranteed to be inside the ultimate bound.
This allows us to compare the magnitude of the robust input required to correct the tracking error to within the ultimate bound.
% experiment details
We compared the robust inputs across 7 different model uncertainties, ranging from 0\% to 30\% in the mass and inertia of each link.
The parameters of the two controllers are chosen such that their ultimate bounds are the same.
We randomized the perturbation added to the initial state of all $7$ joints.
The magnitude of the perturbation added to the initial position and initial velocity are 4.5 degrees and 9 degrees/second, respectively.
For each model uncertainty, we perform 100 experiments where each controller tracks a desired trajectory for 2.5 seconds after the initial state of the robot is randomly perturbed.
Note that the ultimate bound was usually reached within $1.5$ seconds for either controller.
We then take the maximum robust input of each experiment and plot the median over all experiments.
As shown in Fig. \ref{fig:controller_comparison}, our controller applies a smaller robust input to achieve the same ultimate tracking performance.

\subsection{Simulation}
\subsubsection{Setup}
\begin{figure*}
    \centering
    \includegraphics[width=\textwidth]{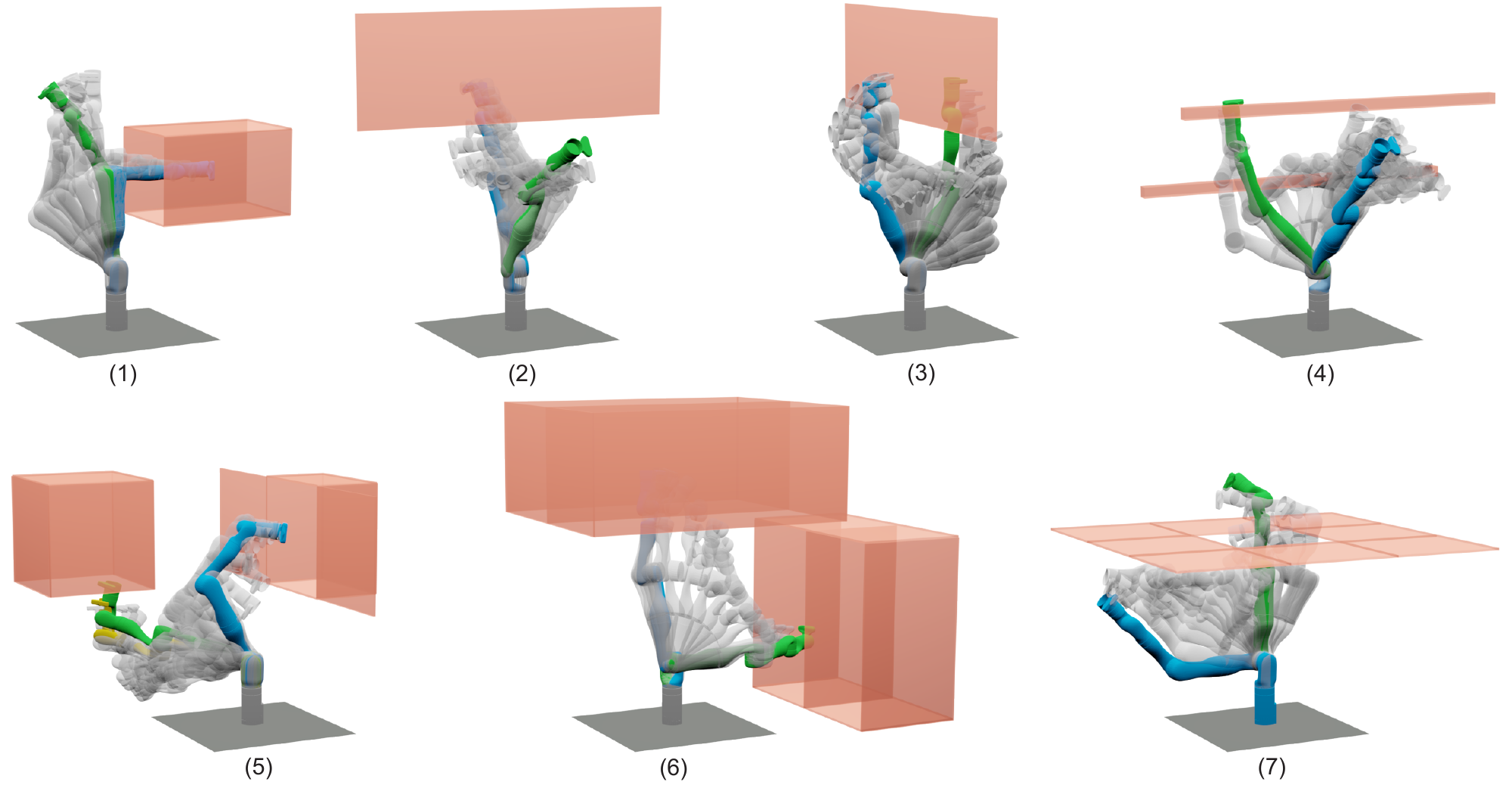}
    \caption{ 
    The set of seven Hard Scenarios (number in the top left), with start pose shown in blue and goal pose shown in green.
    There are seven tasks in the Hard Scenarios set: (1) from inside to outside of a box, (2-3) from one side of a wall to another, (4) between two horizontal posts, (5) from a sink to a cupboard, (6) from one set of shelves to another, (7) through a small window. Note that \methodname fails to reach the goal in scenario (5) and the final pose is shown in yellow.}
    \label{fig:hard_scenarios}
\end{figure*}

% \pat{COPIED AND PASTED FROM RSS PAPER}
As in \cite{holmes2020armtd}, we test \methodname on two sets of scenes.
The first set, Random Obstacles, shows that \methodname can handle arbitrary tasks.
% (see Fig \pat{add fig}).
This set contains 100 tasks with random (but collision-free) start and goal configurations, and random box-shaped obstacles.
Obstacle side lengths vary from $1$ to $50$ cm, with 10 scenes for each $n_O = 13, 16, ..., 37, 40$.
The second set, Hard Scenarios, shows that \methodname guarantees safety where CHOMP converges to an unsafe trajectory.
There are seven tasks in the Hard Scenarios set shown in Fig. \ref{fig:hard_scenarios}.
For both these sets of scenes, \methodname uses the ``standard'' inertial parameters $\intparams$ with $3\%$ uncertainty in each link's mass.
% Note that \methodname is compared to ARMTD and CHOMP {\bf that are both assumed to have access to the robot's nominal parameters $\nomparams$ with no uncertainty}.
% This should give both of these methods a considerable advantage when compared to \methodname.

% \jon{Are we actually doing this?}
% \pat{Commented out ``ablation study'' results for now.}
% Next, we perform an ablation study using the ``dumbbell'' parameters $\intparamsdb$ to show the effectiveness of various components of \methodname.
% Here, we generate a new set of Random Obstacles, and test three different versions of \methodname.
% First, with both collision-avoidance and input constraints active.
% Second, with only collision-avoidance and no input constraints.
% Third, with only collision-avoidance constraints and using the nominal passivity-based controller alone, so that tracking error is not considered by the planner.

In all tests using the Random Obstacles scenario, the straight line high-level planner is used.
For the Hard Scenarios set, we also present results using the RRT\ts{*}.

\subsubsection{Results}
Table \ref{tab:random_obstacles_results} (Random Obstacles) presents \methodname (with a straight line high-level planner) in comparison to ARMTD and CHOMP as in \cite{holmes2020armtd}.
\methodname reached $92/100$ goals and had $0/100$ crashes, meaning \methodname safely stopped in $8/100$ scenes without finding a new trajectory. 
\methodname takes $364.9$ms on average per planning iteration.
Note that \methodname actually outperforms ARMTD using the same high-level planner despite dealing with uncertainty in the inertial parameters $\intparams$.
% This demonstrates that \methodname's collision-avoidance constraints have been made less conservative than ARMTD's.
This is likely due to the updated polynomial zonotope formulation of the robot's forward occupancy.
These sets now more tightly envelope the actual reachable set of the robot, allowing for closer maneuvering around obstacles.
We note that CHOMP always finds a dynamically feasible trajectory, but not necessarily a collision-free one, as demonstrated by the $13/100$ crashes.
In practice, an external collision-checker can be used to verify that these trajectories are collision free before commanding them on the robot, but we note that CHOMP itself does not provide this guarantee.

% \bohao{where shall I report planning time for this scenario?}

Table \ref{tab:hard_scenarios_results} presents results for the Hard Scenarios.
With the straight line high-level planner, \methodname is unable to complete any tasks but also has no collisions.
With the RRT* high-level planner, \methodname successfully completes $6/7$ scenarios and safely stops in the remaining scenario.
This improves on the results for ARMTD with the same high-level planner, which completed $5/7$ scenarios.

% Table \ref{tab:random_obstacles_dumbbell_results} shows results when manipulating a dumbbell of uncertain mass in random environments.
% When using the forward occupancy collision-avoidance constraints and input constraints, along with the full controller \eqref{eq:controller}, \methodname completes $73/100$ tasks and never crashes or saturates inputs.
% Without the input constraints, the arm completes $75/100$ tasks, never crashes, and also does not saturate inputs (though there is no guarantee against this).
% Finally, using the nominal passivity-based controller alone and collision-avoidance constraints that do not account for tracking error, the arm only completes $8/100$ tasks and crashes $78/100$ times.
% These results emphasize the importance of the robust controller's tracking performance and associated tracking error bounds. \jon{My only concern here is that the input constraints are not important}

% \begin{table}[t]
% \centering
% \begin{tabular}{r|c|c|c|c|}
% \multicolumn{1}{c|}{\textbf{Random Obstacles}} & \% goals & \% crashes & MST {[}s{]} & MNPD \\ \hline
% ARMTD + SL & 84 & 0 & 0.273 & 1.076\\ \hline
% % ARMTD + RRT & 62 & 0 & 0.466 & 1.417 \\ \hline
% CHOMP & 82 & 18 & 0.177 & 1.511\\ \hline
% \end{tabular}
% \caption{MST is mean solve time (per planning iteration for ARMTD with a straight-line planner, total for CHOMP) and MNPD is mean normalized path distance. MNPD is only computed for trials where the task was successfully completed, i.e. the path was valid.}
% \label{tab:random_obstacles_results}
% \end{table}

\begin{table}[t]
\caption{Results for the 100 Random Obstacles simulations.
ARMTD and \methodname use straight-line (SL) HLPs. }
\centering
\begin{tabular}{r|c|c|}
\multicolumn{1}{c|}{\textbf{Random Obstacles}} & \% goals & \% crashes \\ \hline
ARMTD + SL + $\nomparams$ & 72 & 0\\ \hline
% ARMTD + RRT & 62 & 0 & 0.466 & 1.417 \\ \hline
CHOMP + $\nomparams$ & 87 & 13\\ \hline
(ours) \methodname + SL + $\intparams$ & 92 & 0\\ \hline
\end{tabular}
\label{tab:random_obstacles_results}
\end{table}

\begin{table}[t]
\caption{
Results for the seven Hard Scenario simulations.
ARMTD and \methodname use straight-line (SL) and RRT* HLPs.
The entries are ``O'' for task completed, ``C'' for a crash, or ``S'' for stopping safely without reaching the goal.
}
\centering
\begin{tabular}{l|c|c|c|c|c|c|c|}
\multicolumn{1}{c|}{\textbf{Hard Scenarios}} & 1 & 2 & 3 & 4 & 5 & 6 & 7 \\ \hline
CHOMP + $\nomparams$ & C & C & C & C & C & C & C \\ \hline
ARMTD + RRT* + $\nomparams$ & O & O & O & S & S & O & O\\ \hline
(ours) \methodname + SL + $\intparams$ & S & S & S & S & S & S & S\\ \hline
(ours) \methodname + RRT* + $\intparams$ & O & O & O & O & S & O & O\\ \hline
\end{tabular}
\label{tab:hard_scenarios_results}
\end{table}

\subsection{Hardware}
Finally, we applied \methodname to perform real-time control and planning in the real-world. using the Kinova Gen3 robot while it holds a dumbbell.
A video showing a hardware demonstration of \methodname safely controlling the Kinova Gen3 robot is available online\footnote{\href{https://youtu.be/-WtxxQyoxGo?si=W2RaziwRLMosZ23L}{https://youtu.be/-WtxxQyoxGo?si=W2RaziwRLMosZ23L}}.
This experiment illustrates that \methodname can operate safely in real-time despite modeling uncertainty. 
\section{Conclusion}
\label{sec:conclusion}
We present \methodname, a real-time planning and control framework with strict safety guarantees for manipulators with set-based uncertainty in their inertial parameters.
A robust control formulation (Thm. \ref{thm:robust_controller}) provides uniform bounds on the worst-case tracking error possible when following desired trajectories.
The PZRNEA algorithm (Alg. \ref{alg:PZRNEA}) enables \methodname to compute sets of possible inputs required to track any desired trajectory, including tracking error.
Polynomial zonotope arithmetic also enables the formulation of continuous-time collision-avoidance constraints.
Strict joint limit, torque limit, and collision-avoidance constraints are implemented within a nonlinear program that is solved at every iteration of \methodname's receding-horizon planning scheme.
By designing the inclusion of a fail-safe braking maneuver in every desired trajectory, \methodname guarantees the safe operation of the robotic arm for all time.

\renewcommand{\bibfont}{\normalfont\footnotesize}
{\renewcommand{\markboth}[2]{}% Remove header adjustment
\printbibliography}

% \appendix
\appendices
\section{Proof of Thm. \ref{thm:robust_controller}}
\label{app:robust_controller_proof}
\begin{proof}
Let $\robh = -\roblyap + \roblevel$.
    % \begin{equation}
    %     \label{eq:CBF_defn}
    %     \robh = -\roblyap + \roblevel.
    % \end{equation}  
This proof shows that $h$ is a minimal Control Barrier Function (CBF) under the input \eqref{eq:robust_input} \cite[Def. 3]{konda2020characterizing}. 
With this property and if \eqref{eq:robust_input} is continuous with respect to its first argument, then all zero super-level sets of $h$, defined as $\robH = \{ \qA(t) \; | \; \robh \geq 0 \}$,
% \begin{equation}
%     \label{eq:lyap_level_set}
%     \robH = \{ \qA(t) \; | \; \robh \geq 0 \},
% \end{equation}
are forward invariant \cite[Thm. 4]{konda2020characterizing}.
If $\robH$ is forward invariant, this proves the bound on $r(t)$.
To see this, note the trajectory starts in $H$, and as a result will remain in $H$. 
Next, note that $\roblyap \leq \roblevel, \; \forall \qA(t) \in \robH$.
By applying Ass. \ref{ass:eigenvalue_bound}:
\begin{equation}
    \frac{1}{2} \sigma_m \norm{\robr(t)}^2 \leq \frac{1}{2} \robr(t)^\top \bM(q(t), \trueparams) \robr(t) = \roblyap.
\end{equation}
Therefore, $\norm{\robr(t)} \leq \epsilon, \quad \forall \qA(t) \in \robH$.
So, we show that $h$ is a minimal control barrier function and \eqref{eq:robust_input} is continuous.

\noindent {\bf $h$ is a minimal CBF:} Recall that the manipulator dynamics are control affine \cite[(6.61)]{spong2005textbook}. 
To prove that $h$ is a minimal control barrier function, we need to prove that under the robust control input:
\begin{equation}
    \label{eq:CBF_cond}
    \dot{h}(\qA(t),\Delta) \geq -\robKinf(\robh).
\end{equation}
To apply \cite[Def. 3]{konda2020characterizing}, $\robKinf$ must be a minimal function.
Any extended class $\mathcal{K}_\infty$ function is a minimal one \cite[Case 2,Thm. 2]{konda2020characterizing}. 
As we show below, $\dot{h}$ is a function of $\Delta, \nomparams,$ and $\intparams$.
These dependencies are clear in context, so we suppress them. 

We make several observations:
$V$ is positive definite because $\bM(q(t),\trueparams))$ is positive definite.
Next,
% a suitable factorization of $\bC(q(t), \dot{q}(t), \trueparams)$ renders
 $N(q(t), \dot{q}(t), \trueparams) = \Mqdot - 2\bC(q(t), \dot{q}(t), \trueparams)$ is skew-symmetric \cite[Ch. 7]{siciliano2009modelling}:
\begin{equation}
    \label{eq:skew_symmetric_N}
    x^\top N(q(t), \dot{q}(t), \trueparams) x = 0, \qquad \forall x \in \R^{n_q}.
\end{equation}
Taking the time derivative of $\robh$, substituting \eqref{eq:modified_manipulator_equation}, and taking advantage of \eqref{eq:skew_symmetric_N}, yields
% \begin{equation}
% \begin{split}
%     \roblyapdot = \robr(t)^\top & \bM(q(t),\trueparams)) \robrdot(t) + \\ + &\frac{1}{2}\robr(t)^\top \Mqdot \robr(t),
%     \end{split}
% \end{equation}
% which after substituting \eqref{eq:modified_manipulator_equation} and taking advantage of \eqref{eq:skew_symmetric_N} yields
% \begin{equation}
%     \label{eq:dot_lyap_r}
%     \begin{split}
%     \roblyapdot =\robr(t)^\top &\robv(\qA(t),\nomparams,\intparams) + \\ &+\robr(t)^\top \robw(\qA(t),\nomparams,\Delta).
%     \end{split}
% \end{equation}
\begin{align}
    \robhdot
    &= -\robr(t)^\top \robv(\qA(t),\nomparams,\intparams) + \\ & \phantom{= -\robr(t)^\top} - \robr(t)^\top \robw(\qA(t),\nomparams,\Delta), \nonumber
\end{align}
$\dot{h}$ is a function of $\Delta, \nomparams,$ and $\intparams$, but we have suppressed these dependencies.
% However, because these dependencies are clear in context we suppress them for convenience. 
% As a result, 
% \begin{align}
%     \robhdot
%     &= -\robr(t)^\top \robv(\qA(t),\nomparams,\intparams) + \\ & \phantom{= -\robr(t)^\top} - \robr(t)^\top \robw(\qA(t),\nomparams,\Delta), \nonumber
% \end{align}
Using this result, \eqref{eq:CBF_cond} becomes
\begin{equation}
    \label{eq:CBF_cond_v}
    \begin{split}
    -\robr(t)^\top &\robv(\qA(t),\nomparams,\intparams) + \\ &-\robr(t)^\top \robw(\qA(t),\nomparams,\Delta) \geq -\robKinf(\robh).
    \end{split}
\end{equation}
Our task is to choose $\robv$ so that \eqref{eq:CBF_cond_v} is always satisfied.

% Rearranging, we obtain
% \begin{equation}
%     \label{eq:CBF_cond_v_rearranged}
%     -\robr(t)^\top \robv(t) \geq -\robKinf(\robh) + \robr(t)^\top \robw(t).
% \end{equation}
% Because $\norm{\robr(t)}\norm{\wmax(\qA(t), \nomparams)} \geq \robr(t)^\top \robw(\qA(t),\nomparams,\Delta)$, we can rearrange to obtain the stricter condition:
By Lem. \ref{lem:disturbance_bound} and \eqref{eq:cond_wmax}, $\abs{\robr(t)}^\top\wmax(\qA(t), \nomparams, \intparams) \geq \robr(t)^\top \robw(\qA(t),\nomparams,\Delta)$, so we rearrange \eqref{eq:CBF_cond_v} to obtain:
% \begin{equation}
%     \label{eq:CBF_cond_v_stricter}
%     \begin{split}
%     -\robr(t)^\top \robv(\qA(t),\nomparams,\intparams) \geq &-\robKinf(\robh) + \\& + \norm{\robr(t)}\norm{\wmax(\qA(t), \nomparams)},
%     \end{split}
%     % \implies -\robr(t)^\top\robv(t) &\geq -\robKinf(\robh) + \robr(t)^\top \robw(t).
% \end{equation}
\begin{equation}
    \label{eq:CBF_cond_v_stricter}
    \begin{split}
    -\robr(t)^\top \robv(\qA(t),\nomparams,\intparams) \geq &-\robKinf(\robh) + \\& + \abs{\robr(t)}^\top\wmax(\qA(t), \nomparams, \intparams),
    \end{split}
    % \implies -\robr(t)^\top\robv(t) &\geq -\robKinf(\robh) + \robr(t)^\top \robw(t).
\end{equation}
whose satisfaction guarantees that \eqref{eq:CBF_cond_v} is satisfied.
The left hand side expression is maximized when $\robv(\qA(t),\nomparams,\intparams)$ points in the $-\frac{\robr(t)}{\norm{\robr(t)}}$ direction.
Choosing $\robv(\qA(t), \nomparams, \intparams ) = -\robcoeff(\qA(t),\nomparams, \intparams) \frac{\robr(t)}{\norm{\robr(t)}}$, where $\robcoeff(\qA(t),\nomparams, \intparams) \geq 0$, yields
% \begin{equation}
% \begin{split}
%     \robcoeff(\qA(t),\nomparams, \intparams)  \norm{\robr(t)} \geq &-\robKinf(\robh) + \\ & + \norm{\robr(t)}\norm{\wmax(\qA(t), \nomparams)}.
%     \end{split}
% \end{equation}
% \begin{equation}
% \begin{split}
%     \robcoeff(\qA(t),\nomparams, \intparams)  \norm{\robr(t)} \geq &-\robKinf(\robh) + \\ & + \abs{\robr(t)}^\top\wmax(\qA(t), \nomparams, \intparams).
%     \end{split}
% \end{equation}
% \begin{equation}
%     \robcoeff(t) \frac{\norm{\robr(t)}^2}{\norm{\robr(t)}} = \robcoeff(t) \norm{\robr(t)} \geq -\robKinf(\robh) + \norm{\robr(t)}\normwmax.
% \end{equation}
% Choosing $\robcoeff(\qA(t),\nomparams, \intparams) \geq 0$ and 
% \begin{equation}
%     \robcoeff(\qA(t),\nomparams, \intparams) \geq \frac{-\robKinf(\robh)}{\norm{\robr(t)}} + \norm{\wmax(\qA(t), \nomparams)}
% \end{equation}
\begin{equation}
    \robcoeff(\qA(t),\nomparams, \intparams) \geq \frac{-\robKinf(\robh) + \abs{\robr(t)}^\top \wmax(\qA(t), \nomparams, \intparams)}{\norm{\robr(t)}}.
\end{equation}
Simultaneously, choosing $\robcoeff(\qA(t),\nomparams, \intparams) \geq 0$ while satisfying the previous equation ensures that $\robh$ is a minimal CBF.
Computing this $\robcoeff$ requires $\robh$.
Because we do not know $\trueparams$, we can not compute $\robh$.
Instead, we bound $\robh$ over the range of inertial parameters.
$\robhmin$ as in \eqref{eq:CBF_lower_bound} lower bounds $\robh$.
% \begin{equation}
%     \robhmin \coloneqq \inf_{\trueparams^* \in \intparams}\left(-\frac{1}{2}\robr^\top M(q, \trueparams^*) \robr \right) + \roblevel \leq \robh
% \end{equation}
The inequality $\robhmin \leq \robh$ implies that $-\robKinf(\robhmin) \geq -\robKinf(\robh)$, so choosing $\robcoeff(\qA(t),\nomparams, \intparams) \geq 0$ and 
% \begin{equation}
%     \hspace*{-0.2cm} \robcoeff(\qA(t),\nomparams, \intparams) \geq \frac{-\robKinf(\robhmin)}{\norm{\robr(t)}} + \norm{\wmax(\qA(t), \nomparams)}
% \end{equation}
\begin{equation}
    \hspace*{-0.2cm} \robcoeff(\qA(t),\nomparams, \intparams) \geq \frac{-\robKinf(\robhmin) + \abs{\robr(t)}^\top \wmax(\qA(t), \nomparams, \intparams)}{\norm{\robr(t)}}
\end{equation}
ensures that \eqref{eq:CBF_cond} is satisfied and $h$ is a minimal CBF.
% The $\robcoeff$ in \eqref{eq:robust_coeff} satisfies these conditions.
% \begin{equation}
%      \robcoeff(\qA(t),\nomparams, \intparams) = \max\left(0, \frac{-\robKinf(\robhmin)}{\norm{\robr(t)}} + \normwmax\right).
% \end{equation}
% \begin{equation}
%     % \label{eq:robust_coeff}
%     \begin{split}
%     \robcoeff(\qA(t),\nomparams,\intparams) = \max\Big(0, \frac{-\robKinf(\robhmin)}{\norm{\robr(t)}} +  \\ \phantom{ \robcoeff(\qA(t),\nomparams,\intparams) =}  + \norm{\wmax(\qA(t),\nomparams)}\Big),
%     \end{split}
% \end{equation}

\noindent {\bf \eqref{eq:robust_input} is continuous:} 
\eqref{eq:robust_input} could only be discontinuous when $\norm{r(t)} = 0$. 
We prove that for all points in a neighborhood of the point $\norm{r(t)} = 0$, $\robcoeff(\qA(t),\nomparams, \intparams) = 0$ where $\nomparams$ and $\intparams$ are held fixed.
% If we prove this result, then the desired result follows. 

Recall from Ass. \ref{ass:eigenvalue_bound} that there exists $\sigma_M > 0$ such that
% \pat{should we include $\sigma_M$ in Assumption \ref{ass:eigenvalue_bound}?}
\begin{equation}
    \label{eq:lyap_upper_bound}
    -\frac{1}{2} r(t)^\top  M(q(t),\Delta) r(t) \geq -\frac{1}{2} \sigma_M \norm{r(t)}^2
\end{equation}
for all $\Delta \in \intparams$ and $q(t) \in Q$. 
% This follows by noticing that $M$ is a continuous function of $q$ and $\Delta$, and $\intparams$ and $Q$ are compact sets. 
Next, there exists $r_M \geq 0$ such that for all $r(t)$ with $\norm{r(t)} \leq r_M$, we have $-\frac{1}{2} \sigma_M \norm{r(t)}^2 + \roblevel > 0$.
This implies that there exists some $\zeta > 0$ such that $\robhmin > \zeta$ for all $\qA(t)$ with $\norm{r(t)} \leq r_M$.
Because $\alpha$ is an extended class $\mathcal{K}_\infty$ function, this implies that
% \begin{equation}
% \begin{split}
%     \frac{-\robKinf(\robhmin)}{\norm{\robr(t)}} + &\norm{\wmax(\qA(t), \nomparams)} < \frac{-\robKinf(\zeta)}{\norm{\robr(t)}} + \\ & \hspace*{1cm} + \norm{\wmax(\qA(t), \nomparams)}.
%     \end{split}
% \end{equation}
\begin{equation}
\begin{split}
    \frac{-\robKinf(\robhmin)}{\norm{\robr(t)}} + &\frac{\abs{\robr(t)}^\top \wmax(\qA(t), \nomparams, \intparams)}{\norm{\robr(t)}} < \frac{-\robKinf(\zeta)}{\norm{\robr(t)}} + \\ & \hspace*{1cm} + \norm{\wmax(\qA(t), \nomparams, \intparams)},
    \end{split}
\end{equation}
where we have used the fact that $\abs{\robr(t)}^\top  \wmax(\qA(t), \nomparams, \intparams) \leq \norm{\robr(t)}\norm{\wmax(\qA(t), \nomparams, \intparams)}$.
Because $\robKinf(\zeta) > 0$, the previous inequality implies $\frac{-\robKinf(\robhmin)}{\norm{\robr(t)}} + \norm{\wmax(\qA(t), \nomparams)} \to -\infty$ as $\norm{r(t)} \to 0$. 
So, $\robcoeff(\qA(t),\nomparams, \intparams) = 0$, $\forall r(t)$ with $\norm{r(t)} \leq r_M$.
% \pat{we may want to remind readers about the $\gamma \geq 0$ requirement which comes into play here}

% The desired result follows because if $r(0) \leq \ultbound$, then as a result of Assumption \ref{ass:eigenvalue_bound} for all $q(0)$:
% \begin{equation}
%     0 \leq -\frac{1}{2} \sigma_m \norm{r(0)}^2 + V_M \leq -\frac{1}{2} \robr(0)^\top \bM(q(0), \trueparams) \robr(0) + V_M,
% \end{equation}
% which proves that $(r(0),q(0)) \in \robH$.

\end{proof}

\section{Proof of Cor. \ref{cor:tracking_error}}
\label{app:tracking_error}

\begin{proof}
Rearranging \eqref{eq:modified_error}, we get $\errdot(t) = -K_r \err(t) + \robr(t)$.
% Consider \eqref{eq:modified_error}, which we rearrange as
% \begin{equation}
%     \label{eq:error_first_order_system}
% \end{equation}
$K_r$ is diagonal, so this defines $\nq$ first-order linear systems where we treat $\robr$ as an input.
The solution of the $j$\ts{th} system is
% \begin{equation}
%     \errj(t) = \exp(-K_{r, j}t)\errj(0) + \int_{0}^{t} \exp(-K_{r, j}(t - s)) \robrj(s) ds.
% \end{equation}
% By hypothesis $\errj(0) = 0$, yielding
\begin{equation}
    \errj(t) = \int_{0}^{t} \exp(-K_{r, j}(t - s)) \robrj(s) ds.
\end{equation}
because by hypothesis $\errj(0) = 0$.
Using $\abs{\robrj(t)} \leq \epsilon$, we obtain $\abs{\errj(t)} \leq \abs{\exp(-K_{r, j}t) \epsilon \int_{0}^{t} \exp(K_{r, j}s) ds}$.
% \begin{equation}
%     \abs{\errj(t)} \leq \abs{\int_{0}^{t} \exp(-K_{r, j}(t - s)) \epsilon ds}.
% \end{equation}
% Rearranging yields
% \begin{equation}
%     \abs{\errj(t)} \leq \abs{\exp(-K_{r, j}t) \epsilon \int_{0}^{t} \exp(K_{r, j}s) ds}.
% \end{equation}
By solving the integral and noticing that $\exp(-K_{r, j}t) \in [0, 1], \; \forall t \geq 0$, we get the result.
% \begin{equation}
%     \label{eq:position_error_bound}
%     \abs{\errj(t)} \leq \frac{1}{K_{r, j}} \epsilon.
% \end{equation}
The bound on the velocity error follows by using the position error bound, \eqref{eq:modified_error}, and the triangle inequality. 
% From \eqref{eq:error_first_order_system}, we have
% \begin{equation}
%     \abs{\errdotj(t)} = \abs{-K_{r, j} \err(t) + \robr(t)}.
% \end{equation}
% Applying the triangle inequality, we obtain
% \begin{equation}
%     \abs{\errdotj(t)} \leq \abs{K_{r, j} \errj(t)} + \abs{\robrj(t)},
% \end{equation}
% and finally plugging in \eqref{eq:position_error_bound} yields
% \begin{equation}
%     \abs{\errdotj(t)} \leq 2\epsilon.
% \end{equation}

\end{proof}

\section{Proof of Lem. \ref{lem:compute_input}}
\label{app:compute_input}

\begin{proof}
Because IRNEA replaces all operations over the inertial parameters in RNEA with interval arithmetic equivalents, $\texttt{RNEA}(\qA(t), \nomparams, a_0^0) \in \texttt{IRNEA}(\qA(t), \intparams, a_0^0)$ for $\nomparams \in \intparams$.
So, $\wdist(\qA(t), \nomparams, \Delta) \in \iv{\wdistinterval(\qA(t), \nomparams, \intparams)}$ and \eqref{eq:cond_wmax} follows.

The norm, $\max$, and absolute value are all continuous functions, so if we prove that $\setop{inf}([\wdistinterval(\qA(t),\nomparams, \intparams)])$ and $\setop{sup}( [\wdistinterval(\qA(t),\nomparams, \intparams)])$ are continuous in their first argument, then 
$\wmax$ is continuous in its first argument. 
 To prove that, note that $\qA(t)$ is a vector rather than interval argument to IRNEA and is used to generate a homogeneous transformation matrix, which is a continuous function of $\qA(t)$.
 Because all operations in IRNEA are either interval additions, subtractions, or matrix multiplications of the interval inertia parameters with the homogeneous transformation matrix or the velocity/acceleration vectors in $\qA(t)$, the lower or upper bound of the interval output to IRNEA is a continuous function of $\qA(t)$. 
 
 To prove the desired result for $\underline{h}$, note that if we use $q_R(t)$ in place of $\qA(t)$, then when we apply \eqref{eq:modified_ref}, we have $\dot{q}_a(t) = 0$ and $\ddot{q}_a(t) = r(t)$.
 Because IRNEA is computing an interval version of \eqref{eq:nominal_controller} and is using interval arithmetic, $\bM(q(t), \nomparams) r(t) \in \texttt{IRNEA}(q_R(t), \intparams, \zeros)$ for $\nomparams \in \intparams$. 
 Therefore by applying the properties of interval arithmetic, the $\underline{h}$ defined as in \eqref{eq:robhmin_compute} satisfies \eqref{eq:CBF_lower_bound}. 
 The continuity of $\underline{h}$ in its first argument follows by applying the same argument as above.
\end{proof}

\section{Lemma \ref{lem:product_of_dot_products}}
\label{app:product_of_dot_products}
The following lemma is utilized in the proof of Thm. \ref{thm:robust_input_bound}.
\begin{lem} \label{lem:product_of_dot_products}
Given vectors $a, b \in \R^{\nq}$ of unit norm $\norm{a} = \norm{b} = 1$, consider the optimization problem $\max \{ f(c) \mid c \in \R^n_q, \norm{c} = 1\}$,
% \begin{align}
%     &\underset{c \in \R^n_q}{\max} && f(c) \\
%     &&& \norm{c} = 1
% \end{align}
where $f(c) = (a^\top c)( b^\top c)$.
At the optimal solution $c^*$, the cost $f(c^*) = \frac{1 + a^\top b}{2}$.
\end{lem}
\begin{proof}
% Let $x = a^\top c$ and $y = b^\top c$, and consider the quantity $(x - y)^2$.
% We know that $(x - y)^2 \geq 0$, and expanding the left hand side yields $x^2 - 2xy + y^2 \geq 0$.
% Adding $4xy$ to both sides yields $x^2 + 2xy + y^2 \geq 4xy$, which yields $(x + y)^2 \geq 4xy$.
% Therefore, we have
% \begin{equation}
%     \label{eq:dot_product_bound}
%     f(c) \leq \frac{(x + y)^2}{4}.
% \end{equation}
% In the case that $x = y$, note that \eqref{eq:dot_product_bound} actually becomes an equality.

% Consider the scalar quantity $(a^\top c - b^\top c)^2$.
$(a^\top c - b^\top c)^2 \geq 0$, so $(a^\top c)^2 - 2(a^\top c)(b^\top c) + (b^\top c)^2 \geq 0$.
Adding $4 (a^\top c)(b^\top c)$ to both sides, we get $(a^\top c + b^\top c)^2 \geq 4(a^\top c) (b^\top c)$.
Therefore $f(c) \leq \frac{(a^\top c + b^\top c)^2}{4}$.
Note that if $a^\top c = b^\top c$, then $f(c) = \frac{(a^\top c + b^\top c)^2}{4}$.
Using the distributive property on $f(c) \leq \frac{(a^\top c + b^\top c)^2}{4}$ gives $f(c) \leq \frac{((a + b)^\top c)^2}{4}.$
The right hand side is maximized when $c$ points in the $a + b$ direction, so we choose $c^* = \frac{(a + b)}{\norm{a+b}}$.
Plugging in, we obtain $f(c^*) \leq \frac{((a + b)^\top \frac{(a + b)}{\norm{a+b}})^2}{4}$.
Note that $\norm{a+b}^2 = (a + b)^\top (a + b)$, and therefore $f(c^*) \leq \frac{((a + b)^\top (a + b))^2}{4(a+b)^\top(a+b)}$.
Cancelling terms, multiplying, and using the fact that $a^\top a = b^\top b = 1$, we obtain $f(c^*) \leq \frac{1 + a^\top b}{2}.$
Finally, note that $a^\top c^* = b ^\top c^* = \frac{1 + a^\top b}{\norm{a + b}}$, and thus $f(c^*) = \frac{(a^\top c^* + b^\top c^*)^2}{4}$ and we have the desired result.

\end{proof}

\section{Proof of Thm. \ref{thm:robust_input_bound}}
\label{app:robust_input_bound}

\begin{proof}
% We seek a bound on the magnitude of the $j$\ts{th} component of the robust input vector \eqref{eq:robust_input}.
For notational convenience, we drop the dependence on $k$ in $\qA$ and $\robr$.
% From \eqref{eq:robust_input} we have that 
% % \begin{equation}
% %     \norm{\robv(\qA(t;k), \nomparams, \intparams )} = \norm{\robcoeff(\qA(t;k), \nomparams, \intparams )}\frac{\norm{\robr(t;k)}}{\norm{\robr(t;k)}},
% % \end{equation}
% % so we seek to bound $\norm{\robcoeff(\qA(t;k), \nomparams, \intparams )}$.
% \begin{equation}
%     \abs{\robv(\qA(t), \nomparams, \intparams )_j} = \robcoeff(\qA(t), \nomparams, \intparams )\frac{\abs{\robr(t)_j}}{\norm{\robr(t)}},
% \end{equation}
% so we seek to bound $\norm{\robcoeff(\qA(t;k), \nomparams, \intparams )}$.
% From \eqref{eq:robust_coeff}, we have 
% % \begin{equation}
% %     \begin{split}
% %     \robcoeff(\qA(t;k),\nomparams,\intparams) = \max\Big(0, \frac{-\robKinf(\underline{h}(\qA(t;k),\intparams)}{\norm{\robr(t;k)}} +  \\ \phantom{ \robcoeff(\qA(t),\nomparams,\intparams) =}  + \norm{\wmax(\qA(t;k),\nomparams,\intparams)}\Big),
% %     \end{split}
% % \end{equation}
% \begin{equation}
%     \begin{split}
%     \robcoeff(\qA(t),\Delta_0,[\Delta]) = \max\Big(0, \frac{-\robKinf(\robhmin)}{\norm{\robr(t)}} +  \\ \phantom{ \robcoeff(\qA(t),\Delta_0,[\Delta]) =}  + \frac{\abs{r(t)}^\top\wmax(\qA(t),\Delta_0,[\Delta])}{\norm{r(t)}}\Big),
%     \end{split}
% \end{equation}
If $0$ achieves the maximum in the RHS of the expression defining $\robcoeff$ (i.e., \eqref{eq:robust_coeff}), then $\robcoeff(\qA(t;k), \nomparams, \intparams ) = 0$ and \eqref{eq:pz_robust_input_bound} holds trivially.
Therefore, we apply \eqref{eq:robust_input} and bound:
\begin{equation}
    \label{eq:rob_input_bound_long}
    \begin{split}
     \abs{\robv(\qA(t), \nomparams, \intparams )_j} \leq \frac{-\robKinf(\robhmin)}{\norm{\robr(t)}} \frac{\abs{\robr(t)_j}}{\norm{\robr(t)}} + \\
     + \frac{\abs{r(t)}^\top\wmax(\qA(t),\Delta_0,[\Delta])}{\norm{r(t)}}\frac{\abs{\robr(t)_j}}{\norm{\robr(t)}}.
     \end{split}
\end{equation}
In the RHS of the above inequality, note that the first term is negative when $\robhmin > 0$, and that the second term is always positive (because all elements of $\wmax(\qA(t),\Delta_0,[\Delta]$ are $\geq 0$ by \eqref{eq:cond_wmax}).
The RHS is largest when $\robhmin$ is as negative as possible, so we examine this case.
% Because $\robcoeff(\qA(t),\Delta_0,[\Delta])$ is required to be positive, we need only consider the case where $\robhmin \leq 0$ in seeking our bound.
% However, when the right hand argument to the max is larger, we need a bound on $\norm{\wmax(\qA(t;k),\nomparams,\intparams)}$ and $-\robKinf(\underline{h}(\qA(t;k),\intparams))$, while a bound on $\norm{\robr(t;k)}$ is already known from \eqref{eq:ultimate_bound}.
% With this in mind

First, want an upper bound on $\frac{-\robKinf(\underline{h}(\qA(t),\intparams)}{\norm{\robr(t)}}\frac{\abs{\robr(t)_j}}{\norm{\robr(t)}}$ (which occurs when $\robhmin$ is as negative as possible).
% Recall that a bound on $\norm{\robr(t;k)} \leq \epsilon$ is already known from \eqref{eq:ultimate_bound}.
Through the same line of reasoning as in \eqref{eq:lyap_upper_bound} in App. \ref{app:robust_controller_proof}, $-\frac{1}{2} \sigma_M \norm{r(t)}^2 + \roblevel \leq \robhmin$.
Plugging in yields
\begin{equation}
    \frac{-\robKinf(\underline{h}(\qA(t),\intparams)}{\norm{\robr(t)}}\frac{\abs{\robr(t)_j}}{\norm{\robr(t)}}
    \leq 
    \frac{-\robKinf(-\frac{1}{2} \sigma_M \norm{r(t)}^2 + \roblevel)}{\norm{\robr(t)}}
\end{equation}
where we have used $\abs{\robr(t)_j} \leq \norm{\robr(t)}$.
Using $\alpha(x) = \alpha_c x$, we have
\begin{equation}
    \hspace*{-0.2cm} \frac{-\robKinf(-\frac{1}{2} \sigma_M \norm{r(t)}^2 + \roblevel)}{\norm{\robr(t)}}
    = 
    \alpha_c \frac{1}{2}\sigma_M \norm{\robr(t)} - \alpha_c \frac{\roblevel}{\norm{\robr(t)}}.
\end{equation}
The negative sign in front of $\roblevel$ means that we want to divide by as large a value of $\norm{\robr(t)}$ to make this negative quantity as small as possible to construct an upper bound.
Recall that a bound on $\norm{\robr(t)} \leq \epsilon$ is already known from \eqref{eq:ultimate_bound}.
With the uniform bound \eqref{eq:ultimate_bound_defn}, this becomes
\begin{equation}
    \alpha_c \frac{1}{2}\sigma_M \norm{\robr(t)} - \alpha_c \frac{\roblevel}{\norm{\robr(t)}}
    \leq 
    \alpha_c \frac{1}{2}\sigma_M \epsilon - \alpha_c \frac{\roblevel}{\epsilon}.
\end{equation}
Notice from \eqref{eq:ultimate_bound_defn} that $\roblevel = \frac{1}{2}\sigm\epsilon^2$, and substituting gives
\begin{equation}
    \label{eq:rob_input_lyap_bound}
    \frac{-\robKinf(\underline{h}(\qA(t;k),\intparams)}{\norm{\robr(t)}}\frac{\abs{\robr(t;k)_j}}{\norm{\robr(t;k)}}
    \leq 
    \frac{\alpha_c \epsilon(\sigma_M - \sigma_m)}{2}.
\end{equation}

Next, we seek an upper bound on $\frac{\abs{r(t)}^\top\wmax(\qA(t),\Delta_0,[\Delta])}{\norm{r(t)}}\frac{\abs{\robr(t)_j}}{\norm{\robr(t)}}$.
With $\wmax(\pzgreek{*})$ as in \eqref{eq:pz_w_max}
\begin{equation}
    \frac{\abs{r(t)}^\top\wmax(\qA(t),\Delta_0,[\Delta])}{\norm{r(t)}} \leq \frac{\abs{r(t)}^\top \wmax(\pzgreek{*})}{\norm{r(t)}}.
\end{equation}
where $\wmax(\pzgreek{*}) = \wmax(\pzi{\qA},\Delta_0,[\Delta])$ for brevity.
Bringing the norm of $\wmax(\pzgreek{*})$ outside, we know that
\begin{equation}
    \frac{\abs{r(t)}^\top\wmax(\pzgreek{*})}{\norm{r(t)}}\frac{\abs{r(t)}_j}{\norm{r(t)}} = \norm{\wmax(\pzgreek{*})} \left ( \frac{\wmax(\pzgreek{*})^\top}{\norm{\wmax(\pzgreek{*})}} \frac{\abs{r(t)}}{\norm{r(t)}} \hat{z}_j^\top \frac{\abs{r(t)}}{\norm{r(t)}} \right ),
\end{equation}
where $\hat{z}_j^\top$ is a unit vector comprised of all zeros except a $1$ in the $j$\ts{th} dimension.
The quantity in parentheses is a product of two dot products of unit vectors.
Applying Lem. \ref{lem:product_of_dot_products}
\begin{equation}
    \label{eq:rob_input_dist_bound}
    \frac{\abs{r(t)}^\top\wmax(\qA(t),\Delta_0,[\Delta])}{\norm{r(t)}}\frac{\abs{r(t)}_j}{\norm{r(t)}} \leq \frac{\norm{\wmax(\pzgreek{*})} + \wmax(\pzgreek{*})_j}{2}.
\end{equation}
Combine \eqref{eq:rob_input_lyap_bound} and \eqref{eq:rob_input_dist_bound} with \eqref{eq:rob_input_bound_long} to get the desired result.
% \begin{equation}
%     % \label{eq:pz_robust_input_bound}
%     \abs{\robv(\qA(t;k), \nomparams, \intparams )_j} \leq \frac{\alpha_c \epsilon(\sigma_M - \sigma_m) + \norm{\wmax(\pzgreek{*})} + \wmax(\pzgreek{*})_j}{2}.
% \end{equation}

% we have 
% \begin{equation}
%     \norm{\rho(\pzi{w(\pzi{\qA},\nomparams,\intparams)})} \geq \norm{\wmax(\qA(t;k),\nomparams,\intparams)}.
% \end{equation}

\end{proof}

\end{document}